\documentclass[10pt,twocolumn,a4]{article}
\usepackage{booktabs} % For formal tables
\usepackage{subcaption}
\usepackage{mwe}
\usepackage{graphicx}
\usepackage{capt-of}
\usepackage{stackengine}

\usepackage{svg}
\usepackage{xcolor}
\usepackage[export]{adjustbox}
\usepackage{mdframed}
\usepackage{caption}
\usepackage{subcaption}
\usepackage{multirow}
\usepackage{diagbox}
\usepackage{tabularx}
\usepackage[normalem]{ulem}

\usepackage{lipsum} %dummy text

\usepackage[colorinlistoftodos,prependcaption,textsize=tiny]{todonotes}
\usepackage{xargs}

\usepackage[pagenumbers]{cvpr} % To force page numbers, e.g. for an arXiv version

\usepackage{amsmath}
\usepackage{amssymb}
\usepackage{enumitem}% http://ctan.org/pkg/enumitem
\usepackage[pagebackref,breaklinks,colorlinks]{hyperref}

\usepackage{xspace}
\usepackage{setspace}
\usepackage[capitalize]{cleveref}
\crefname{section}{Sec.}{Secs.}
\Crefname{section}{Section}{Sections}
\Crefname{table}{Table}{Tables}
\crefname{table}{Tab.}{Tabs.}

\makeatletter
\@namedef{ver@everyshi.sty}{}
\makeatother
\usepackage{tikz}

\makeatletter
\DeclareRobustCommand\onedot{\futurelet\@let@token\@onedot}
\def\@onedot{\ifx\@let@token.\else.\null\fi\xspace}
 
\def\ie{\emph{i.e}\onedot} 
 
\def\etc{\emph{etc}\onedot}

\makeatother

\newcommand{\norm}[1]{\left\lVert#1\right\rVert}

\newcommandx{\doit}[2][1=]{\todo[linecolor=olive,backgroundcolor=olive!25,bordercolor=olive,#1]{#2}}
\newcommandx{\unsure}[2][1=]{\todo[linecolor=red,backgroundcolor=red!25,bordercolor=red,#1]{#2}}
\newcommandx{\change}[2][1=]{\todo[linecolor=blue,backgroundcolor=blue!25,bordercolor=blue,#1]{#2}}
\newcommandx{\info}[2][1=]{\todo[linecolor=teal,backgroundcolor=teal!25,bordercolor=teal,#1]{#2}}
\newcommandx{\improvement}[2][1=]{\todo[linecolor=Plum,backgroundcolor=Plum!25,bordercolor=Plum,#1]{#2}}
\newcommandx{\thiswillnotshow}[2][1=]{\todo[disable,#1]{#2}}

\newcommand{\chesstable}{\textsc{Chess Table}~}
\newcommand{\colorfountain}{\textsc{Color Fountain}~}
\newcommand{\stove}{\textsc{Stove}~}
\newcommand{\shoerack}{\textsc{Shoe Rack}~}
\newcommand{\garden}{\textsc{Garden}~}
\newcommand{\kitchen}{\textsc{Kitchen}~}
\newcommand\Bstrut{\rule[-0.9ex]{0pt}{0pt}}   % = `bottom' strut

\newcommand{\Comments}[1]{}

\newlength\paramargin
\newlength\figmargin
\newlength\secmargin
\newlength\figcapmargin
\newlength\tabcapmargin

\newcommand{\Paragraph}[1]
{\vspace{2mm} \noindent \textbf{#1}}

\newcommand{\sqboxs}{1.2ex}% the square size
\newcommand{\sqbox}[1]{\textcolor{#1}{\rule{\sqboxs}{\sqboxs}}}
\definecolor{yellow}{HTML}{E6E600}
\definecolor{green}{HTML}{33CC33}
\definecolor{red}{HTML}{E60000}
\begin{document}
    %% HEADER : TITLE + AUTHOR info
    \title{Interactive Segmentation of Radiance Fields}
\author{
Rahul Goel\footnotemark[1]
\and
Dhawal Sirikonda\footnotemark[1]
\and
Saurabh Saini
\and
P J Narayanan\and \\[-2mm]
CVIT, Kohli Center on Intelligent Systems (KCIS), IIIT Hyderabad\\
{\tt\small {\{rahul.goel, dhawal.sirikonda, saurabh.saini\}@research.iiit.ac.in}},
{\tt\small pjn@iiit.ac.in}
}
    
    \twocolumn[{
    \renewcommand\twocolumn[1][]{#1}
    \maketitle
    \begin{center}
    \includegraphics[width=1.0\linewidth]{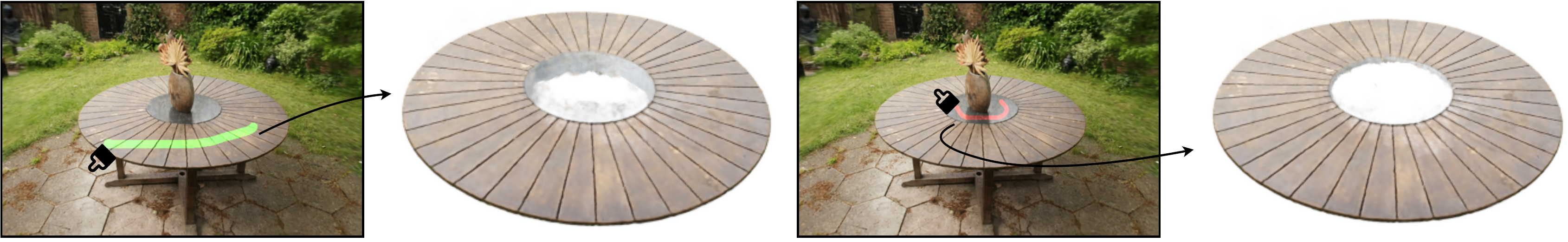}
    \captionof{figure}{
    We present ISRF, an interactive method to segment objects in radiance fields. Users can draw positive strokes to segment multiple objects at a time in 3D and negative strokes to remove unwanted regions repeatedly. In the figure, the \textsc{Wooden Table Top} is segmented using one positive and one negative stroke as shown.}
    \label{fig:teaser}
\end{center}
    }]
    
    % \maketitle
    %% -------------------------------------------------------------
    %% ABSTRACT + TEASER
    \begin{abstract}
Radiance Fields (RF) are popular to represent casually-captured scenes for new view synthesis and several applications beyond it. Mixed reality on personal spaces needs understanding and manipulating scenes represented as RFs, with semantic segmentation of objects as an important step. Prior segmentation efforts show promise but don't scale to complex objects with diverse appearance. We present the ISRF method to {\em interactively} segment objects with fine structure and appearance. Nearest neighbor feature matching using distilled semantic features identifies high-confidence seed regions. Bilateral {search} in a joint spatio-semantic space grows the region to recover accurate segmentation. We show state-of-the-art results of segmenting objects from RFs and compositing them to another scene, changing appearance, etc., and an interactive segmentation tool that others can use.
\end{abstract}
    \let\thefootnote\relax\footnote{Project Page: \href{https://rahul-goel.github.io/isrf/}{https://rahul-goel.github.io/isrf/}}
    \footnotetext[1]{$^\ast$ Equal Contribution}
    % \input{src/figures/0_teaser}
    %% -------------------------------------------------------------
    %% KEYWORDS
    % \input{src/imports/4_keywords}
    %% -------------------------------------------------------------
    
    %% -------------------------------------------------------------
    %% SECTIONS
    \begin{figure*}
    \centering
    \begin{minipage}{\linewidth}
        \centering
         \includegraphics[width=\textwidth]{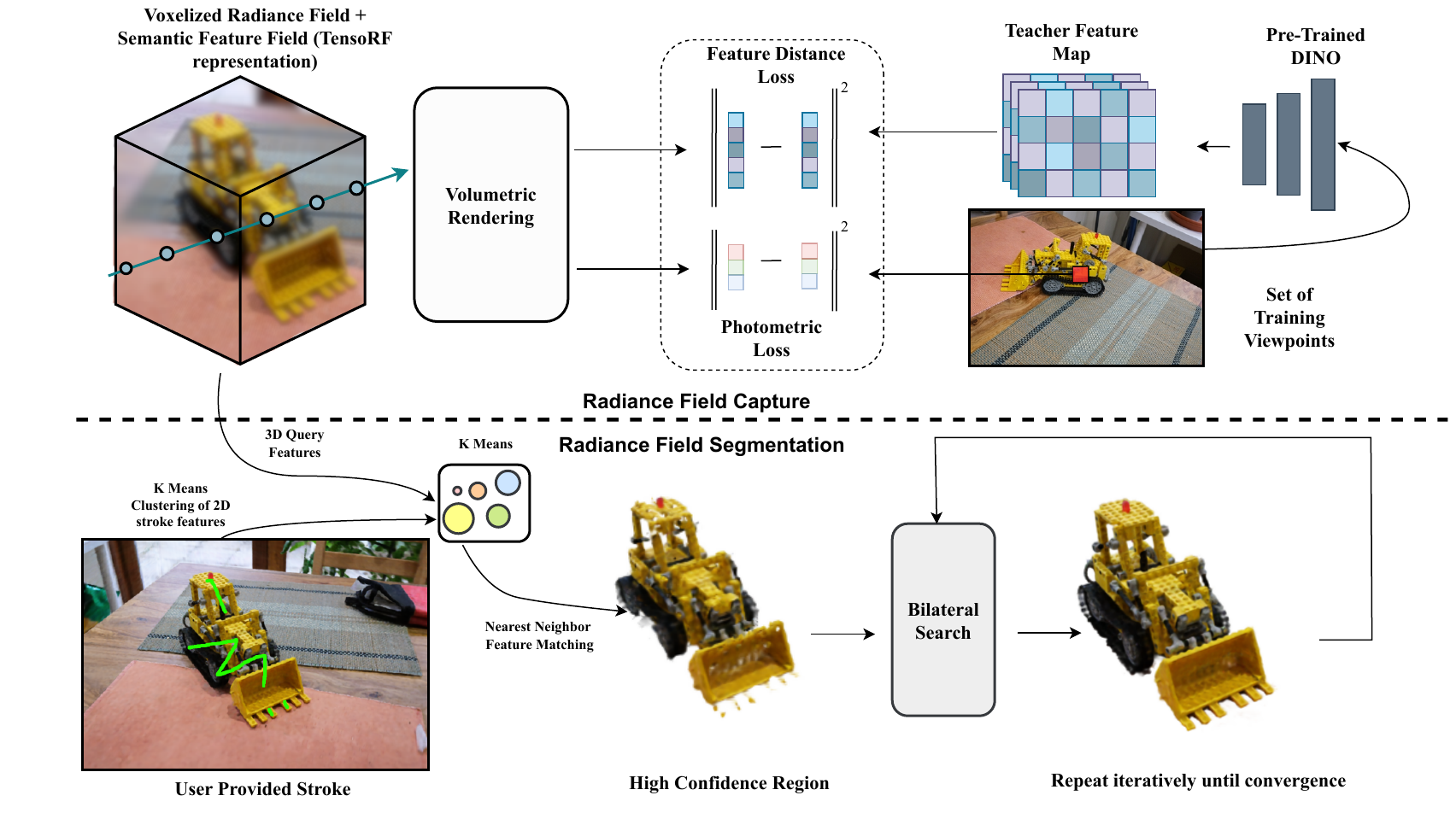}
    \end{minipage}
        \caption{\emph{ISRF System overview}: We capture a 3D scene of voxelized radiance field and distill the semantic feature into it. Once captured, the user can easily mark regions using a brush tool on a reference view (green[\sqbox{green}] stroke). The features are collected corresponding to the marked pixels and clustered using K-Means. The voxel-grid is then matched using NNFM (nearest neighbor feature matching) to obtain a high confidence seed using a tight threshold. The seed is then grown using bilateral search to smoothly cover the boundaries of the object, conditioning the growth in the spatio-semantic domain.}
    \label{fig:sysDiag}
\end{figure*}

    \section{Introduction}
{
    \label{sec:intro}    
    Scene representation is a crucial step for any scene understanding or manipulation task. Relevant scene parameters, be it shape, appearance, or illumination, can be represented using various modalities like $2$D (depth/texture) maps, point clouds, surface meshes, voxels, parametric functions, \etc. Each modality has its strengths and weaknesses.  For example, shape correspondence is straightforward between point clouds compared to surface meshes but compromises rendering fidelity. Thus, choosing an appropriate representation has a major impact on downstream analyses and applications.
        
    Neural implicit representations have emerged as a promising modality for 3D analysis recently. Although initially proposed only for shapes \cite{deepsdf, occupancy_networks}, they have been extended to encode complete directional radiance at a point \cite{nerf}, other rendering parameters like lightfields, specularity, textual context, object semantics, \etc \cite{lightfieldNerf, pandora, nerf_audio, nerf_text, nerf_semantic_1, nerf_semantic_2, nerf_semantic_3}. The representation was extended beyond static inward-looking and front-facing scenes to complex outward-looking unbounded $360^{\circ}$ views, dynamic clips, occluded egocentric videos, and unconstrained images.
    
    Radiance fields have also been used beyond Novel View Synthesis (NVS) for other applications \cite{cla_nerf, nerf_supervision, loc_nerf, nerf2real, block_nerf, city_nerf, iron, nerf_nonrigid, hyper_nerf}. Segmenting objects of the scene representation is a first step towards its understanding and manipulation for different downstream tasks. There have been a few efforts at segmenting and editing of radiance fields. Recently, N3F \cite{N3F}, and DFF \cite{DFF} presented preliminary solutions to this in the neural space of radiance fields. Both use distillation for feature matching between user-provided cues with the learned 3D feature volume, with N3F using user-provided patches and DFF using textual prompts or patches as the segmentation cues. These methods struggle to segment objects with a wide appearance variation. The NVOS system provides segmentation with strokes but have poor quality and non-interactive computations \cite{nvos}.
     
    In this paper, we present a simple and efficient method to interactively segment objects in a radiance field representation. Our ISRF method uses an intuitive process with the user providing easy strokes to guide it interactively. We use the fast and memory-efficient TensoRF representation \cite{tensorf} to train and render. TensoRF uses an explicit voxel representation that is more amenable to manipulation. We include a DINO feature \cite{dino} at every voxel to facilitate semantic matching from 2D to 3D. DINO features are trained on a large collection of images and are known to capture semantics effectively. We condense the DINO features from the user-specified regions to create a fixed-length set using K-Means. A nearest neighbor feature matching (NNFM) on this set in the 3D voxels identifies a high-confidence  \emph{seed region} of the object to be segmented. The seed region is grown using a bilateral filtering-inspired search to include neighboring proximate voxels in a joint feature-geometric space. We show results of segmenting several challenging objects in forward facing \cite{mildenhall2019llff} and 360 degrees \cite{mipnerf360} scenes. The explicit voxel space we use facilitates simple modification for segmenting objects. We also show examples of compositing objects from one RF into another. In summary, the following are the core contributions of ISRF:
    \begin{itemize}[label={$\circ$}]
    {
        \itemsep=0mm
        \item An easily interpretable and qualitatively improved 3D object segmentation framework for radiance fields. 
        \item Interactive modification of segmentation to capture fine structure, starting with high-confidence matching. Our representation allows a spatio-semantic bilateral search to make this possible. The framework can also use other generalized distances to grow the region for specific applications.
        \item A hybrid implicit-explicit representation that is memory-efficient and fast to render also facilitates the distillation of semantic information for improved segmentation. Our results show improved accuracy and fine-grain object details in very challenging situations over contemporary efforts.
        \item An easy-to-use, GUI based tool to interactively segment objects from an RF representation to facilitate object replacement, alteration, \etc.
        \item Consistent 2D/3D segmentation masks for a few scenes and objects created manually using our method to facilitate future work in segmentation, manipulation, and understanding of RFs.
    }
    \end{itemize}
}
    \section{Related Work}
{
   \label{sec:related}
    Radiance field research work is extensive and fast-growing. Hence, we restrict our related works discussion to three relevant topics, \ie, hybrid representations, manipulation of radiance fields and feature-encoded semantics. For a more comprehensive background, we encourage the reader to refer to the latest surveys in this area \cite{nerf_survey, nerf_survey2}.

    \Paragraph{Hybrid Representations:} 
    % The field of NVS has been a center of interest for a long time in the vision community.
    In the past several years, various representations have been employed for the NVS applications \cite{multi_plane_images, lumigraph, plenoptic}. The latest line of works based on implicit volumetric representations \cite{nerf,nerf++,mipnerf360} specifically has shown great promise by leveraging the Radiance Fields (RF)\cite{radianceFields} for comprehensive scene representation. NVS, from the perspective of radiance fields, involves volumetric rendering \cite{nerf} from a particular viewpoint. However, despite its vast utility, NeRF demands tens of hours of training time per scene. The computational overhead issue has been the focus of subsequent works like PlenOctrees\cite{plenoctrees}, KiloNeRF\cite{kilonerf}, \etc. which borrows efficient techniques from the traditional 3D literature. Later, Plenoxels\cite{plenoxels} and DVGO\cite{dvgo,dvgov2} have advanced on this front by harnessing the lattice structure in a hybrid representation with implicit field features encoded on an explicit spatial grid. This significantly reduces the training time overhead to 5-10 minutes per scene by trading it off with increased storage requirements. InstantNGP \cite{instantngp} reduced the training time to few seconds using multilevel hash encodings. Recently, TensoRF\cite{tensorf} proposed a tensor-decomposed set of matrix-vector representation for the radiance feature lattice structure, which addresses both the storage and time overhead issues. We base our method on TensoRF representation to exploit this gained efficiency and explicit geometric information.

    \Paragraph{Editing:} The advent of RFs has paved a principled way for altering the appearance of 3D scene content. Many extended this approach to solve problems in varied domains on the editing problem. More specifically, works like \cite{neural_pil, extracting_triangle, nerfactor, nerv2020, NeRD} have disentangled the photo-realistic rendering equation to account for the material and lighting edits. Others like \cite{snerf_siggraph} and \cite{stylizednerf_cvpr, arf_eccv} have aimed to alter the appearance via post-hoc image-based stylization modules \cite{gatys, johnson}. Apart from such appearance edits, methods like \cite{neumesh}, \cite{cagenerf} have concentrated on geometric deformations of object-centric scenes represented as Radiance Fields. Our proposed method allows both appearance and geometric scene manipulations.

    \Paragraph{Semantics:} For the scene, understanding the semantic information of the scene is essential; still, only a few solutions have been proposed in this area. Initial methods like Semantic NeRF \cite{semanticnerf}  tried directly regressing semantic labels in the novel views from sparse priors. A few leveraged deep image features like DINO \cite{dino} and LSeg \cite{lseg} to attribute semantics to the 3D scene points. N3F\cite{N3F} and DFF\cite{DFF} demonstrate object-specific segmentation using deep semantics. Though these methods support segmentation, interactive content addition and removal are not supported by them, as the underlying scene representation is an implicit neural function that prohibits easy alterations and extensions into other application scenarios.
    
    In this work, we use a 2D-3D distillation-based approach similar to N3F and DFF, but focus on fine-grained interactive segmentation. Initial works like GrabCut\cite{grabcut} and its variants utilized positive and negative user strokes to obtain the correct segmented objects. Subsequent variants \cite{mask_dino, fbrs} have augmented this by leveraging deep learning in the temporal and non-temporal domains. NVOS \cite{nvos} follows a 3D variant of GrabCut using the positive and negative strokes for segmentation of scenes represented as RFs and MPI\cite{multi_plane_images} but struggles to produce faithful segmentation while incurring significant performance overhead. We draw inspiration from them and build upon the proven methods like semantic features, nearest neighbor matching, and voxel carving techniques \cite{space_carving, guided_filtering} by extending them to radiance fields. We experimentally prove how such simple techniques combined with the appropriate scene representation can improve result quality and fine details while simultaneously being quite intuitive, interpretable, and efficient (80$\times$ faster than NVOS). 
}
    \section{Method}
{
    \label{sec:method}
    We first provide the basics on radiance fields and the feature distillation strategy related to our scene representation. We then detail our proposed interactive segmentation workflow comprising 2D-3D feature matching, region growing, and manipulation techniques on this learned representation.

    \subsection{Radiance Field Representation}
    {
        \label{subsection:method_background_rf}
        A radiance field (RF) \cite{radianceFields} $\mathcal{F}$ maps the scene radiance values as view dependent RGB color $c\in \mathbb{R}^3$, given a continuous point $x \in \mathbb{R}^3$ and viewing direction $d \in \mathbb{S}^2$ in space as inputs: $ \mathcal{F}(x,d): \mathbb{R}^3 \times \mathbb{S}^2 \rightarrow \mathbb{R}^3. $
        
        NeRF \cite{nerf}, and its variants \cite{nerf++,mipnerf360,nerf-w} encoded this mapping as the neural function using an MLP, with a low memory footprint but high training and rendering overhead. They also store scalar point density $\sigma \in \mathbb{R}$ which is used for differentiable volumetric rendering %Eq \ref{eq:volume_rendering}
        to train the network:

       \begin{align}
            \label{eq:volume_rendering}
            \small
            \hat{C}(r) = \left( \textstyle\sum_{i=i}^{K} T_i \alpha_i c_i \right)
            \qquad \textrm{where} \\
            \alpha_i = 1 - e^{-\sigma_i \delta_i} 
            \qquad \textrm{and} \qquad
            T_i = \textstyle\prod_{j=1}^{i-1} (1 - \alpha_j).
        \end{align}
       Here for a given point $i$ along a ray, $\delta_i$ is the distance to the sampled point, $T_i$ is the accumulated transmittance, and $c_i$ is the view-dependent color for the point.
       Later efforts like Plenoxels \cite{plenoxels}, and DVGO \cite{dvgo} stored the field variables in a lattice structure akin to a 3D voxel grid, significantly improving the training and rendering times at the cost of high storage requirements. These quantized values are trilinearly interpolated and decoded to render color value at any point. The grid structure provides easy spatial context and explicit representation leading to higher efficiency. Recently, TensoRF \cite{tensorf} proposed a matrix-vector decomposition representation of this lattice, reducing storage requirements while facilitating efficient training and view generation.
       We use TensoRF as the basis of our work. The top part of \cref{fig:sysDiag} shows our radiance field capture step, with the volume represented using TensoRF. In the case of the quantized representation of radiance fields, the radiance is obtained as follows:
       \begin{equation}
            \begin{split}
            \label{eq:trilinear_radiance}
            \sigma_i = \psi(V^{\sigma}, x_i) 
            \quad \textrm{and} \quad  
            c_{i} = \mu^{\scriptstyle{rbg}}_{\theta}(\psi(V^{f}, x_i), d).
            \end{split}
       \end{equation}
       Here $\sigma$ is the density of the volumetric space, $V^{f}$ the radiance feature grid of appearance features $f$, and $\psi$ indicates trilinear interpolation. While rendering a given sample point $x_i \in \mathbb{R}^3$ along the ray direction $d$, a small decoding MLP $\mu^{\scriptstyle{rbg}}_{\theta}(f_i, d) \rightarrow c_i$ is evaluated. The final color of a ray is calculated by combining all sample colors $c_i$ at every point $x_i$ along it using the  \cref{eq:volume_rendering}. This is used to reduce the photometric loss $\mathcal{L}^{(rgb)}$ optimizing for both the radiance feature lattice $V^f$ and parameters $\theta$ of MLP ($\mu$).
       
    }
    
    \subsection{Semantic Features Distillation}
    {
        \label{subsection:method_background_fd}
     
        Object segmentation requires knowledge of scene semantics. We include an additional feature into the radiance field for this. In order to attribute semantics to the radiance field, we distill contextual knowledge from a large pre-trained teacher model similar to the prior art \cite{N3F, DFF}. Specifically, our teacher is a vision transformer model trained using self-supervision and is shown to pay attention to semantically meaningful objects in the scene in a class-agnostic manner. This knowledge from the teacher is distilled into the student radiance field in addition to the color and density values as point semantic features $\phi \in \mathbb{R}^m$. Thus the mapping now becomes: $ \mathcal{F}(x,d): \mathbb{R}^3 \times \mathbb{S}^2 \rightarrow \mathbb{R}^3 \times \mathbb{R} \times \mathbb{R}^m.$ 
        More concretely, we use 2D semantic features using the DINO ViT-b8 model \cite{dino} for each input posed image.
        Recent efforts \cite{N3F, DFF} also use DINO; unlike them, we directly optimize for the features on the voxel grid in the TensoRF representation without a neural network. We also do not encode the direction dependence in these semantic features since the object semantics are direction agnostic. We trilinearly interpolate the distilled semantic feature $\phi_{i} = \psi(V^{\phi}, x_i)$
        for a point $x_i$ from the learned feature lattice $V^{\phi}$. We combine the $\phi_i$ along the ray using the \cref{eq:volume_rendering} like color $c_i$. The TensoRF representation is optimized to minimize the total loss
        \begin{equation}
            \label{eq:weighted_loss}
            \mathcal{L} = \mathcal{L}_{rgb} + \lambda \mathcal{L}_{feature}
        \end{equation}
        to obtain the final radiance field with $\phi$, $V^f$, and $V^\phi$. Both losses $\mathcal{L}_{rgb}$ and $\mathcal{L}_{feature}$ are calculated using $L^2$ norm. %(Refer to Fig \ref{fig:sysDiag})
        
        High-resolution feature rendering results in high-frequency feature fields similar to N3F \cite{N3F}. (See the supplementary document for distilled feature field visualizations.)
        Explicit semantic features at every point open the way to adapt traditional 3D analysis techniques to radiance fields in a semantically meaningful fashion. Segmenting objects in 3D voxel space and using bilateral filtering inspired search are examples that go beyond what prior neural representations have shown.    
    }

    \subsection{2D-3D Feature Matching}
    {
        \label{sub_sec:feature_matching}
        For object segmentation, the user picks a(few) reference views and annotates the regions of interest using a brush stroke. Semantic DINO features associated with the marked pixels are collected.
        DINO features were shown to fare well using 1-NN feature matching for good 2D semantic segmentation \cite{dino}. However, a single DINO feature will not suffice to segment complex objects with diversity. We cluster the input features using K-Means to obtain a fixed-size exemplar set of features for matching in 3D space. We use nearest neighbor feature matching (NNFM) on the exemplar set to label each voxel as foreground or background. The result is stored in a 3D bitmap. In this step, we use a tight threshold to identify a high-confidence seed region, which is processed further. Prior methods \cite{DFF, N3F} used a single averaged semantic feature from the user-specified patch to match 2D to 3D. Their implicit neural representation can only be segmented after $\phi$ values are rendered. Feature matching methods like NNFM are too \emph{costly} to evaluate at every point on the ray using a neural representation.

       The segmentation results can also be precomputed and stored, facilitating downstream tasks like view generation and editing on the fly without repeated processing.
    }
    
    \subsection{Region Growing}
    {
        \label{sub_sec:region_growing_blf}
        The high confidence seed region $(M^0)$ from the previous step is grown in the volume-space to delineate the complete object volume. We do this in joint spatio-semantic space to include proximate voxels that are also semantically close. We adopt a {\em Bilateral Filtering}\cite{bilateral_filter} inspired search dubbed as {\em Bilateral Search} on the voxel grid using the spatial feature $x$ and semantic feature $\phi$ values as filter's domain and range kernels, respectively.
        We iteratively grow the current bitmap region $M^r$ till convergence, as given below.
            \begin{align*}
            \label{eq:bilateral_filtering}
                % \begin{split}
                \small
                    M^{r+1}(x) = {\cal T}_\tau(\dfrac{1}{W} \sum_{x_i \in \Omega_x} M^{r}(x_i) \,
                    g_{\sigma_\phi}(\phi_{i}^{2}) \,
                    g_{\sigma_s}(s_{i}^{2}))\\
                    \textrm{where} \quad 
                    \phi_i = \norm{\phi_{x_i} - \phi_{x}}, \quad
                    s_i = \norm{{x_i} - {x}} \\
                    \textrm{and} \quad
                    W = \sum_{x_i \in \Omega_x} g_{\sigma_f}(\phi^2_i) \, g_{\sigma_s}(s^2_i).
                % \end{split}
            \end{align*}
        Here $M^r$ is the $r^{th}$ iteration of filtering; $\phi_x$ is the distilled semantic feature at point $x$ in the volumetric space; $g_{\sigma}$ is the Gaussian smoothing functions with variance $\sigma$; ${\cal T}_\tau$ is binary thresholding against value $\tau$; and $\Omega_x$ is the immediate voxel neighbors of $x$. 
        We find that $\tau = 0.2$ works well for our scenes. The seed region expands to the boundaries of the desired object in a few iterations of bilateral filtering.
        \begin{figure}[b]
    \centering
    \begin{minipage}{0.49\linewidth}
        \centering
        \subcaptionbox{Stroke 1}
        {\includegraphics[width=\linewidth]{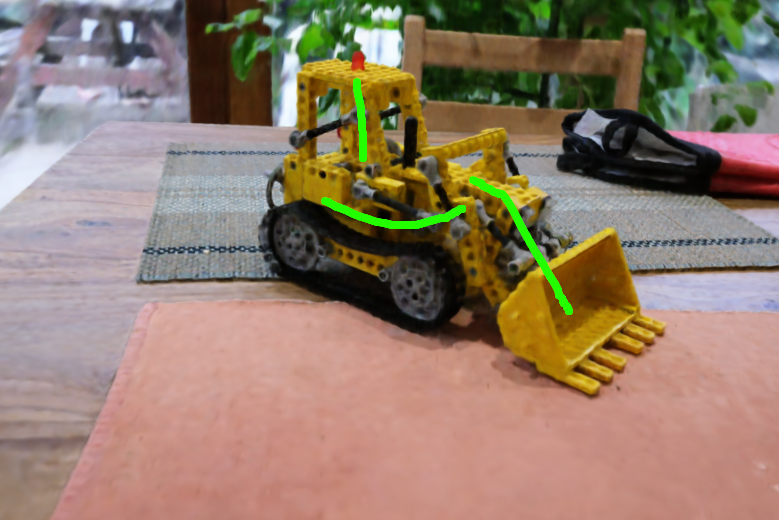}}
    \end{minipage}
    \begin{minipage}{0.49\linewidth}
        \centering
        \subcaptionbox{Output 1}
         {\includegraphics[width=\linewidth]{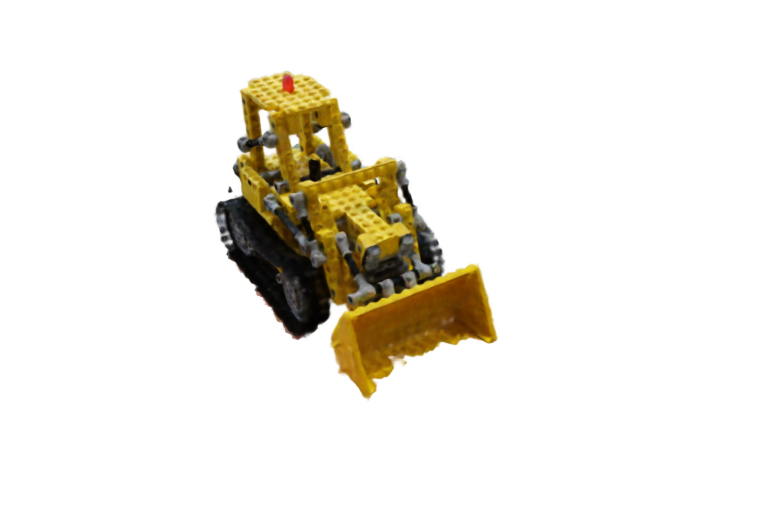}}
    \end{minipage}
    
    \begin{minipage}{0.49\linewidth}
        \centering
        \subcaptionbox{Stroke 2}
        {\includegraphics[width=\linewidth]{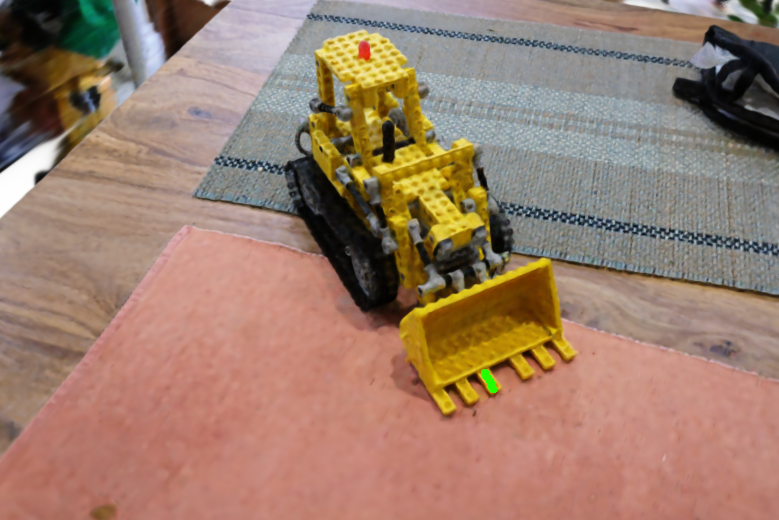}}
    \end{minipage}
    \begin{minipage}{0.49\linewidth}
        \centering
        \subcaptionbox{Merged Output}
        {\includegraphics[width=\linewidth]{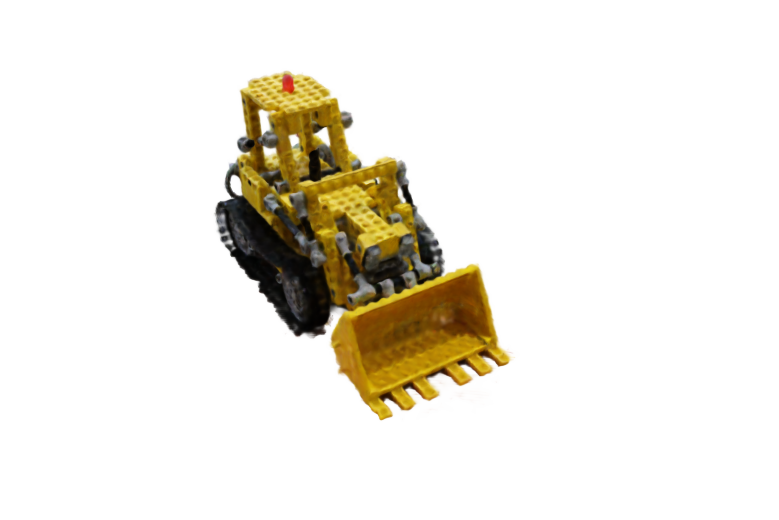}}
    \end{minipage}
    \caption{\emph{Multiple Positive Strokes}:
    {When the method fails to capture some of the details using initial set of strokes, the user can iteratively add more positive strokes to recover the desired object. (a) depicts the initial strokes which lead to missing teeth as shown in (b). Addition of a small stroke on one of the teeth (c) and followed by region grown captures full-details as shown in (d).}}
    \label{fig:double_positive}
\end{figure}
    }

    \subsection{User Interactivity}
    {
        \label{sub_sec:interacitvity}
        Region growing results in a stable voxel content based on the input strokes. The user can add or remove parts interactively if the extracted content misses out on a few details or when some extraneous content floods into the segmented region. We use positive and negative strokes to add and remove the content in the image space, as followed by methods like GrabCut \cite{grabcut}. The mask of the negative segment is subtracted from the mask of the positive segment to get the final segmented objects. We find practically that even complex objects can be segmented well with a few positive and negative strokes, as shown in the results in the paper and in the supplementary material. Additionally, our method provides interactive feedback for every stroke (as can be seen in \cref{tab:time_taken}) that allows users to segment interactively unlike methods like NVOS\cite{nvos}. Implementation details have been reported in the supplementary document.
    } 

}
    \begin{figure*}[!ht]
    \centering
    \rotatebox[origin=c]{90}{Reference \Bstrut}\begin{minipage}{0.24\linewidth}
    \centering
    % \stackinset{l}{}{b}{}
    % {\fcolorbox{black}{green}{\includegraphics[scale=0.15]{assets/images/cmp/styles/santamaria.jpg}}}
    \frame{\includegraphics[width=\textwidth]{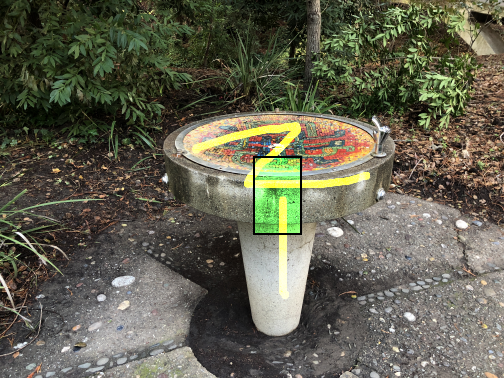}}
    \end{minipage}
    \begin{minipage}{0.24\linewidth}
        \centering
         \frame{\includegraphics[width=\textwidth]{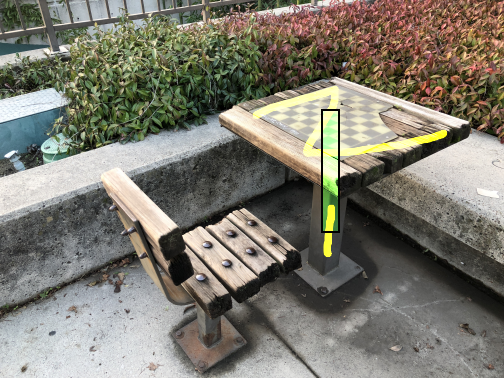}}
    \end{minipage}
    \begin{minipage}{0.24\linewidth}
        \centering
         \frame{\includegraphics[width=\textwidth]{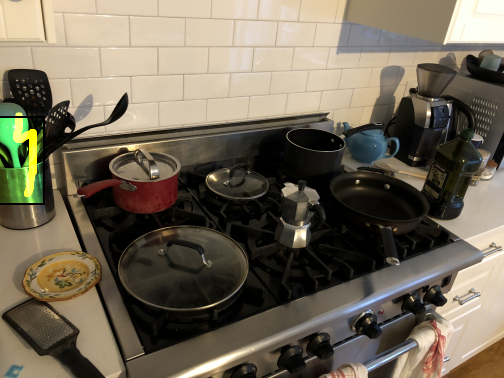}}
    \end{minipage}
    \begin{minipage}{0.24\linewidth}
        \centering
         \frame{\includegraphics[width=\textwidth]{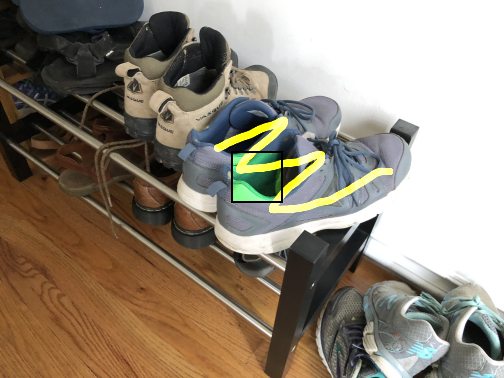}}
    \end{minipage}
    
    \rotatebox[origin=c]{90}{N3F \cite{N3F}/DFF \cite{DFF} \Bstrut}\begin{minipage}{0.24\linewidth}
        \centering
        % \stackinset{l}{}{b}{}
        % {\fcolorbox{black}{cyan}{\includegraphics[scale=0.15]{assets/images/cmp/styles/udnie.jpg}}}
        \frame{\includegraphics[width=\textwidth]{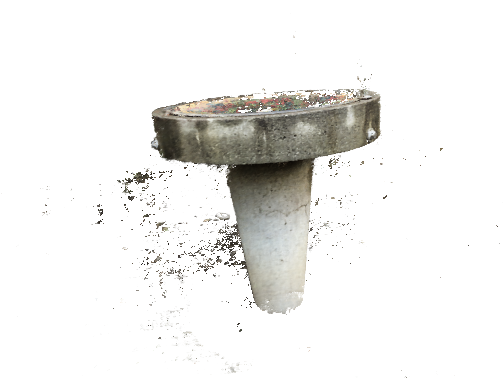}}
    \end{minipage}
    \begin{minipage}{0.24\linewidth}
        \centering
         \frame{\includegraphics[width=\textwidth]{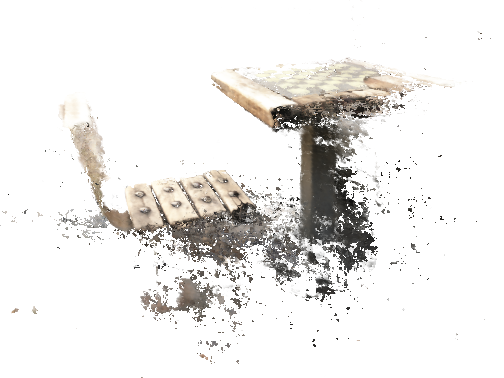}}
    \end{minipage}
    \begin{minipage}{0.24\linewidth}
        \centering
         \frame{\includegraphics[width=\textwidth]{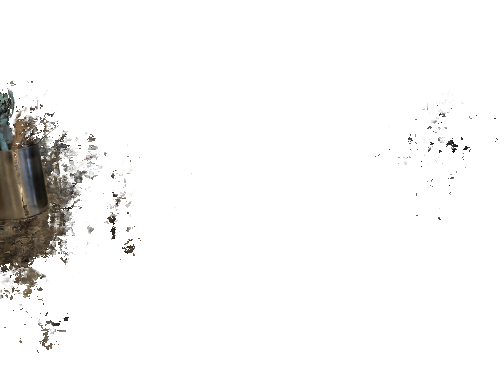}}
    \end{minipage}
    \begin{minipage}{0.24\linewidth}
        \centering
         \frame{\includegraphics[width=\textwidth]{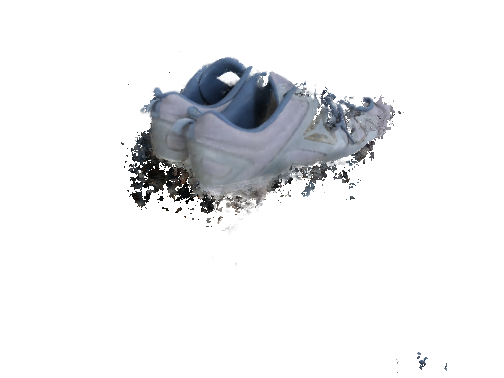}}
    \end{minipage}
    
    \rotatebox[origin=c]{90}{Ours (patch) \Bstrut}\begin{minipage}{0.24\linewidth}
        \centering
        \frame{\includegraphics[width=\textwidth]{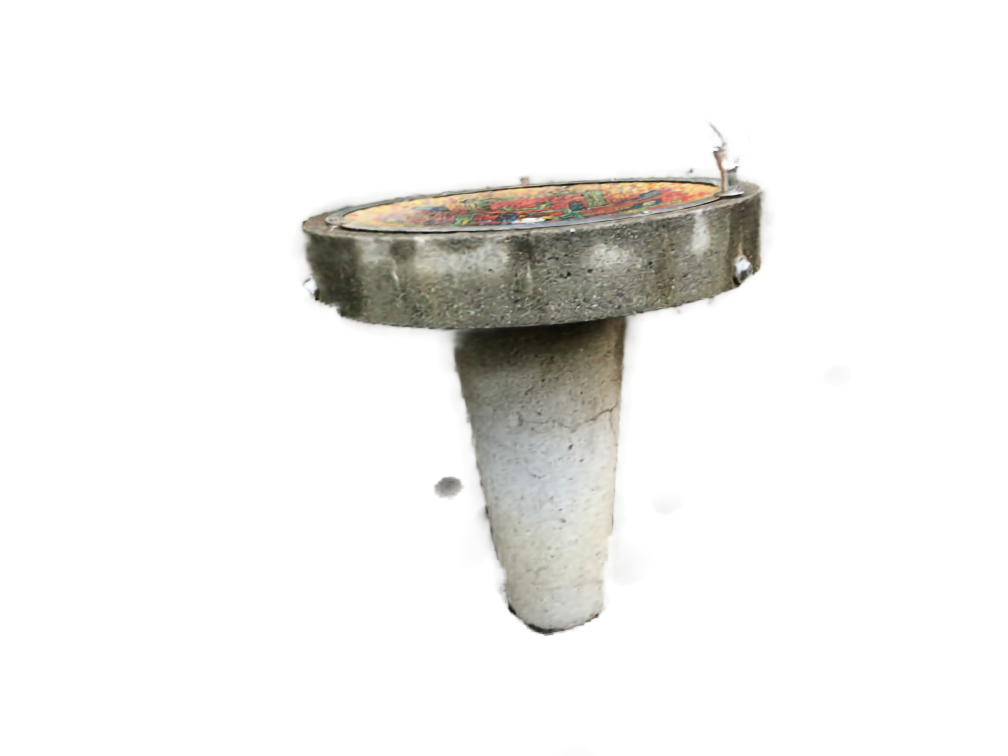}}
        % \stackinset{l}{}{b}{}
        % {\fcolorbox{black}{orange}{\includegraphics[scale=0.65]{assets/images/cmp/styles/mediter.jpg}}}
    \end{minipage}
    \begin{minipage}{0.24\linewidth}
        \centering
         \frame{\includegraphics[width=\textwidth]{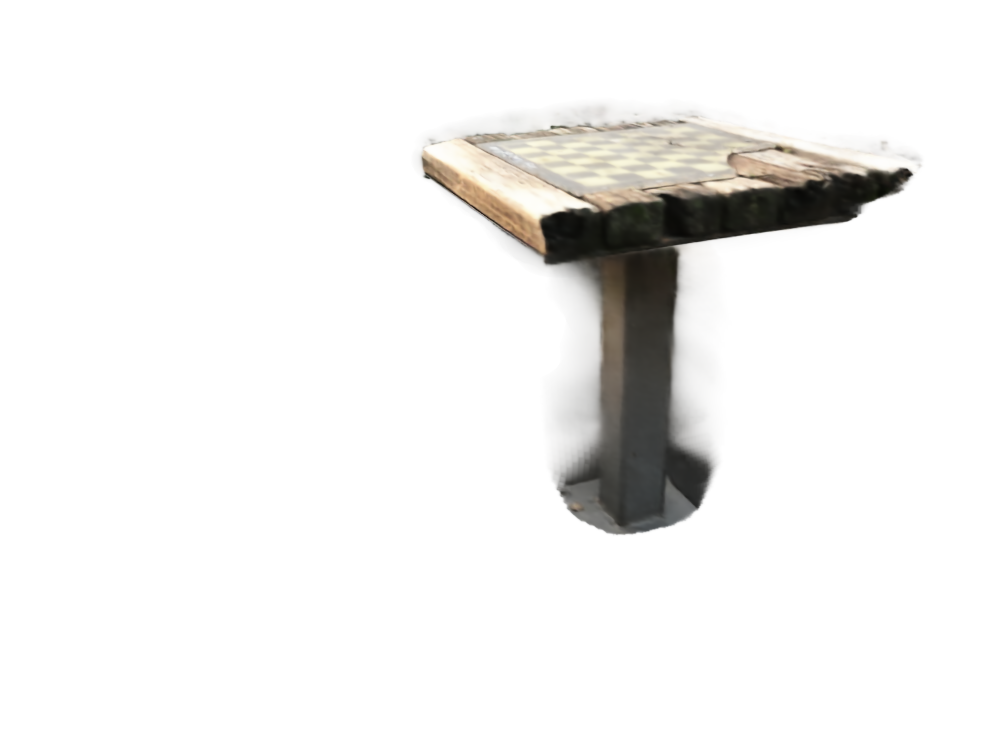}}
    \end{minipage}
    \begin{minipage}{0.24\linewidth}
        \centering
         \frame{\includegraphics[width=\textwidth]{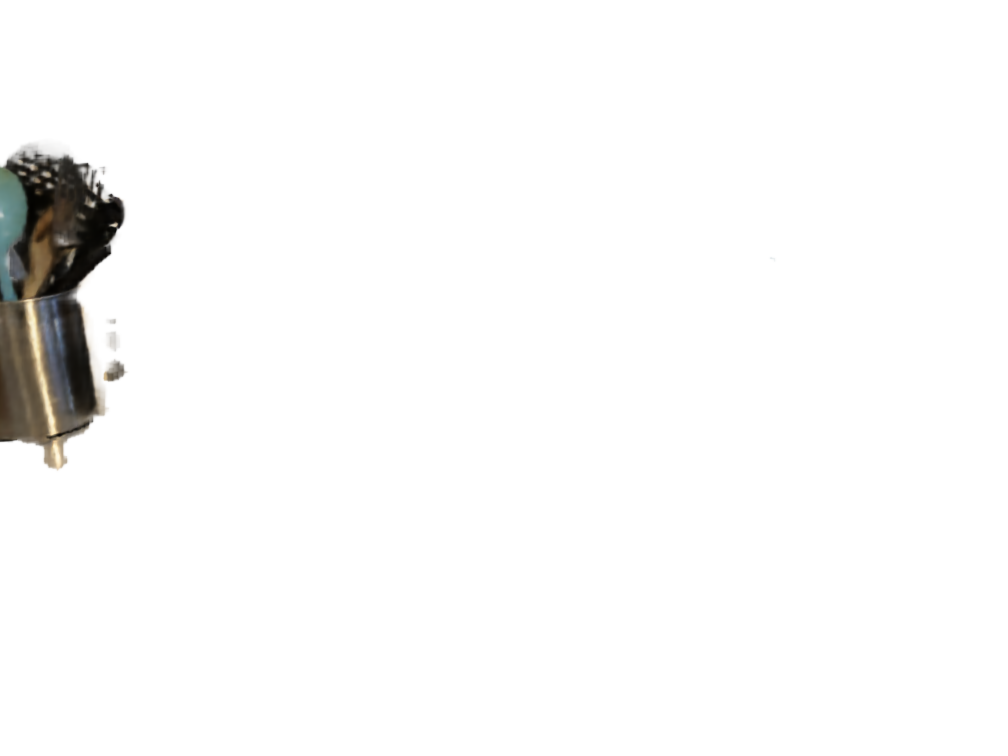}}
    \end{minipage}
    \begin{minipage}{0.24\linewidth}
        \centering
         \frame{\includegraphics[width=\textwidth]{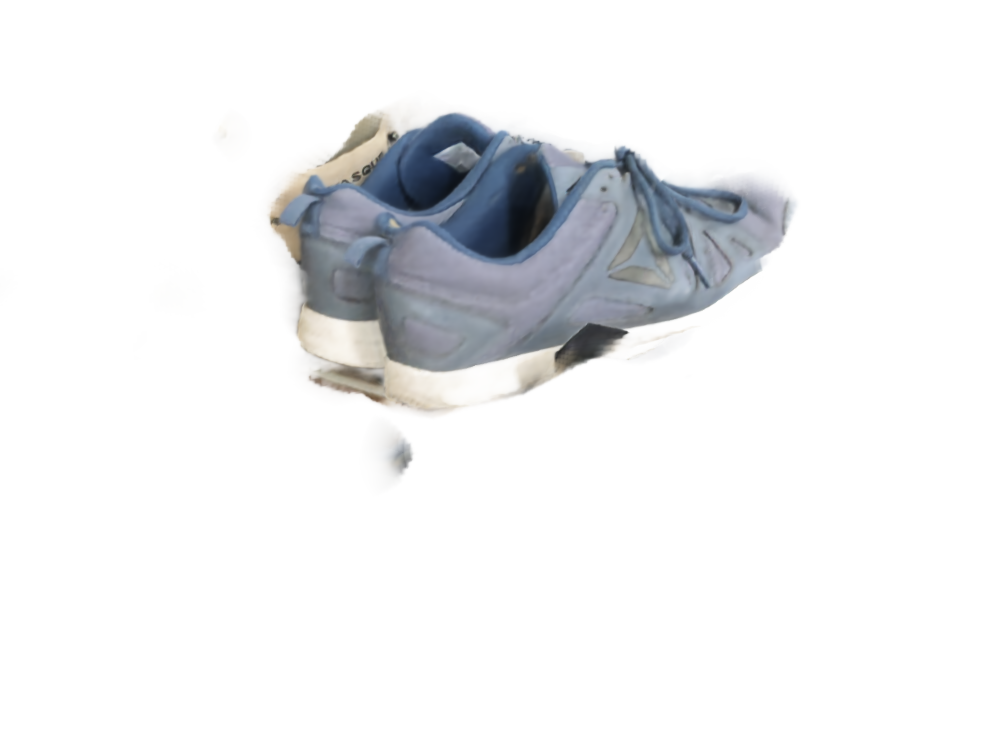}}
    \end{minipage}
    
    \rotatebox[origin=c]{90}{Ours (stroke) \Bstrut}\begin{minipage}{0.24\linewidth}
        \centering
        \subcaptionbox{\colorfountain}
        {\frame{\includegraphics[width=\textwidth]{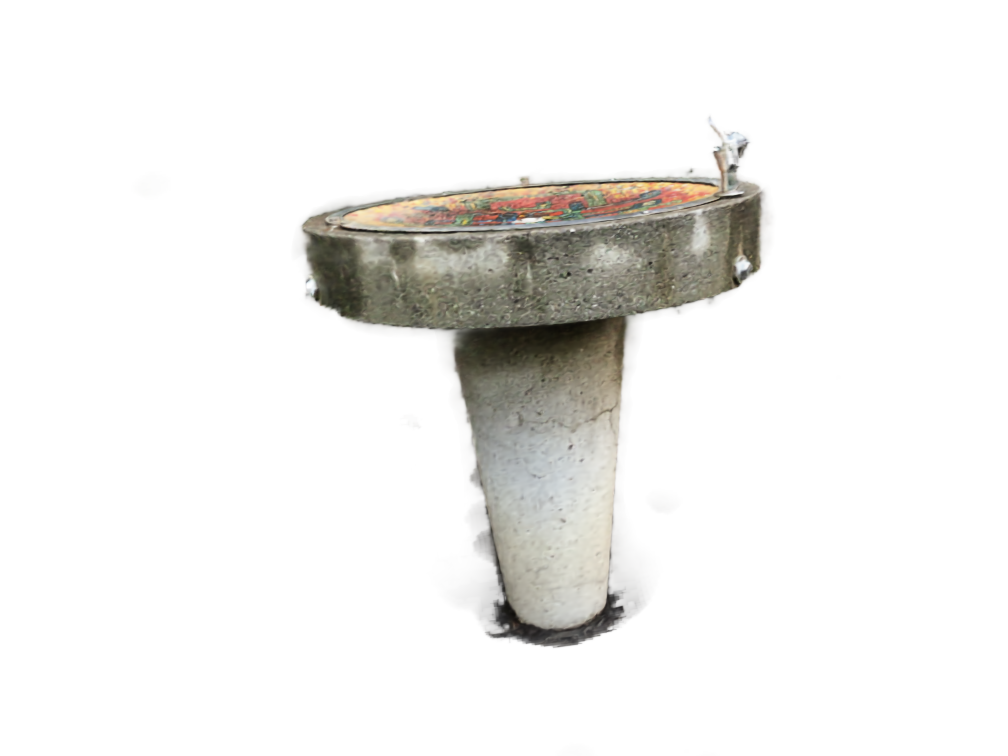}}}
        % \stackinset{l}{}{b}{}
        % {\fcolorbox{black}{orange}{\includegraphics[scale=0.65]{assets/images/cmp/styles/mediter.jpg}}}
    \end{minipage}
    \begin{minipage}{0.24\linewidth}
        \centering
         \subcaptionbox{\chesstable}{{\frame{\includegraphics[width=\textwidth]{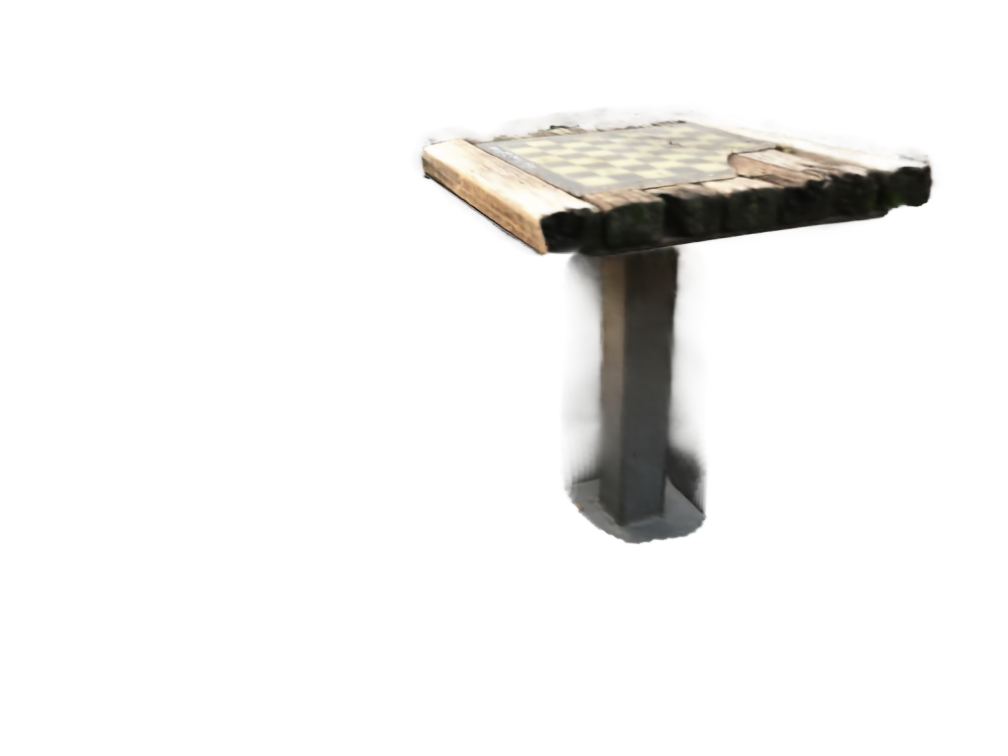}}}}
    \end{minipage}
    \begin{minipage}{0.24\linewidth}
        \centering
         \subcaptionbox{\stove}{{\frame{\includegraphics[width=\textwidth]{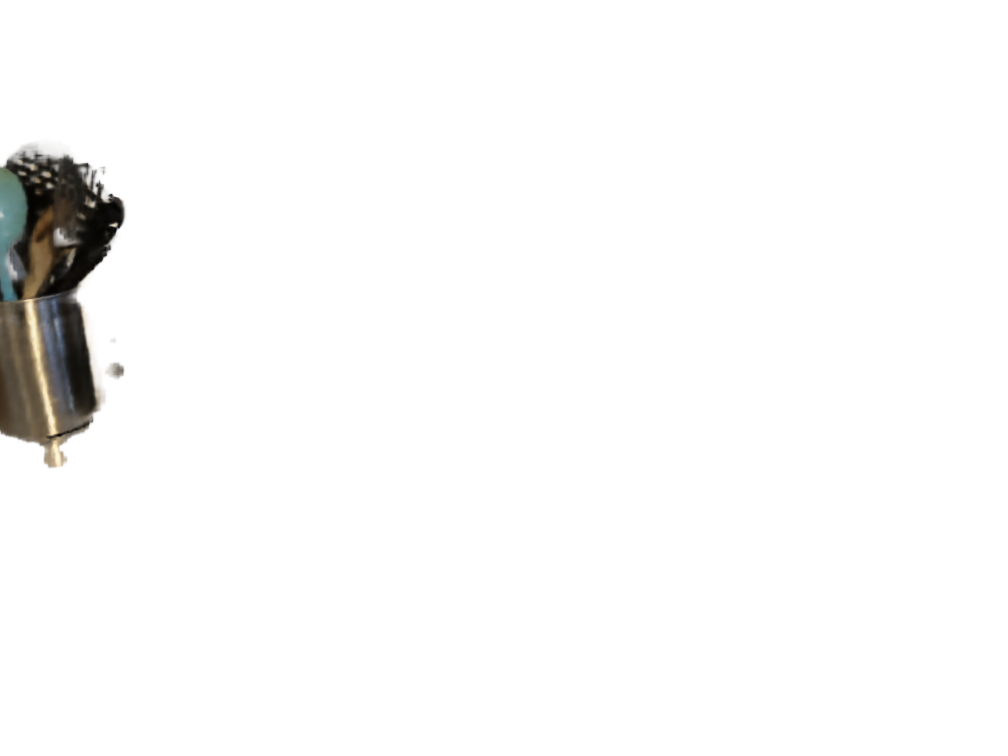}}}}
    \end{minipage}
    \begin{minipage}{0.24\linewidth}
        \centering
         \subcaptionbox{\shoerack}
         {{\frame{\includegraphics[width=\textwidth]{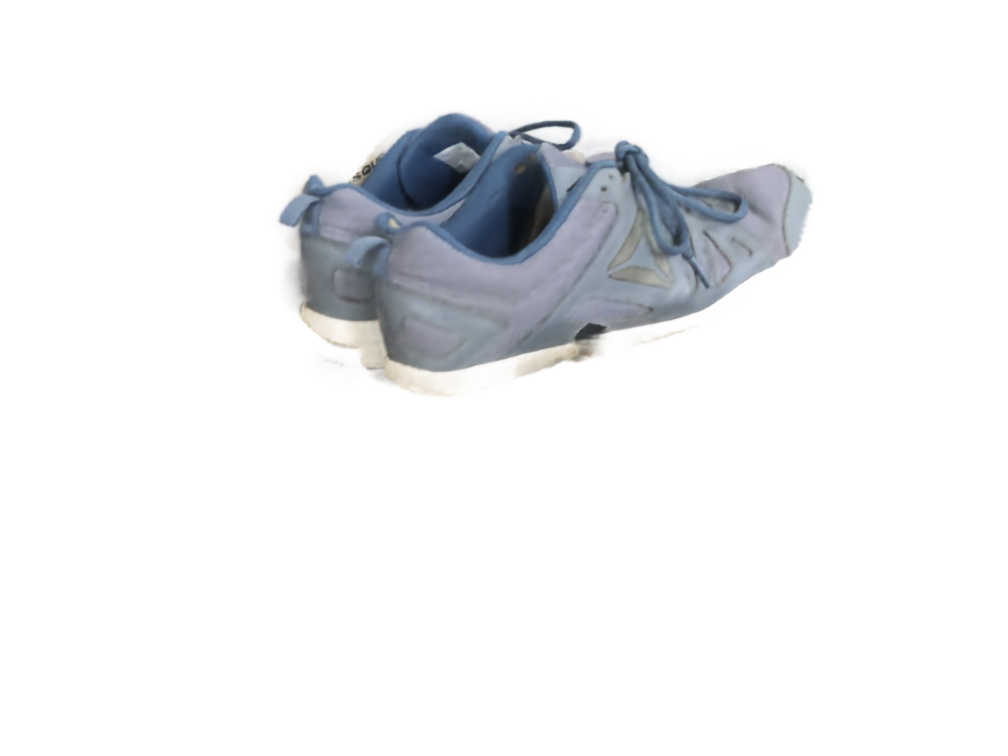}}}}
    \end{minipage}
    \caption{{\em Our ISRF vs N3F/DFF\cite{N3F,DFF}:} Both N3F and DFF employ a similar strategy for segmentation. We tweak the threshold for their method and bring out the best results and show their respective results in the Row 2.
    Row 3 shows our results with the same queried patch (highlighted in green[\sqbox{green}] in Row 1). Since our method works best on user provided strokes (shown in yellow[\sqbox{yellow}] in Row 1), we show the corresponding results in Row 4.
    While N3F/DFF are able to recover simpler objects like \colorfountain, they fail to capture other objects. Our method faithfully recovers the queried objects with clear and smooth boundaries. For more details, please refer to \cref{sub_sec:region_growing_blf}. \vspace{5mm}
    }
    \label{fig:results}
\end{figure*}
    \section{Results}
{
    \label{sec:Results}
    In this section, we discuss the comparisons and results of our proposed method against the existing semantic features distillation-based Radiance fields segmentation approaches. Specifically, we focus on the two recent approaches: DFF-DINO \cite{DFF} and N3F \cite{N3F}. Both use extracted features from input images and fuse them into the volumetric space. DFF additionally concentrates on the language queries using LSeg \cite{lseg}, but both approaches are similar regarding semantic features. As the code of DFF is not publicly available, we compare our method against N3F, which is similar to DFF for this part. %{\textcolor{red}{PJN: Don't we want to say we also compare with NVOS, which we do?}}

    \subsection{Comparison}
    {
        \label{sub_sec:comparison}
        As discussed earlier, our approach supports region selection either by a patch or a hand-drawn brush stroke as shown in \cref{fig:teaser}. To obtain the desired volumetric content, we follow the methods described in the Sections (\cref{sub_sec:feature_matching,sub_sec:region_growing_blf}). \cref{fig:results} shows our segmentation results on a few challenging scenes.
        
        The usage of clustering followed by NNFM clearly outperforms the prior approach of average matching \cite{DFF, N3F}. 
        The direct incorporation of nearest neighbor feature matching (NNFM) in these approaches leads to significant rendering delays, while the choice of neural space limits them from using elegant techniques like bilateral search. 

        In \cref{fig:results} it can be observed that in the case of the \colorfountain, the simple average feature matching technique faithfully recovers the region of interest albeit with some additional noise. However, as the scene's complexity and region of interest grows, the prior art fails to garner pleasing results. This can be observed clearly in the case of the three LLFF\cite{mildenhall2019llff} scenes (\chesstable, \shoerack, \stove).
        When only simple averaging is employed, the \chesstable scene suffers due to the erroneous feature matches. The clustered matching mitigates the errors and confines the segmented volume to the \textsc{Table}. A similar effect can be observed in the case of \stove where the object of interest is sparingly covered in the input images but is faithfully recovered with distinct boundaries, unlike N3F.
        The last scene \shoerack is a {\em classic} example where recovering white-sole might be challenging even with the best feature matching scheme. This is where the bilateral search helps in exploiting multi-domain content by conditioning on the spatio-semantics.
        
        We also qualitatively compare our results with another stroke-based approach NVOS\cite{nvos} in \cref{fig:nvos_comp}. Quantitative evaluation of \emph{mIOU/mAcc} scores on all the NVOS dataset also reflect similar behavior. Using the input strokes and GT masks of NVOS, we obtain an {\emph{mIoU} of {\bf 83.75\%} (compared to {\bf 70.1\%} of NVOS) and an \emph{mAcc}} of {\bf 96.4\%} (compared to {\bf 92.0\%} of NVOS), on the same LLFF dataset. Additionally, our interactive scheme allows for improving the segmentation in subsequent iterations. We achieved an {\em mIoU} of 90.8\% and an {\em mAcc} of 98.2\% on the same dataset using multiple strokes. A detailed depiction of results is discussed in the supplementary document.
    }
    \begin{figure*}[!ht]
    \centering
    \begin{minipage}{0.30\linewidth}
            \adjustbox{trim={0} {0} {0} {0.05\height},clip}
            {\includegraphics[width=\linewidth]{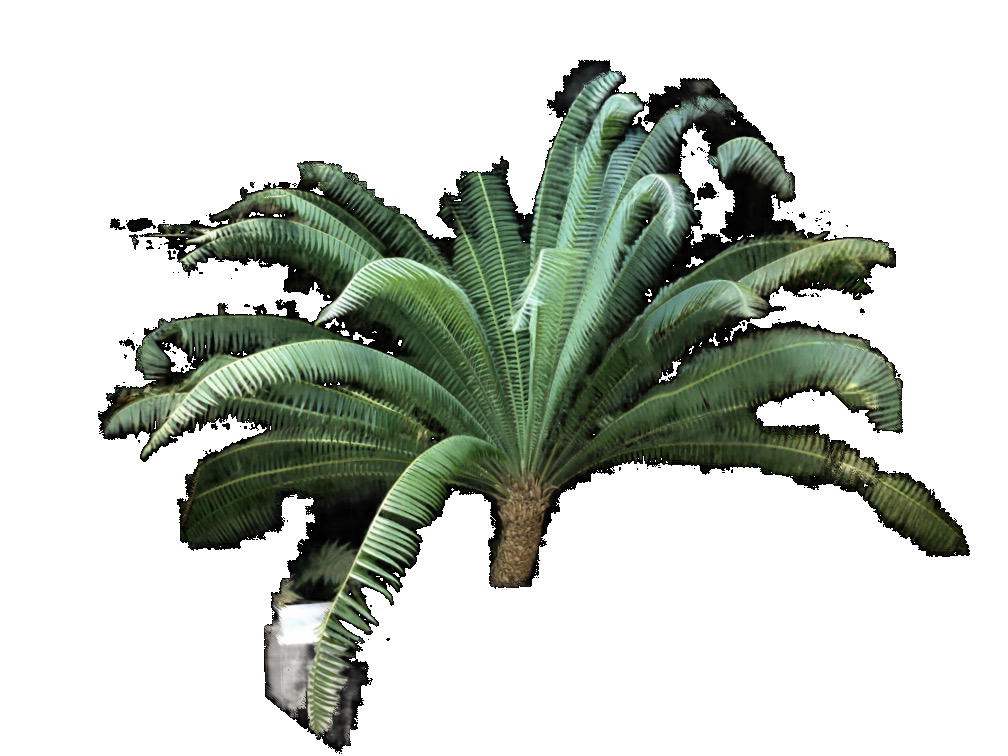}}
            \adjustbox{trim={0} {0} {0} {0.2\height},clip}
            {\includegraphics[width=\linewidth]{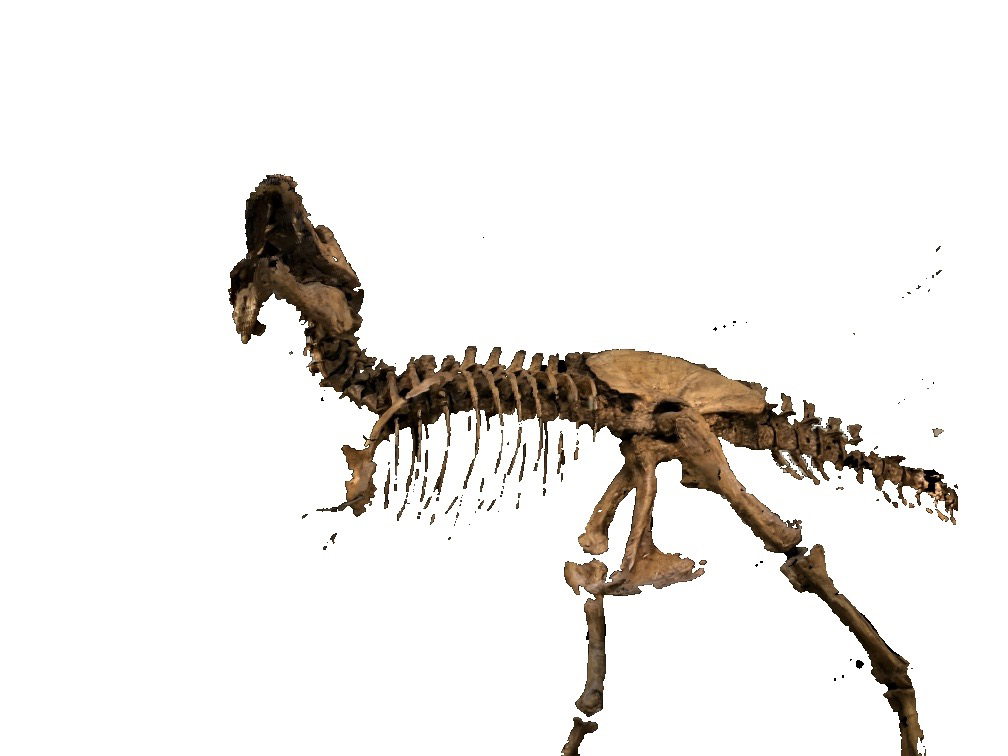}}
%            \subcaption{NVOS}
    \end{minipage}
    \begin{minipage}{0.30\linewidth}
            \adjustbox{trim={0} {0} {0} {0.05\height},clip}
            {\includegraphics[width=\linewidth]{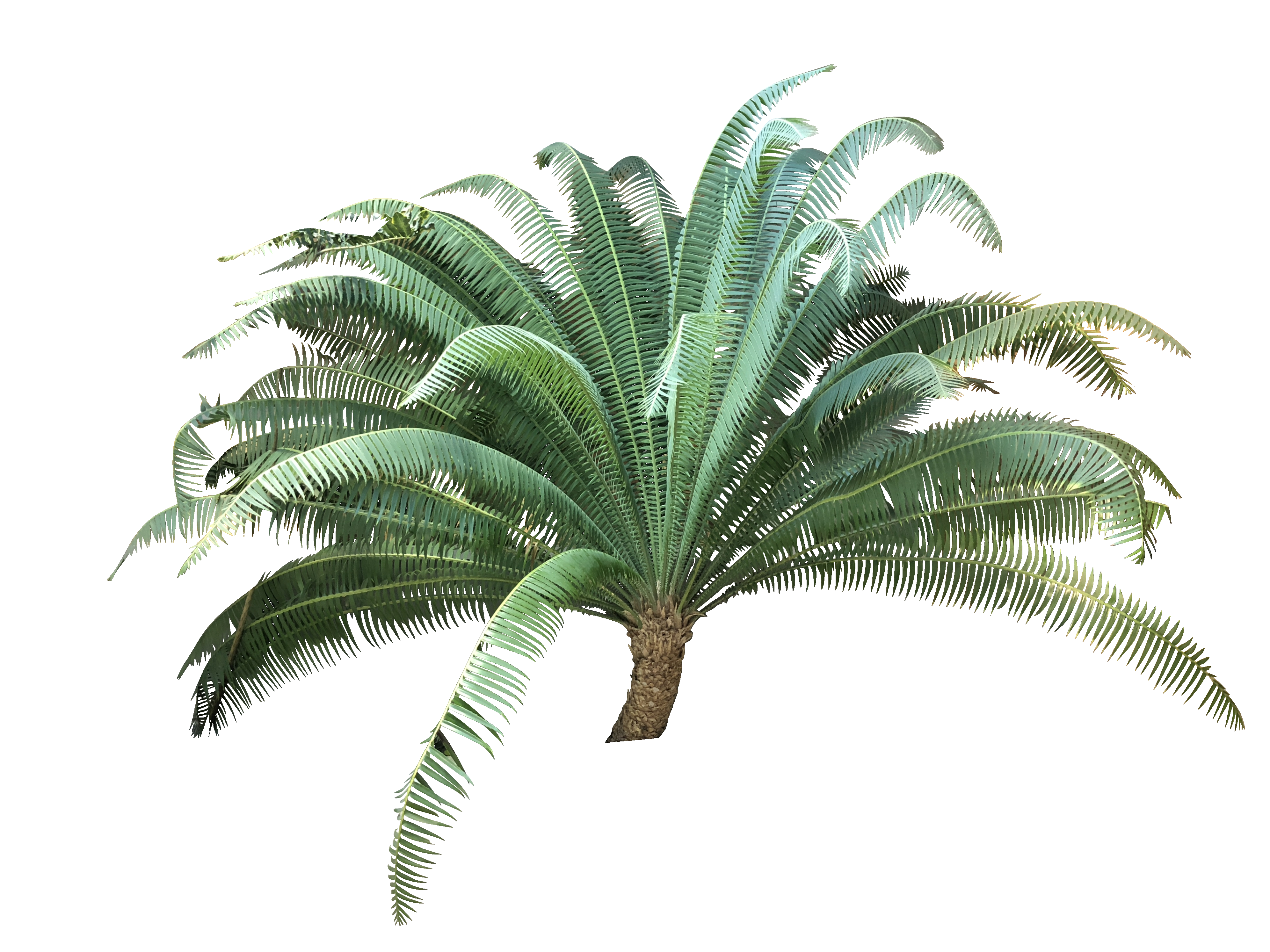}}
            \adjustbox{trim={0} {0} {0} {0.2\height},clip}
            {\includegraphics[width=\linewidth]{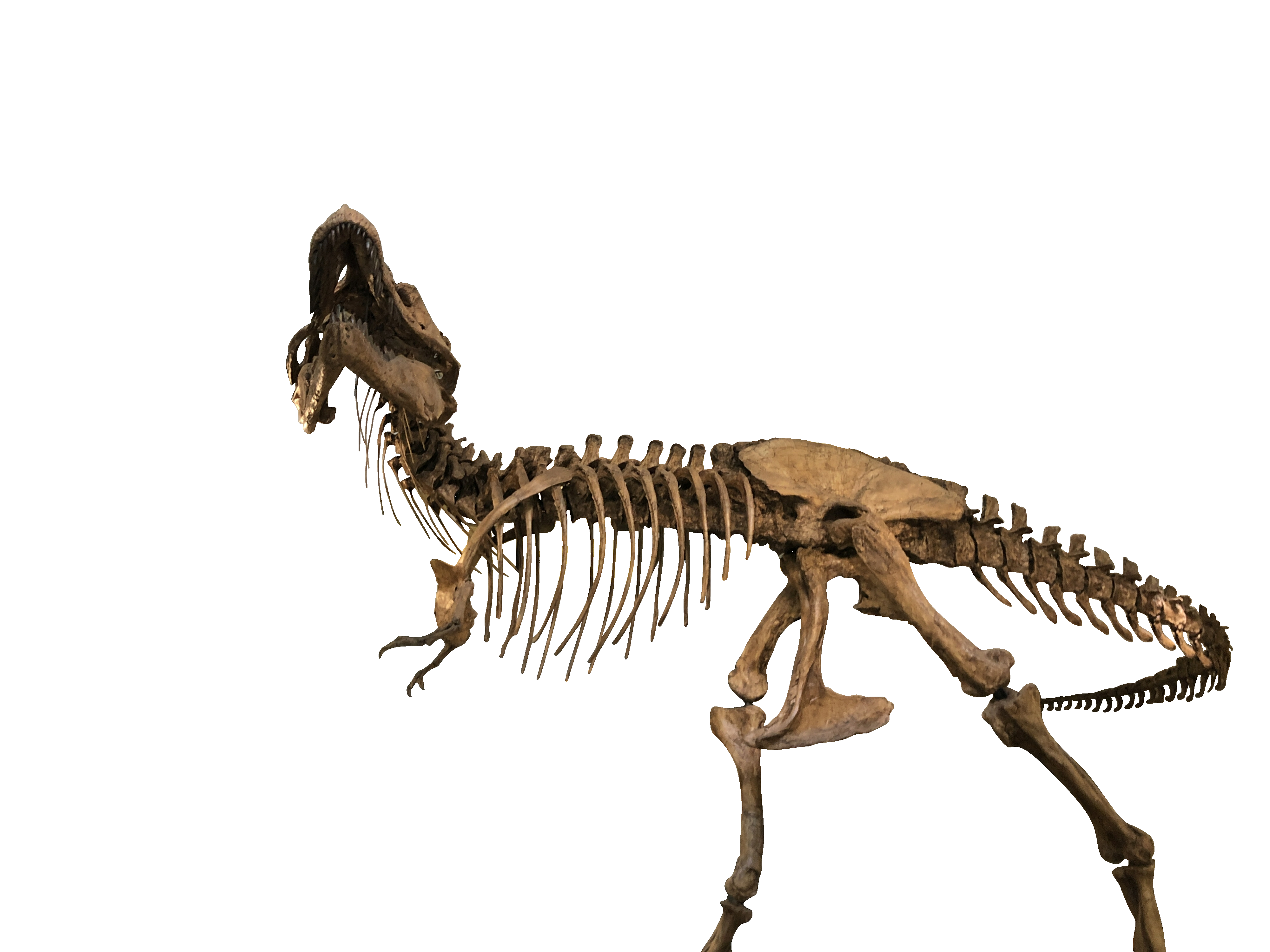}}
 %           \subcaption{Reference}
    \end{minipage}
    \begin{minipage}{0.30\linewidth}
            \adjustbox{trim={0} {0} {0} {0.05\height},clip}
            {\includegraphics[width=\linewidth]{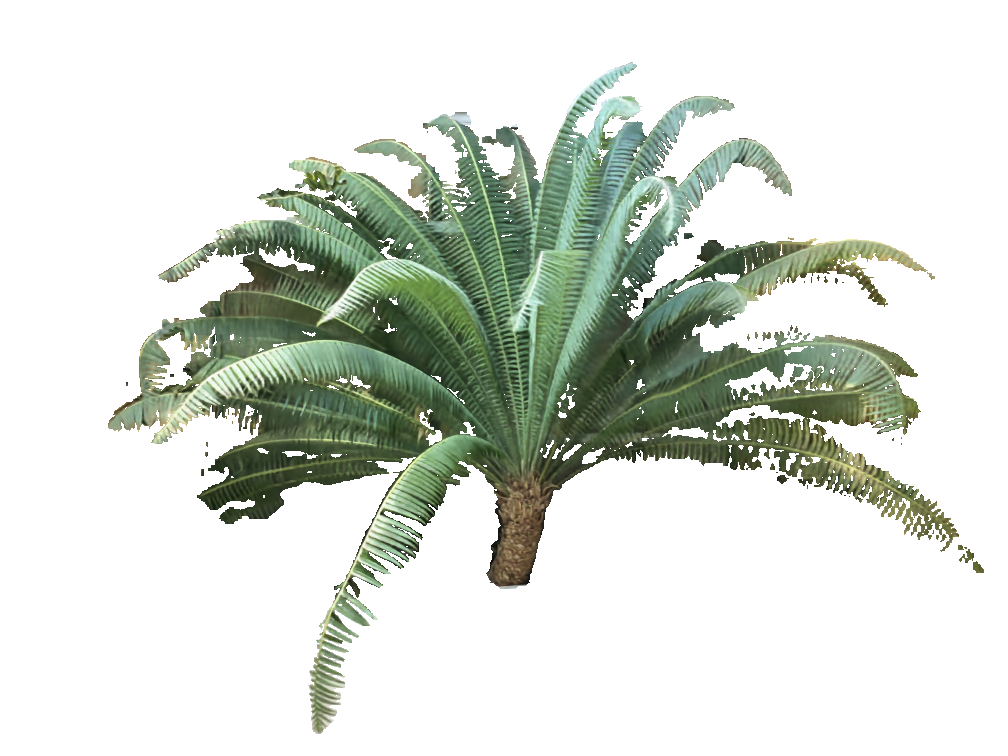}}
            \adjustbox{trim={0} {0} {0} {0.2\height},clip}
            {\includegraphics[width=\linewidth]{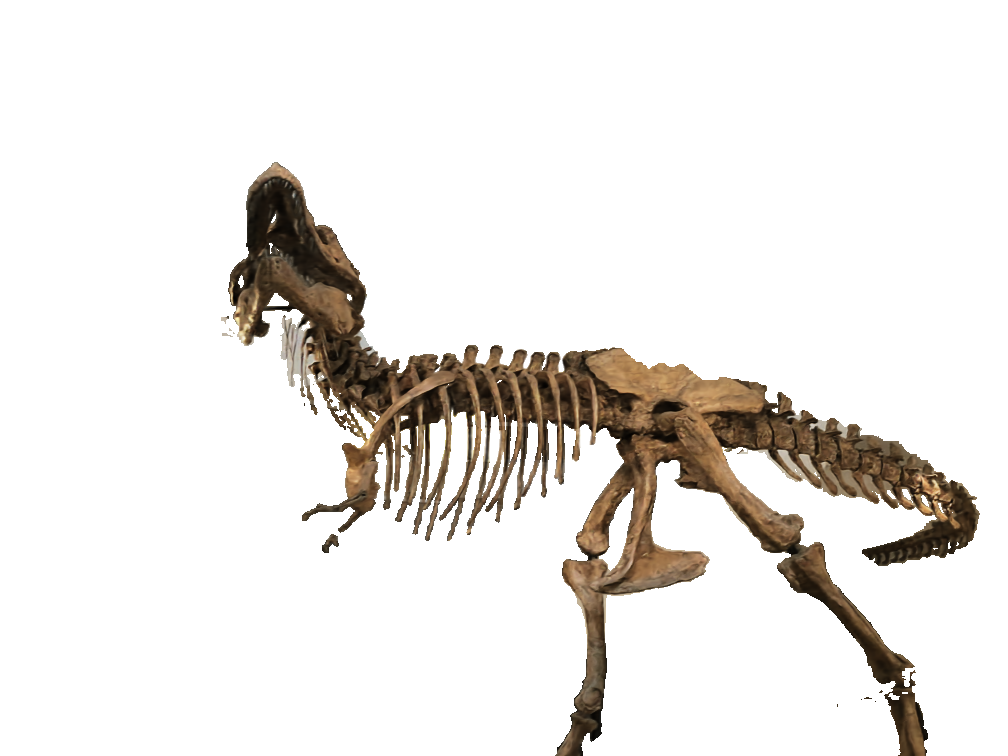}}
  %          \subcaption{ISRF}
    \end{minipage}  
    \caption{Results. Left: NVOS\cite{nvos} (from their paper), Middle: reference masks from NVOS-dataset, Right: Our ISRF system. More results can be seen in the supplementary document.
%Our \href{https://arxiv.org/abs/2212.13545}{arxiv version} provides high quality renders; please use zoom functionality of Adobe Acrobat/Okular viewer to see fine details.
    }
    \label{fig:nvos_comp}
\end{figure*}
    
    \subsection{Interactive Segmentation with User Strokes}    
    {
        \label{sub_sec:interactive_segmentation}
        Our method allows both adding and removing content using positive and negative strokes.
        The cases where the single stroke fails to obtain the desired content in the extracted space, the user can add another positive stroke to add more content. \cref{fig:double_positive} shows one such example where the excavator (`JCB') has missing teeth in the extracted region. Drawing an additional stroke and bilaterally growing the region again brings out the full desired result. This effectively grows the bit-map $M^r$ by segmenting more desirable regions from the volumetric space.

        Similar to adding new content, some scenarios demand the need to remove extraneous content from the extracted region. In such scenarios, we mark the region to be removed and grow it independently of the positive content. Once fully grown, the full extent of the negative/undesirable content is obtained which we subtract from the previously extracted regions obtaining the edited bit-map $M^r$. \cref{fig:teaser} shows one such example where the \textsc{Reflective Granite} floods into the \textsc{Table} region. We add a negative stroke (red) to remove this undesired region.

        Incorporating these functionalities is not trivial in the case of the prior art, as an additional negative match or a positive match calculation at the time of rendering is a tedious task.
    }
}

    \begin{table}[b]
    \begin{center}
        \begin{tabular}{|c|c|} 
         \hline
         \textbf{Step} & \textbf{Time Taken}\\
         \hline
         Pre-training radiance field & 7 mins \\
         \hline
         Training feature field & 2.5 mins \\
         \hline
         \hline
         K-Means Clustering & 2 secs \\
         \hline
         3D Feature Query & 1 secs \\
         \hline
         Bilateral Region Growing & 0.3 secs \\
         \hline
        \end{tabular}
    \end{center}
    \vspace{-5mm}
   \caption{\label{tab:time_taken} Timings of different steps of the ISRF pipeline}
\vspace{-5mm}
\end{table}
    \section{Experiments}
{ 
    \label{sec:experiments}
    In this section, we discuss various feature-matching variants we used to obtain the high-confidence seed region. Additionally, we show some immediate applications of radiance field segmentation.
    \subsection{Ablations}
    {
        \label{sub_sec:ablations}
        In order to obtain a high-confidence region, which acts as a seed for the bilateral filter, a feature-matching technique is required to match the marked features with the distilled semantic features in the volumetric space.
        To this end, we experimented with three different feature matching techniques, namely (1) Average Feature Matching, (2) Nearest neighbor Feature matching(NNFM) (3) K means + NNFM, which are compared in the \cref{fig:feature_matching}.
        It can be easily inferred from \cref{fig:fm_afm} that average feature matching performs poorly in this task. In order to improve these results, we resort to the nearest neighbor feature matching. Though this recovers a good high confidence region, it is accompanied by additional noise as seen in \cref{fig:fm_nnfm} Furthermore, as the marked region's size grows, computation also becomes tedious in this case. To address this, we cluster the features using K-means clustering and then do an NNFM that reduces computational overhead and avoids noisy matches, as seen in \cref{fig:fm_km_nnfm}. When $K=1$, clustering results in mean features of the selected stroke, and as $K$ approaches high values, the search approximates NNFM.
    }
    \begin{figure}[tb]
    \centering
    \begin{minipage}{0.49\linewidth}
        \centering
        \subcaptionbox{\label{fig:fm_gt}Ground Truth}
        {\includegraphics[width=\linewidth]{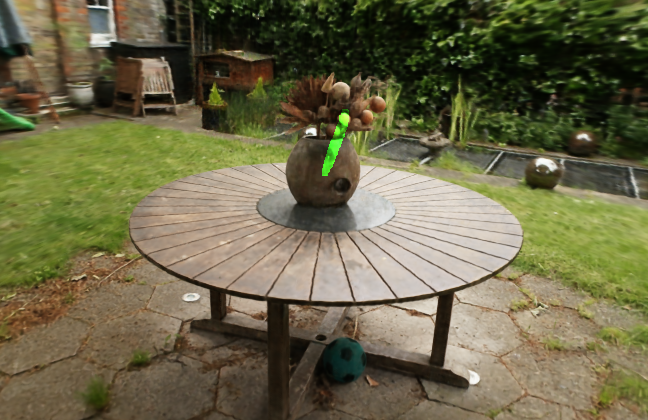}}
    \end{minipage}
    \begin{minipage}{0.49\linewidth}
        \centering
        \subcaptionbox{\label{fig:fm_afm}Average Feature Matching}
         {\includegraphics[width=\linewidth]{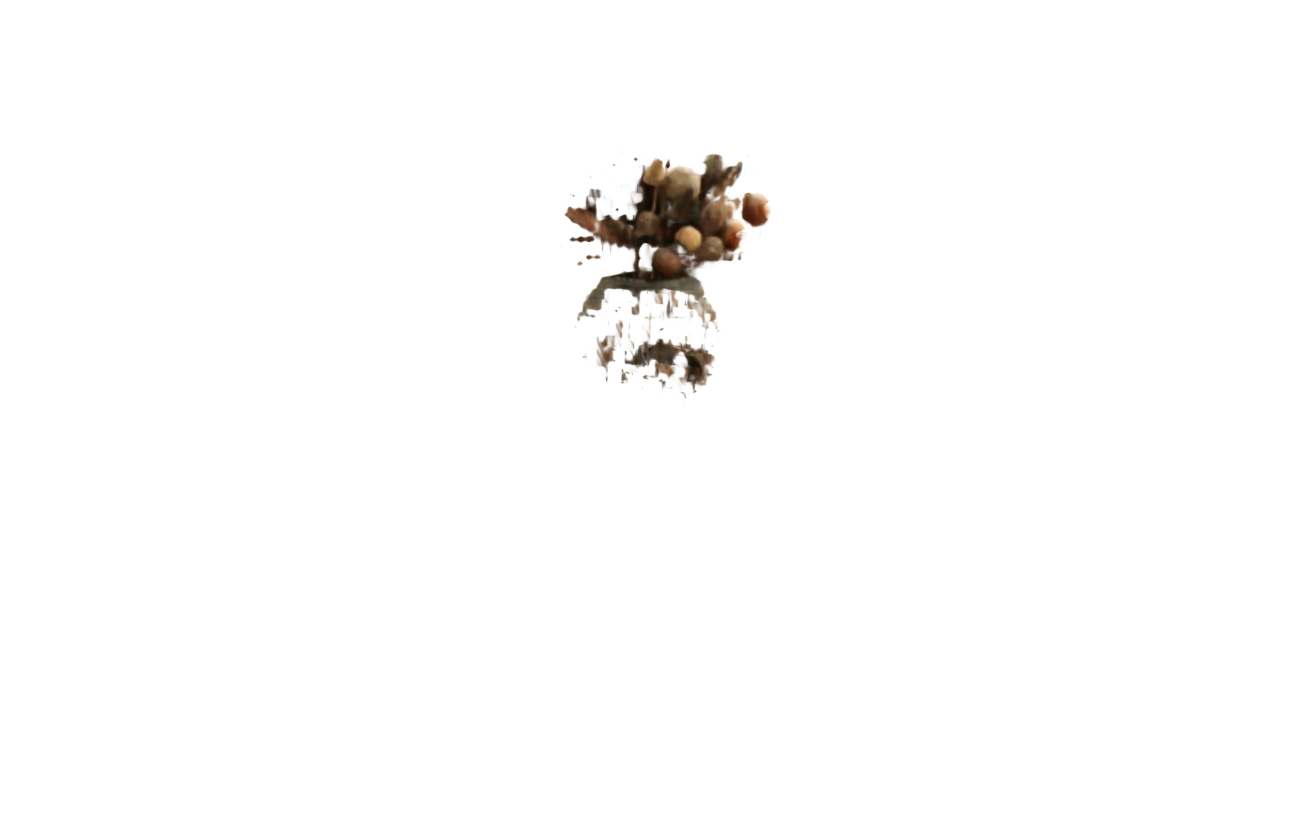}}
    \end{minipage}
    
    \begin{minipage}{0.49\linewidth}
        \centering
        \subcaptionbox{\label{fig:fm_nnfm}NNFM}
        {\includegraphics[width=1.0\linewidth]{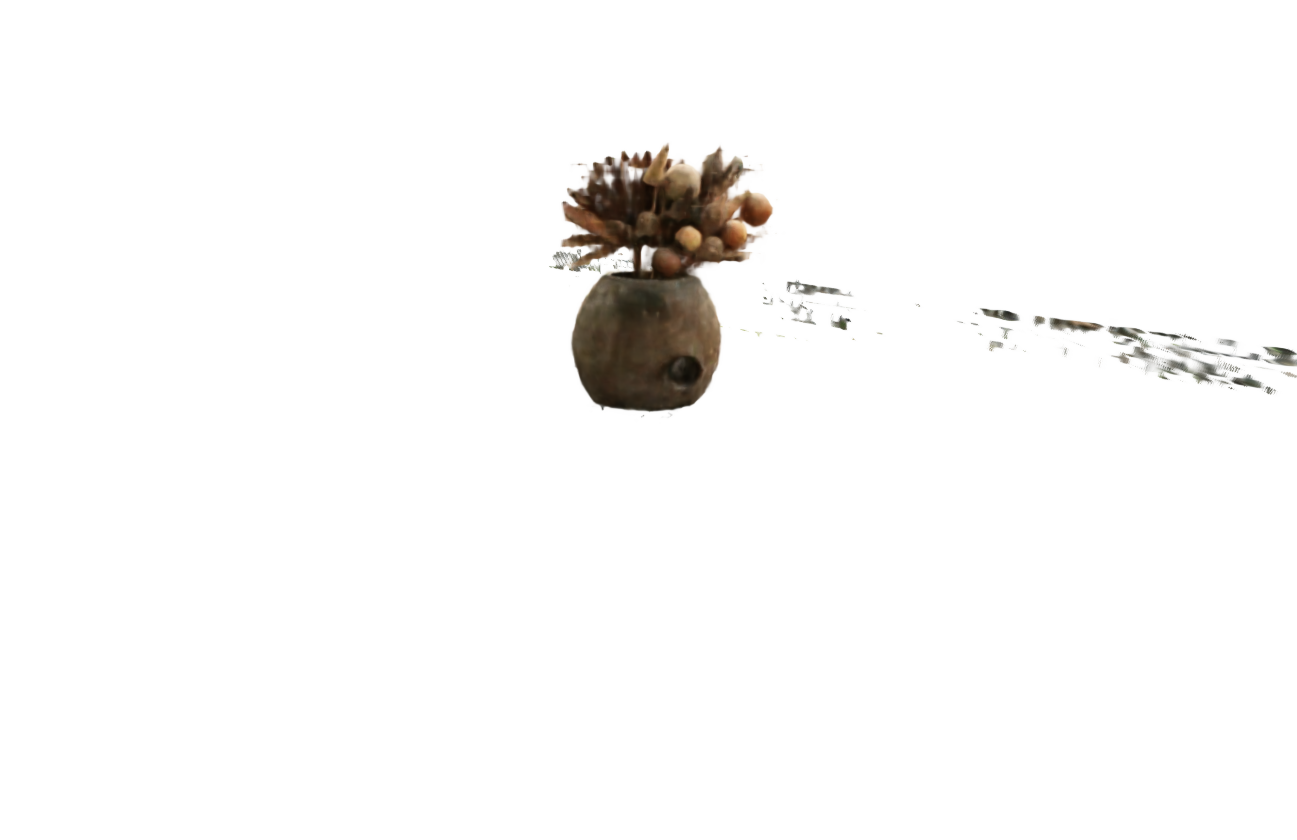}}
    \end{minipage}
    \begin{minipage}{0.49\linewidth}
        \centering
        \subcaptionbox{\label{fig:fm_km_nnfm}K-Means + NNFM}
        {\includegraphics[width=1.0\linewidth]{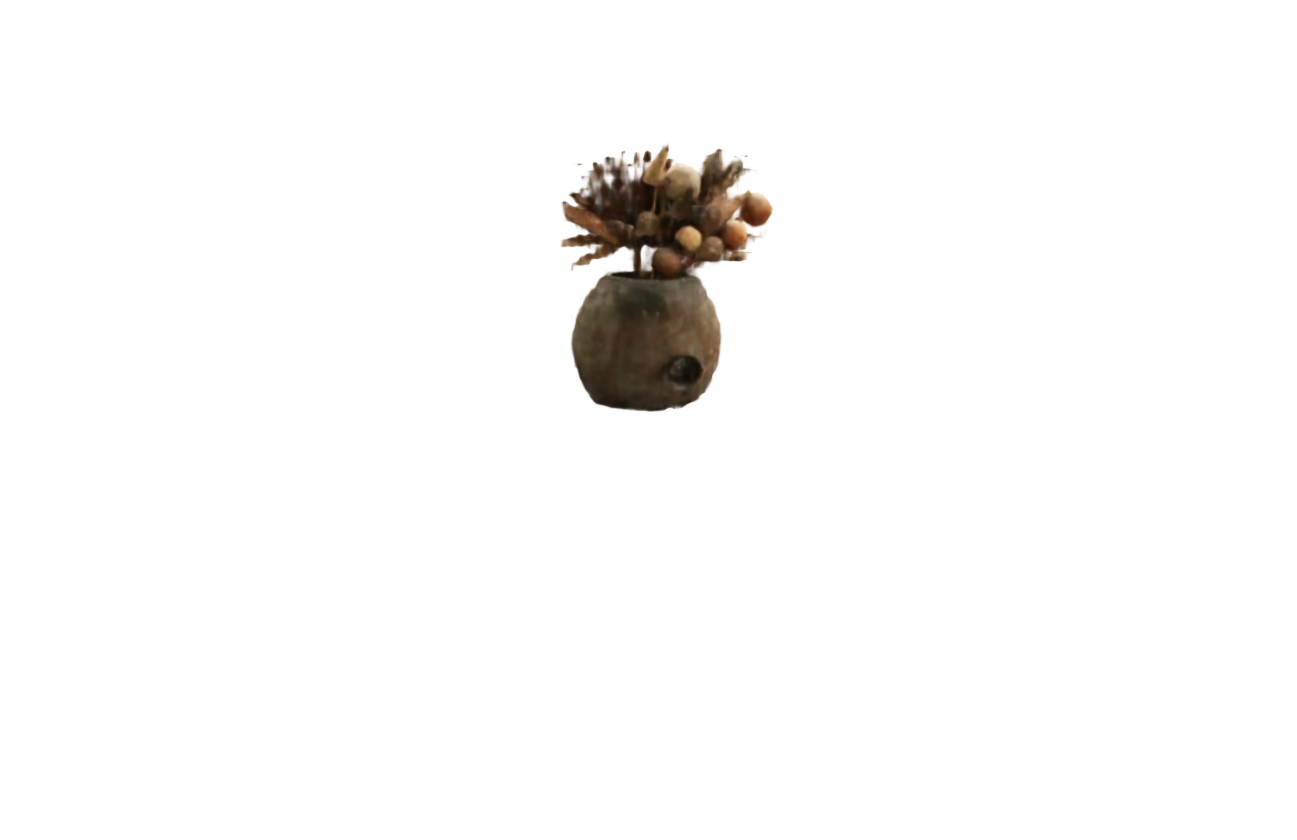}}
    \end{minipage}
    \caption{ \emph{Feature Matching}:
    {This figure shows the high confidence region of the RF (a) obtained using different feature-matching techniques for a particular stroke. While {\em Average feature matching} (b) fails to cover the entire object due to loss of information during the averaging process, {\em NNFM} (c) without clustering leads to noise bleeding. Use of {\em NNFM after clustering} (d) eliminates noisy regions while also considering multiple features at once.}}
    \label{fig:feature_matching}
    \vspace{-5mm}
\end{figure}
    
    \subsection{Editing}
    {
        \label{sub_sec:editing}
        After obtaining a good segmentation, many appealing opportunities for editing open up:

        \Paragraph{Object Removal: }The removal of an object from the scene is a simple task and is shown in \cref{fig:editing_removal} where the \textsc{Pot} in \garden scene is removed. Please note that we do not inpaint the scene post content removal.

        \Paragraph{Affine transformation: } As we have good quality segmented volumetric content, we can perform affine transformations on voxel space (\textsc{Pot}) for object position manipulation. We demonstrate this in \cref{fig:editing_trans}. One can look \textit{behind} the \textsc{Table} on the \textsc{Ground} to see the repositioned \textsc{Pot}.
        
        \Paragraph{Geometric Scene Composition: } With high-quality 3D segmentation masks, we can also composite two different radiance fields. 
        We demonstrate this in \cref{fig:editing_composition}. We follow the composition technique of \cite{d2nerf} to accomplish this task. The JCB is picked from the \kitchen scene from the \cite{mipnerf360} dataset and placed in \garden. Please note that we do not take global illumination into account for these edits.

        \Paragraph{Appearance Editing: } As the appearance vector is associated with each voxel in the grid, we can alter the appearance of the individual segmented objects. We demonstrate this by stylization the content using \cite{goel2022styletrf} in \cref{fig:editing_style}.
        \begin{figure}[b]
    \centering    
    \begin{minipage}{0.49\linewidth}
        \centering
        \subcaptionbox{\label{fig:editing_removal}Removal}{{
        \begin{tikzpicture}
            \node[anchor=south west,inner sep=0] (image) at (0,0) {\includegraphics[width=\textwidth]{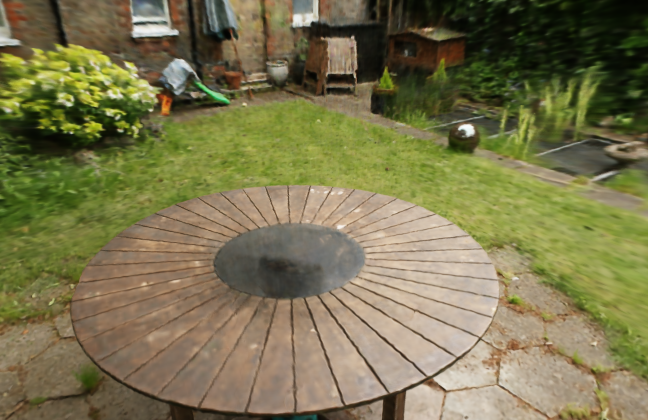}};
        \end{tikzpicture}
        }}
    \end{minipage}
    \begin{minipage}{0.49\linewidth}
        \centering
        \subcaptionbox{\label{fig:editing_trans}Translation}{{
        \begin{tikzpicture}
            \node[anchor=south west,inner sep=0] (image) at (0,0) {\includegraphics[width=\textwidth]{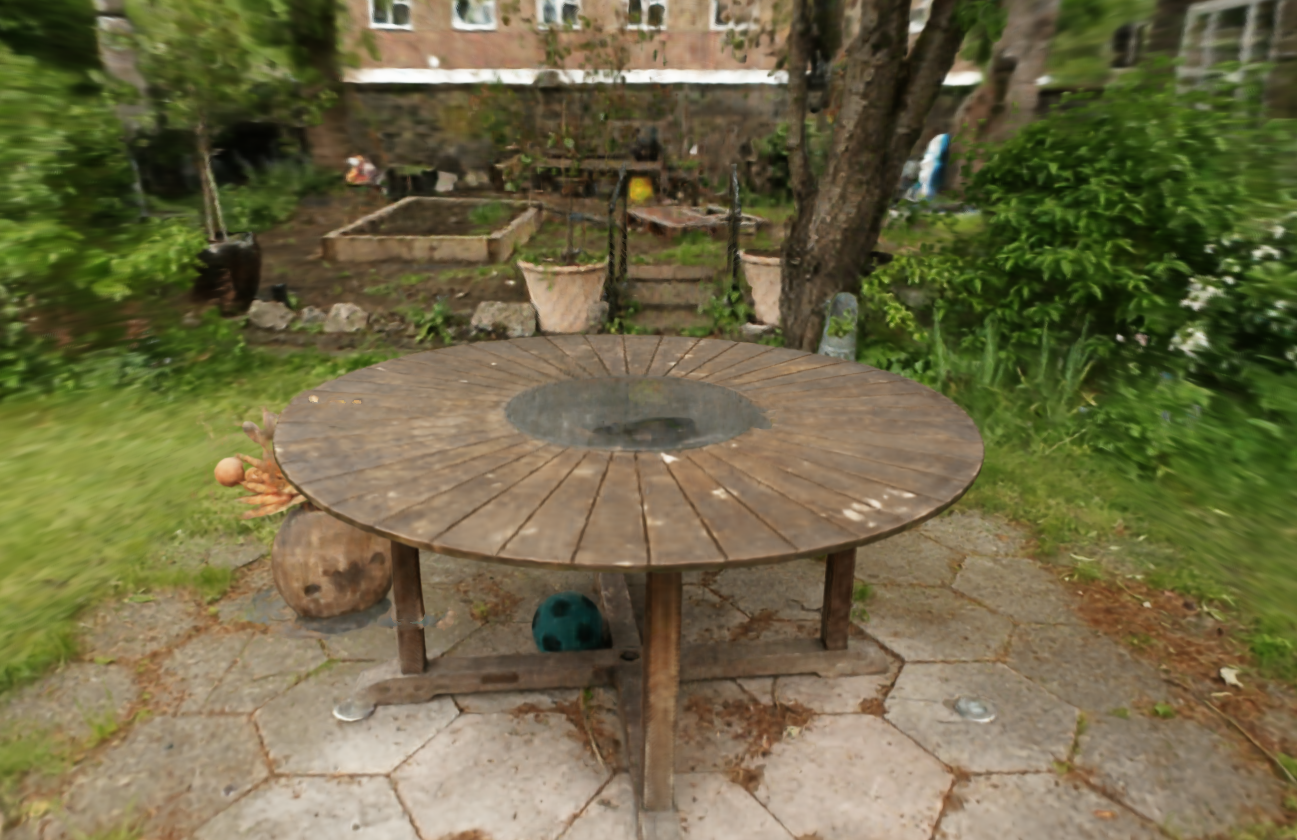}};
            \begin{scope}[x={(image.south east)},y={(image.north west)}]
                \draw[cyan,ultra thick] (0.15,0.2) rectangle (0.4,0.6);
            \end{scope}
        \end{tikzpicture}
        }}
    \end{minipage}
    \begin{minipage}{0.49\linewidth}
        \centering
        \subcaptionbox{\label{fig:editing_composition}Scene Composition}{{
        \begin{tikzpicture}
            \node[anchor=south west,inner sep=0] (image) at (0,0) {\includegraphics[width=\textwidth]{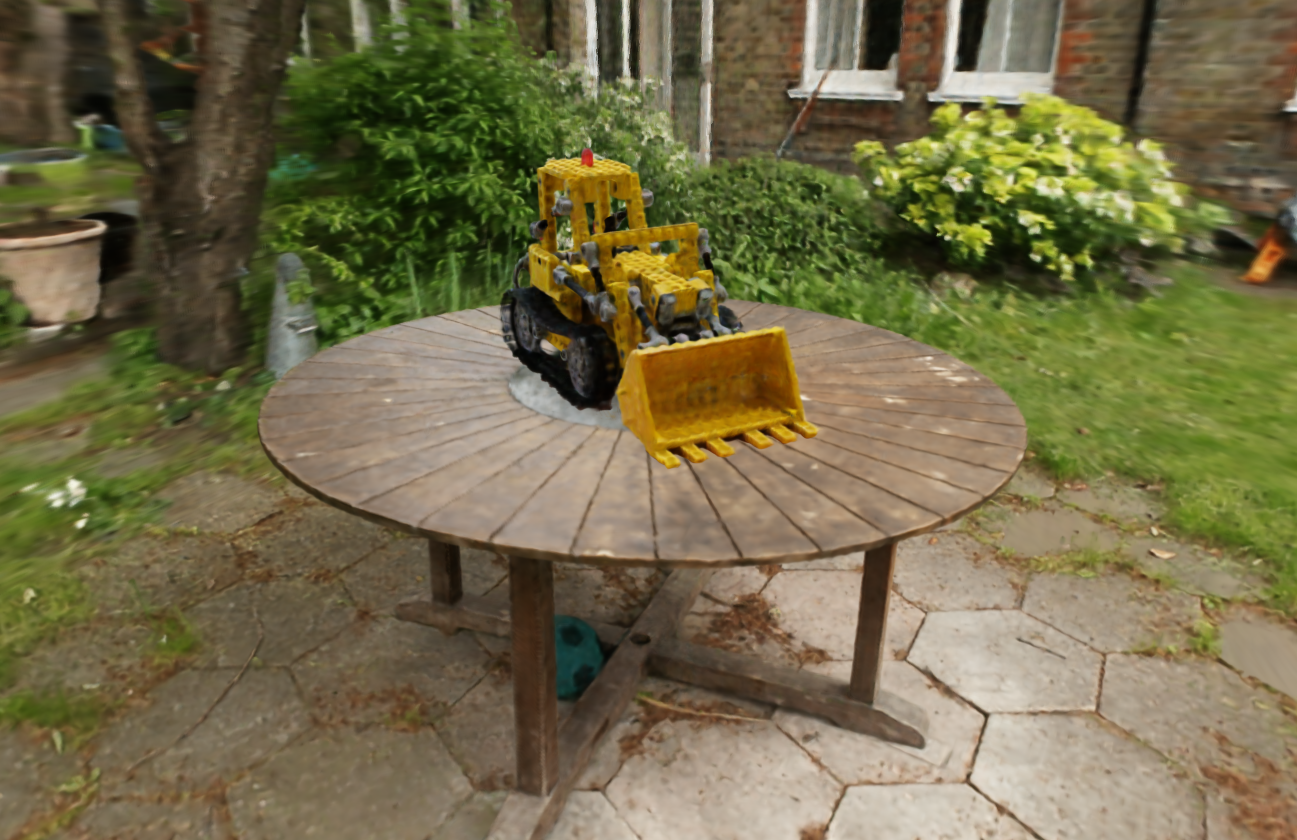}};
        \end{tikzpicture}
        }}
    \end{minipage}
    \begin{minipage}{0.49\linewidth}
        \centering
        \subcaptionbox{\label{fig:editing_style}Appearance Editing}{{
        \begin{tikzpicture}
            \node[anchor=south west,inner sep=0] (image) at (0,0) {\includegraphics[width=\textwidth]{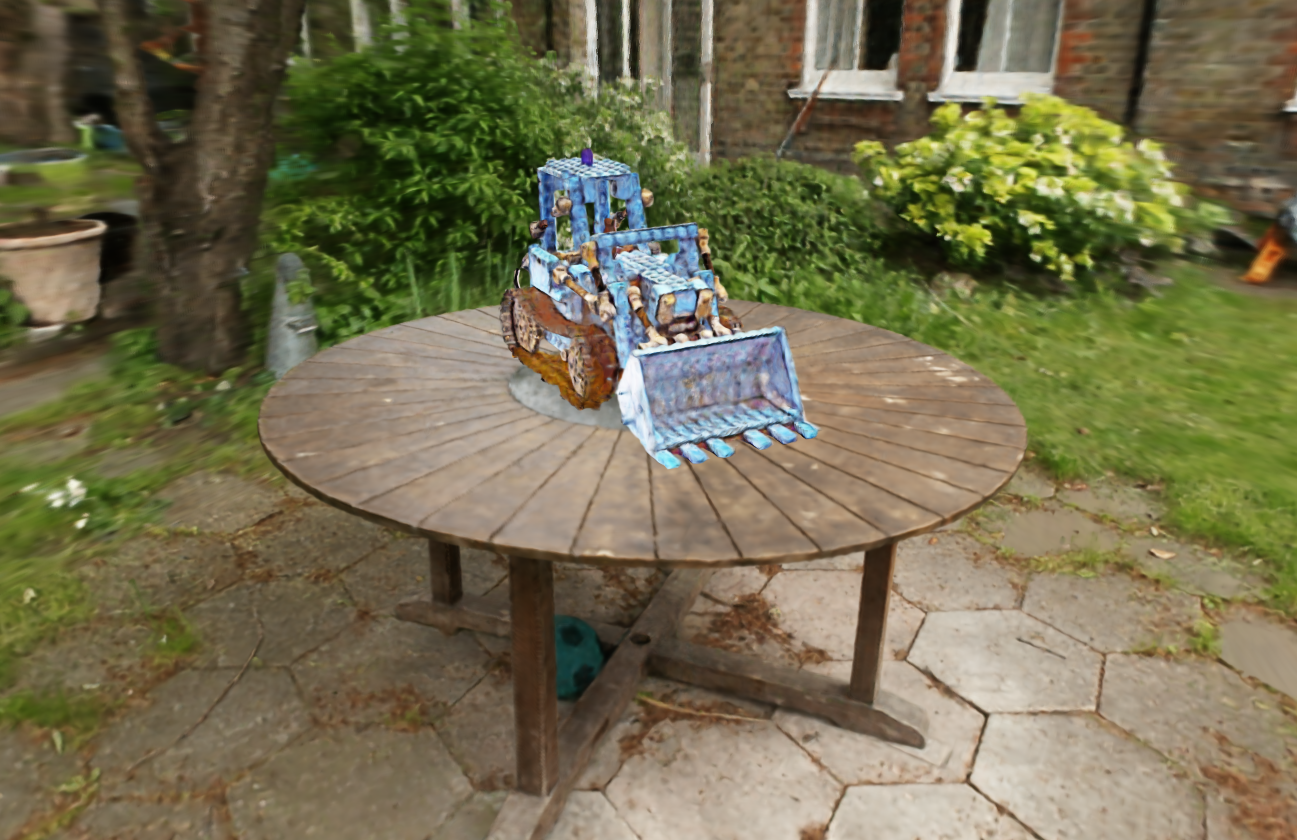}};
        \end{tikzpicture}
        }}
    \end{minipage}
    \caption{ \emph{Scene manipulation}:
    {Segmented object(s) can be edited in different ways. In (a), we remove the \textsc{Pot} from the center of the table. In (b), we translate the \textsc{Pot} to the \textsc{Ground} behind the \textsc{Table}. In (c), we replace the \textsc{Pot} with the \textsc{lego JCB} obtained from a different scene (\kitchen). We stylize the newly added \textsc{lego JCB} using  \cite{goel2022styletrf} in (d). All scenes are from \cite{mipnerf360}.}}
    \label{fig:editing}
    \vspace{-4mm}
\end{figure}

    }
    \subsection{Discussions and Limitations}
    {
        Our method improves upon the prior art on several fronts but has its own shortcomings. Like prior works, we rely on DINO features to represent object semantics and this can result in artefacts if the features do not capture the semantics properly. Third column in the last row in \cref{fig:results} shows a small false appendage at the bottom of the utensil holder which can not be easily removed interactively without eating into object's body. Better semantic features can resolve this problem. Also, the leftmost example in \cref{fig:editing} shows that the shadow of the pot is left behind on the granite center of the table even after the pot is edited out. Removing the pot from the geometric representation does not guarantee removal of its secondary effects on neighbouring objects like shadows or highlights, without elaborate geometric post-processing. Our method may also struggle in segmenting geometry well if the voxel resolution is low compared to the scale of object details as shown in  supplementary results. Multiresolution voxel representations can solve this problem with additional overhead.
    }
}
    \section{Conclusions and Future Work}
{
    \label{sec:conclusions}
    \noindent
    In this paper, we presented an easy and accurate method to segment objects from a TensoRF representation of radiance fields and showed simple scene editing operations facilitated by this. The efficient voxel-based representation we use makes our method more versatile and simple compared to the prior works in this direction. We show several results on multiple challenging scenes (and present more in the supplementary document). Semantic segmentation is a first step towards interpretation, understanding, and manipulation of 3D scenes. This work provides high quality segmentation that can be the basis for several such downstream tasks. { A simple extension to the current method would be to generalize the distance used for matching in the NNFM and region-growing steps to include other features like color latent vectors. Extending the current method to a InstantNGP\cite{instantngp} framework, while incorporating additional multi-domain explorations strategies like guided filtering\cite{guided_filtering} would be a good direction to explore.}
    
    In the future, multi-representation processing might be needed by combining parts of captured RFs, graphics models, SDFs, etc., to provide maximum flexibility in Virtual Reality and Augmented Reality applications. This requires processing parts of the RFs directly without going through the full learning process post-editing. This is a promising direction of work that we intend to pursue in the future.\\
}

    % \acknowledgement
    {\noindent {\bf Potential negative societal impact:} Our work presents a tool to manipulate radiance fields captured casually. While ill-intentioned manipulation to create the appearance to fake scene content is possible using such a tool, the risk is negligible compared to the sophisticated image or geometry editing tools that are already prevalent. Our method needs very little additional data and doesn't directly use vast internet collections with or without consent.}\\

    {
%        \vspace{-5mm}
        \noindent{\bf Acknowledgements:} We thank {Rajvi Shah} and {Parikshit Sakurikar} for their valuable inputs and \href{http://ishaanshah.github.io/}{Ishaan Shah} for his effort in creation of object masks used for evaluation of quantitative metrics. We thank the TCS Foundation and the Department of Science and Technology for partially funding the students.
    }
    
    % \input{src/tables/0_table_consistency_metrics}
    % \input{src/tables/1_table_time}
    % \input{src/sections/selfNotes.tex}

    %%%%%%%%% REFERENCES
    {\small
    \bibliographystyle{ieee_fullname}
    \bibliography{egbib}

\begin{thebibliography}{10}\itemsep=-1pt

\bibitem{lightfieldNerf}
Benjamin Attal, Jia-Bin Huang, Michael Zollh{\"o}fer, Johannes Kopf, and
  Changil Kim.
\newblock {Learning Neural Light Fields with Ray-Space Embedding Networks}.
\newblock In {\em Proceedings of the IEEE/CVF Conference on Computer Vision and
  Pattern Recognition (CVPR)}, 2022.

\bibitem{mipnerf360}
Jonathan~T. Barron, Ben Mildenhall, Dor Verbin, Pratul~P. Srinivasan, and Peter
  Hedman.
\newblock {Mip-NeRF 360: Unbounded Anti-Aliased Neural Radiance Fields}.
\newblock In {\em Proceedings of the IEEE/CVF Conference on Computer Vision and
  Pattern Recognition (CVPR)}, 2022.

\bibitem{NeRD}
Mark Boss, Raphael Braun, Varun Jampani, Jonathan~T. Barron, Ce Liu, and
  Hendrik~P.A. Lensch.
\newblock {NeRD: Neural Reflectance Decomposition from Image Collections}.
\newblock In {\em Proceedings of the IEEE/CVF International Conference on
  Computer Vision (ICCV)}, 2021.

\bibitem{neural_pil}
Mark Boss, Varun Jampani, Raphael Braun, Ce Liu, Jonathan~T. Barron, and
  Hendrik~P.A. Lensch.
\newblock {Neural-PIL: Neural Pre-Integrated Lighting for Reflectance
  Decomposition}.
\newblock In {\em Adv. Neural Inform. Process. Syst.}, 2021.

\bibitem{nerf2real}
Arunkumar Byravan, Jan Humplik, Leonard Hasenclever, Arthur Brussee, Francesco
  Nori, Tuomas Haarnoja, Ben Moran, Steven Bohez, Fereshteh Sadeghi, Bojan
  Vujatovic, and Nicolas Heess.
\newblock {NeRF2Real: Sim2real Transfer of Vision-guided Bipedal Motion Skills
  using Neural Radiance Fields}.
\newblock 2022.

\bibitem{dino}
Mathilde Caron, Hugo Touvron, Ishan Misra, Herv\'e J\'egou, Julien Mairal,
  Piotr Bojanowski, and Armand Joulin.
\newblock {Emerging Properties in Self-Supervised Vision Transformers}.
\newblock In {\em Proceedings of the IEEE/CVF International Conference on
  Computer Vision (ICCV)}, 2021.

\bibitem{tensorf}
Anpei Chen, Zexiang Xu, Andreas Geiger, Jingyi Yu, and Hao Su.
\newblock {TensoRF: Tensorial Radiance Fields}.
\newblock In {\em Proceedings of the European Conference on Computer Vision
  (ECCV)}, 2022.

\bibitem{neumesh}
{Chong Bao and Bangbang Yang}, Zeng Junyi, Bao Hujun, Zhang Yinda, Cui
  Zhaopeng, and Zhang Guofeng.
\newblock {NeuMesh: Learning Disentangled Neural Mesh-based Implicit Field for
  Geometry and Texture Editing}.
\newblock In {\em Proceedings of the European Conference on Computer Vision
  (ECCV)}, 2022.

\bibitem{pandora}
Akshat Dave, Yongyi Zhao, and Ashok Veeraraghavan.
\newblock {PANDORA: Polarization-Aided Neural Decomposition Of Radiance}.
\newblock In {\em Proceedings of the European Conference on Computer Vision
  (ECCV)}, 2022.

\bibitem{plenoxels}
{Fridovich-Keil and Yu}, Matthew Tancik, Qinhong Chen, Benjamin Recht, and
  Angjoo Kanazawa.
\newblock {Plenoxels: Radiance Fields without Neural Networks}.
\newblock In {\em Proceedings of the IEEE/CVF Conference on Computer Vision and
  Pattern Recognition (CVPR)}, 2022.

\bibitem{nerf_semantic_1}
Xiao Fu, Shang-Wei Zhang, Tianrun Chen, Yichong Lu, Lanyun Zhu, Xiaowei Zhou,
  Andreas Geiger, and Yiyi Liao.
\newblock {Panoptic NeRF: 3D-to-2D Label Transfer for Panoptic Urban Scene
  Segmentation}.
\newblock In {\em International Conference on 3D Vision (3DV)}, 2022.

\bibitem{nerf_semantic_3}
Xuan Gao, Chenglai Zhong, Jun Xiang, Yang Hong, Yudong Guo, and Juyong Zhang.
\newblock {Reconstructing Personalized Semantic Facial NeRF Models from
  Monocular Video}.
\newblock {\em ACM Trans. Graph.}, 2022.

\bibitem{gatys}
Leon~A. Gatys, Alexander~S. Ecker, and Matthias Bethge.
\newblock {Image Style Transfer Using Convolutional Neural Networks}.
\newblock In {\em Proceedings of the IEEE/CVF Conference on Computer Vision and
  Pattern Recognition (CVPR)}, 2016.

\bibitem{goel2022styletrf}
Rahul Goel, Dhawal Sirikonda, Saurabh Saini, and P.~J. Narayanan.
\newblock {StyleTRF: Stylizing Tensorial Radiance Fields}.
\newblock In {\em Proceedings of the Thirteenth Indian Conference on Computer
  Vision, Graphics and Image Processing}, ICVGIP '22, 2022.

\bibitem{lumigraph}
Steven~J. Gortler, Radek Grzeszczuk, Richard Szeliski, and Michael~F. Cohen.
\newblock {The Lumigraph}.
\newblock In {\em Proceedings of the 23rd Annual Conference on Computer
  Graphics and Interactive Techniques}, SIGGRAPH '96, 1996.

\bibitem{nerf_audio}
Yudong Guo, Keyu Chen, Sen Liang, Yongjin Liu, Hujun Bao, and Juyong Zhang.
\newblock {AD-NeRF: Audio Driven Neural Radiance Fields for Talking Head
  Synthesis}.
\newblock 2021.

\bibitem{guided_filtering}
Kaiming He, Jian Sun, and Xiaoou Tang.
\newblock {Guided Image Filtering}.
\newblock {\em IEEE Trans. Pattern Anal. Mach. Intell. (T-PAMI)}, 2013.

\bibitem{stylizednerf_cvpr}
Yi-Hua Huang, Yue He, Yu-Jie Yuan, Yu-Kun Lai, and Lin Gao.
\newblock {StylizedNeRF: Consistent 3D Scene Stylization as Stylized NeRF via
  2D-3D Mutual Learning}.
\newblock In {\em Proceedings of the IEEE/CVF Conference on Computer Vision and
  Pattern Recognition (CVPR)}, 2022.

\bibitem{nerf_semantic_2}
Ajay Jain, Matthew Tancik, and P. Abbeel.
\newblock {Putting NeRF on a Diet: Semantically Consistent Few-Shot View
  Synthesis}.
\newblock In {\em Proceedings of the IEEE/CVF International Conference on
  Computer Vision (ICCV)}, 2021.

\bibitem{johnson}
Justin Johnson, Alexandre Alahi, and Li Fei-Fei.
\newblock {Perceptual Losses for Real-Time Style Transfer and
  Super-Resolution}.
\newblock In {\em Proceedings of the European Conference on Computer Vision
  (ECCV)}, 2016.

\bibitem{DFF}
Sosuke Kobayashi, Eiichi Matsumoto, and Vincent Sitzmann.
\newblock {Decomposing NeRF for Editing via Feature Field Distillation}.
\newblock In {\em Adv. Neural Inform. Process. Syst.}, 2022.

\bibitem{space_carving}
K.N. Kutulakos and S.M. Seitz.
\newblock A theory of shape by space carving.
\newblock In {\em {Proceedings of the Seventh IEEE International Conference on
  Computer Vision}}, 1999.

\bibitem{plenoptic}
Marc Levoy and Pat Hanrahan.
\newblock {Light field rendering}.
\newblock {\em Proceedings of the 23rd annual conference on Computer graphics
  and interactive techniques}, 1996.

\bibitem{lseg}
Boyi Li, Kilian~Q Weinberger, Serge Belongie, Vladlen Koltun, and Rene Ranftl.
\newblock {Language-driven Semantic Segmentation}.
\newblock In {\em Int. Conf. Learn. Represent.}, 2022.

\bibitem{mask_dino}
Feng Li, Hao Zhang, Huaizhe xu, Shilong Liu, Lei Zhang, Lionel~M. Ni, and
  Heung-Yeung Shum.
\newblock {Mask DINO: Towards A Unified Transformer-based Framework for Object
  Detection and Segmentation}.
\newblock {\em arXiv, abs:2206.02777}, 2022.

\bibitem{loc_nerf}
Dominic Maggio, Marcus Abate, J. Shi, Courtney Mario, and Luca Carlone.
\newblock {Loc-NeRF: Monte Carlo Localization using Neural Radiance Fields}.
\newblock {\em ArXiv}, abs/2209.09050, 2022.

\bibitem{nerf-w}
Ricardo Martin-Brualla, Noha Radwan, Mehdi S.~M. Sajjadi, Jonathan~T. Barron,
  Alexey Dosovitskiy, and Daniel Duckworth.
\newblock {NeRF in the Wild: Neural Radiance Fields for Unconstrained Photo
  Collections}.
\newblock In {\em Proceedings of the IEEE/CVF Conference on Computer Vision and
  Pattern Recognition (CVPR)}, 2021.

\bibitem{occupancy_networks}
Lars Mescheder, Michael Oechsle, Michael Niemeyer, Sebastian Nowozin, and
  Andreas Geiger.
\newblock {Occupancy Networks: Learning 3D Reconstruction in Function Space}.
\newblock In {\em Proceedings of the IEEE/CVF Conference on Computer Vision and
  Pattern Recognition (CVPR)}, 2019.

\bibitem{mildenhall2019llff}
Ben Mildenhall, Pratul~P. Srinivasan, Rodrigo Ortiz-Cayon, Nima~Khademi
  Kalantari, Ravi Ramamoorthi, Ren Ng, and Abhishek Kar.
\newblock {Local Light Field Fusion: Practical View Synthesis with Prescriptive
  Sampling Guidelines}.
\newblock {\em ACM Trans. Graph.}, 2019.

\bibitem{nerf}
Ben Mildenhall, Pratul~P. Srinivasan, Matthew Tancik, Jonathan~T. Barron, Ravi
  Ramamoorthi, and Ren Ng.
\newblock {NeRF: Representing Scenes as Neural Radiance Fields for View
  Synthesis}.
\newblock In {\em Proceedings of the European Conference on Computer Vision
  (ECCV)}, 2020.

\bibitem{instantngp}
Thomas M\"uller, Alex Evans, Christoph Schied, and Alexander Keller.
\newblock {Instant Neural Graphics Primitives with a Multiresolution Hash
  Encoding}.
\newblock {\em ACM Trans. Graph.}, 2022.

\bibitem{extracting_triangle}
Jacob Munkberg, Jon Hasselgren, Tianchang Shen, Jun Gao, Wenzheng Chen, Alex
  Evans, Thomas Mueller, and Sanja Fidler.
\newblock {Extracting Triangular 3D Models, Materials, and Lighting From
  Images}.
\newblock In {\em Proceedings of the IEEE/CVF Conference on Computer Vision and
  Pattern Recognition (CVPR)}, 2022.

\bibitem{snerf_siggraph}
Thu Nguyen-Phuoc, Feng Liu, and Lei Xiao.
\newblock {SNeRF: Stylized Neural Implicit Representations for 3D Scenes}.
\newblock {\em ACM Trans. Graph.}, 2022.

\bibitem{deepsdf}
Jeong~Joon Park, Peter Florence, Julian Straub, Richard Newcombe, and Steven
  Lovegrove.
\newblock {DeepSDF: Learning Continuous Signed Distance Functions for Shape
  Representation}.
\newblock In {\em Proceedings of the IEEE/CVF Conference on Computer Vision and
  Pattern Recognition (CVPR)}, 2019.

\bibitem{hyper_nerf}
Keunhong Park, U. Sinha, Peter Hedman, Jonathan~T. Barron, Sofien Bouaziz,
  Dan~B. Goldman, Ricardo Martin-Brualla, and Steven~M. Seitz.
\newblock {HyperNeRF: A Higher-Dimensional Representation for Topologically
  Varying Neural Radiance Fields}.
\newblock {\em ACM Trans. Graph.}, 2021.

\bibitem{pytorch}
Adam Paszke, Sam Gross, Francisco Massa, Adam Lerer, James Bradbury, Gregory
  Chanan, Trevor Killeen, Zeming Lin, Natalia Gimelshein, Luca Antiga, Alban
  Desmaison, Andreas Kopf, Edward Yang, Zachary DeVito, Martin Raison, Alykhan
  Tejani, Sasank Chilamkurthy, Benoit Steiner, Lu Fang, Junjie Bai, and Soumith
  Chintala.
\newblock {PyTorch: An Imperative Style, High-Performance Deep Learning
  Library}.
\newblock In H. Wallach, H. Larochelle, A. Beygelzimer, F. d\textquotesingle
  Alch\'{e}-Buc, E. Fox, and R. Garnett, editors, {\em Advances in Neural
  Information Processing Systems 32}, pages 8024--8035. Curran Associates,
  Inc., 2019.

\bibitem{kilonerf}
Christian Reiser, Songyou Peng, Yiyi Liao, and Andreas Geiger.
\newblock {KiloNeRF: Speeding up Neural Radiance Fields with Thousands of Tiny
  MLPs}.
\newblock In {\em Proceedings of the IEEE/CVF International Conference on
  Computer Vision (ICCV)}, 2021.

\bibitem{nvos}
Zhongzheng Ren, Aseem Agarwala, Bryan Russell, Alexander~G. Schwing, and Oliver
  Wang.
\newblock {Neural Volumetric Object Selection}.
\newblock In {\em Proceedings of the IEEE/CVF Conference on Computer Vision and
  Pattern Recognition (CVPR)}, 2022.

\bibitem{grabcut}
Carsten Rother, Vladimir Kolmogorov, and Andrew Blake.
\newblock {GrabCut: interactive foreground extraction using iterated graph
  cuts}.
\newblock {\em ACM SIGGRAPH 2004 Papers}, 2004.

\bibitem{fbrs}
Konstantin Sofiiuk, Ilya~A. Petrov, Olga Barinova, and Anton Konushin.
\newblock {F-BRS: Rethinking Backpropagating Refinement for Interactive
  Segmentation}.
\newblock In {\em Proceedings of the IEEE/CVF Conference on Computer Vision and
  Pattern Recognition (CVPR)}, 2020.

\bibitem{nerv2020}
Pratul~P. Srinivasan, Boyang Deng, Xiuming Zhang, Matthew Tancik, Ben
  Mildenhall, and Jonathan~T. Barron.
\newblock {NeRV: Neural Reflectance and Visibility Fields for Relighting and
  View Synthesis}.
\newblock In {\em Proceedings of the IEEE/CVF Conference on Computer Vision and
  Pattern Recognition (CVPR)}, 2021.

\bibitem{dvgo}
Cheng Sun, Min Sun, and Hwann{-}Tzong Chen.
\newblock {Direct Voxel Grid Optimization: Super-fast Convergence for Radiance
  Fields Reconstruction}.
\newblock In {\em Proceedings of the IEEE/CVF Conference on Computer Vision and
  Pattern Recognition (CVPR)}, 2022.

\bibitem{dvgov2}
Cheng Sun, Min Sun, and Hwann-Tzong Chen.
\newblock {Improved Direct Voxel Grid Optimization for Radiance Fields
  Reconstruction}.
\newblock {\em arXiv,abs/2206.05085}, 2022.

\bibitem{block_nerf}
Matthew Tancik, Vincent Casser, Xinchen Yan, Sabeek Pradhan, Ben Mildenhall,
  Pratul~P. Srinivasan, Jonathan~T. Barron, and Henrik Kretzschmar.
\newblock {Block-NeRF: Scalable Large Scene Neural View Synthesis}.
\newblock In {\em Proceedings of the IEEE/CVF Conference on Computer Vision and
  Pattern Recognition (CVPR)}, 2022.

\bibitem{nerf_survey}
Ayush Tewari, Justus Thies, Ben Mildenhall, Pratul Srinivasan, Edith Tretschk,
  Yifan Wang, Christoph Lassner, Vincent Sitzmann, Ricardo Martin-Brualla,
  Stephen Lombardi, Tomas Simon, Christian Theobalt, Matthias Nießner, Jon~T.
  Barron, Gordon Wetzstein, Michael Zollhöfer, and Vladislav Golyanik.
\newblock {Advances in Neural Rendering}.
\newblock {\em Comput. Graph. Forum}, 2022.

\bibitem{bilateral_filter}
C. Tomasi and R. Manduchi.
\newblock {Bilateral filtering for gray and color images}.
\newblock In {\em {Sixth International Conference on Computer Vision}}, 1998.

\bibitem{nerf_nonrigid}
Edgar Tretschk, Ayush Tewari, Vladislav Golyanik, Michael Zollh{\"o}fer,
  Christoph Lassner, and Christian Theobalt.
\newblock Non-rigid neural radiance fields: Reconstruction and novel view
  synthesis of a dynamic scene from monocular video.
\newblock In {\em Proceedings of the IEEE/CVF International Conference on
  Computer Vision (ICCV)}, 2021.

\bibitem{N3F}
Vadim Tschernezki, Iro Laina, Diane Larlus, and Andrea Vedaldi.
\newblock {Neural Feature Fusion Fields}: {3D} distillation of self-supervised
  {2D} image representations.
\newblock In {\em International Conference on 3D Vision (3DV)}, 2022.

\bibitem{cla_nerf}
Wei-Cheng Tseng, Hung‐Ju Liao, Yen-Chen Lin, and Min Sun.
\newblock {CLA-NeRF: Category-Level Articulated Neural Radiance Field}.
\newblock {\em Proc. of the {IEEE} International Conference on Robotics and
  Automation}, 2022.

\bibitem{multi_plane_images}
Richard Tucker and Noah Snavely.
\newblock {Single-View View Synthesis With Multiplane Images}.
\newblock In {\em Proceedings of the IEEE/CVF Conference on Computer Vision and
  Pattern Recognition (CVPR)}, 2020.

\bibitem{nerf_text}
Can Wang, Menglei Chai, Mingming He, Dongdong Chen, and Jing Liao.
\newblock {CLIP-NeRF: Text-and-Image Driven Manipulation of Neural Radiance
  Fields}.
\newblock In {\em Proceedings of the IEEE/CVF Conference on Computer Vision and
  Pattern Recognition (CVPR)}, 2022.

\bibitem{d2nerf}
Tianhao Wu, Fangcheng Zhong, Andrea Tagliasacchi, Forrester Cole, and Cengiz
  Oztireli.
\newblock {D$^2$NeRF: Self-Supervised Decoupling of Dynamic and Static Objects
  from a Monocular Video}.
\newblock In {\em Adv. Neural Inform. Process. Syst.}, 2022.

\bibitem{city_nerf}
Yuanbo Xiangli, Linning Xu, Xingang Pan, Nanxuan Zhao, Anyi Rao, Christian
  Theobalt, Bo Dai, and Dahua Lin.
\newblock {BungeeNeRF: Progressive Neural Radiance Field for Extreme
  Multi-scale Scene Rendering}.
\newblock In {\em Proceedings of the European Conference on Computer Vision
  (ECCV)}, 2022.

\bibitem{nerf_survey2}
Yiheng Xie, Towaki Takikawa, Shunsuke Saito, Or Litany, Shiqin Yan, Numair
  Khan, Federico Tombari, James Tompkin, Vincent Sitzmann, and Srinath Sridhar.
\newblock {Neural Fields in Visual Computing and Beyond}.
\newblock {\em Comput. Graph. Forum}, 2022.

\bibitem{cagenerf}
Tianhan Xu and Tatsuya Harada.
\newblock {Deforming Radiance Fields with Cages}.
\newblock In {\em Proceedings of the European Conference on Computer Vision
  (ECCV)}, 2022.

\bibitem{nerf_supervision}
Lin Yen-Chen, Peter~R. Florence, Jonathan~T. Barron, Tsung-Yi Lin, Alberto
  Rodriguez, and Phillip Isola.
\newblock {NeRF-Supervision: Learning Dense Object Descriptors from Neural
  Radiance Fields}.
\newblock In {\em Proc. of the {IEEE} International Conference on Robotics and
  Automation}, 2022.

\bibitem{plenoctrees}
Alex Yu, Ruilong Li, Matthew Tancik, Hao Li, Ren Ng, and Angjoo Kanazawa.
\newblock {PlenOctrees for Real-time Rendering of Neural Radiance Fields}.
\newblock In {\em Proceedings of the IEEE/CVF Conference on Computer Vision and
  Pattern Recognition (CVPR)}, 2021.

\bibitem{arf_eccv}
Kai Zhang, Nick Kolkin, Sai Bi, Fujun Luan, Zexiang Xu, Eli Shechtman, and Noah
  Snavely.
\newblock {ARF: Artistic Radiance Fields}.
\newblock In {\em Proceedings of the European Conference on Computer Vision
  (ECCV)}, 2022.

\bibitem{iron}
Kai Zhang, Fujun Luan, Zhengqi Li, and Noah Snavely.
\newblock {IRON: Inverse Rendering by Optimizing Neural SDFs and Materials from
  Photometric Images}.
\newblock 2022.

\bibitem{nerf++}
Kai Zhang, Gernot Riegler, Noah Snavely, and Vladlen Koltun.
\newblock {NeRF++: Analyzing and Improving Neural Radiance Fields}.
\newblock {\em arXiv:2010.07492}, 2020.

\bibitem{nerfactor}
Xiuming Zhang, Pratul~P. Srinivasan, Boyang Deng, Paul Debevec, William~T.
  Freeman, and Jonathan~T. Barron.
\newblock {NeRFactor: Neural Factorization of Shape and Reflectance under an
  Unknown Illumination}.
\newblock {\em ACM Trans. Graph.}, 2021.

\bibitem{semanticnerf}
Shuaifeng Zhi, Tristan Laidlow, Stefan Leutenegger, and Andrew Davison.
\newblock {In-Place Scene Labelling and Understanding with Implicit Scene
  Representation}.
\newblock In {\em Proceedings of the IEEE/CVF International Conference on
  Computer Vision (ICCV)}, 2021.

\bibitem{radianceFields}
Kun Zhou, Yaohua Hu, Stephen Lin, Baining Guo, and Heung-Yeung Shum.
\newblock {Precomputed Shadow Fields for Dynamic Scenes}.
\newblock {\em ACM Trans. Graph.}, 2005.

\end{thebibliography}
    }
    \clearpage
    \clearpage
    %  supplementary material added here comment out if not required
    \setcounter{section}{0}
        \section{Implementation Details}
    {
        \label{sec:sup_implementation_details}
        All the methods proposed in the paper have been implemented using PyTorch \cite{pytorch} branching off the code provided by DVGOv2 \cite{dvgov2}. All experiments are performed using a commodity hardware equipped with AMD Ryzen 5800x and a NVIDIA RTX 3090. 
        
        The feature components of the radiance fields namely radiance latent vectors and the learnt DINO features are stored using VM decomposition proposed by TensoRF \cite{tensorf}. For radiance latent vectors, we use \emph{VM-48} representation of TensoRF and for DINO features, we use \emph{VM-64} variant of TensoRF. The segmentation masks and densities have been stored as a full voxel grids. 
        % Though the bitmap is having a memory in the orders of $\mathcal{O}(n^3)$ the actual memory foot is substantially small as bitmap only requires a storage of one bit for each location in the full volumetric grid.
        
        The DINO \emph{ViT-b8} \cite{dino} model provides 768 features for each patch of $8 \times 8$ pixels in an image. We reduce the dimensionality of these features by doing a principal component analysis reducing the effective dimension to 64. This is consistent with the prior works \cite{N3F,DFF}. For each pixel, the feature is calculated by referring to the feature of the respective patch that pixel corresponds to.
        
        We first pre-train the model for the volumetric density and radiance for $20,000$ iterations. Once the radiance field is stabilized on the VM-48 TensoRF representation, we introduce distillation using \emph{student-teacher} strategy similar to that of \cite{N3F, DFF} on the VM-64 TensoRF variant. Upon adoption, the resultant VM-48 variant of TensoRF along with its shallow MLP represents the radiance field, and VM-64 constitute the distilled features. It is to be noted that the distilled features are not accompanied by a shallow MLP. The features are store at voxel lattice locations and tri-linearly interpolated to be compared and optimized against the DINO features without the involvement of any non-linearity.  The adoption is done with $\lambda = 0.001$ for the weighted loss function for $5,000$ iterations. The loss is taken on the features and radiance together to maintain consistency.
        
        We choose $K=10$ when applying K-Means to the set of features selected from the user's brush stroke. For the bilateral search, the value of $\sigma_\phi$ and $\sigma_s$ are set to $10.0$ and the $1.0$ respectively while the threshold value $\tau$ is $0.1$.
    }
    
    \section{Scene Editing}
    {
        \label{sec:sup_scene_editing}
        In this section, we explain the procedures that were followed for editing the 3D scenes post segmentation. The segmentation procedure provides a 3D bit map representing the segmented voxels.
        Utilization of an additional bitmap also assists in faster rendering as the voxels with segmentation mask values of $0$ can easily be filtered out. \cref{fig:sup_editing} shows the additional results of scene editing.
        
        \subsection{Object Removal}
        {
            \label{sub_sec:sup_object_removal}
            For removing a segmented object from the scene, we alter the evaluation of the density for a 3D point. We simultaneously evaluate the bit map value $b_x$ at the queried point. To segmented the object of interest (foreground), the effective density $\sigma'_x$ is $\sigma_x * b_x$. Similarly, to render the background the effective density $\sigma'_x$ is $\sigma_x * (1.0 - b_x)$.
        }
        
        \subsection{Translation}
        {
            \label{sub_sec:sup_translation}
            If an object needs to be moved to another location, the ray queries lying inside the object's voxel space can be shifted to the desired location. Let $t$ be the translation vector for the object to be moved, then the object's ray-point query changes as shown below.

            \begin{align*}
                \small
                    \sigma'_x, rgb'_x = \sigma_x, rgb_x \ \forall \ b_x = 0\\
                    \sigma'_x, rgb'_x = \sigma_{x+t}, rgb_{x+t} \ \forall \ b_x = 1
            \end{align*}
        }
        
        \subsection{Scene Composition}
        {
            \label{sub_sec:sup_scene_composition}
            To perform scene composition, we follow a similar strategy used by D$^2$NeRF \cite{d2nerf}. We alter the volumetric rendering equation to account for density and color from both the scenes as shown below:
            \begin{align*}
            %\label{eq:sup_composition}
                \small
                    \hat{C}(r) = \int_{t_n}^{t_f} T(t) \left(\sigma_1(t) c_1(t) + \sigma_2(t) c_2(t)\right) dt \\
                    T(t) = exp\left(- \int_{t_n}^{t} (\sigma_1(s) + \sigma_2(s))\right) ds \\
            \end{align*}
            
            The results for scene composition have been shown in the main paper and \cref{fig:sup_editing} of the supplementary.
        }
        
        \subsection{Appearance Editing}
        {
            \label{sub_sec:sup_appearance_editing}
            Here, we apply style transfer on an already composed scene. We first calculate a 3D bitmap for the JCB lego in the \kitchen scene. Then, we generate a new set of stylized training images using the method proposed by \cite{johnson, goel2022styletrf} using a reference image. The appearance latent vectors and the rendering MLP is fine-tuned according to the new training images while keeping the density and feature weights frozen. This transfers the style from a reference image to the 3D object.
        }
        \begin{figure*}[t]
    \centering
    \begin{minipage}{0.24\linewidth}
        \centering
        \subcaptionbox{Original Rendered Image}{\includegraphics[width=\textwidth, height=3cm]{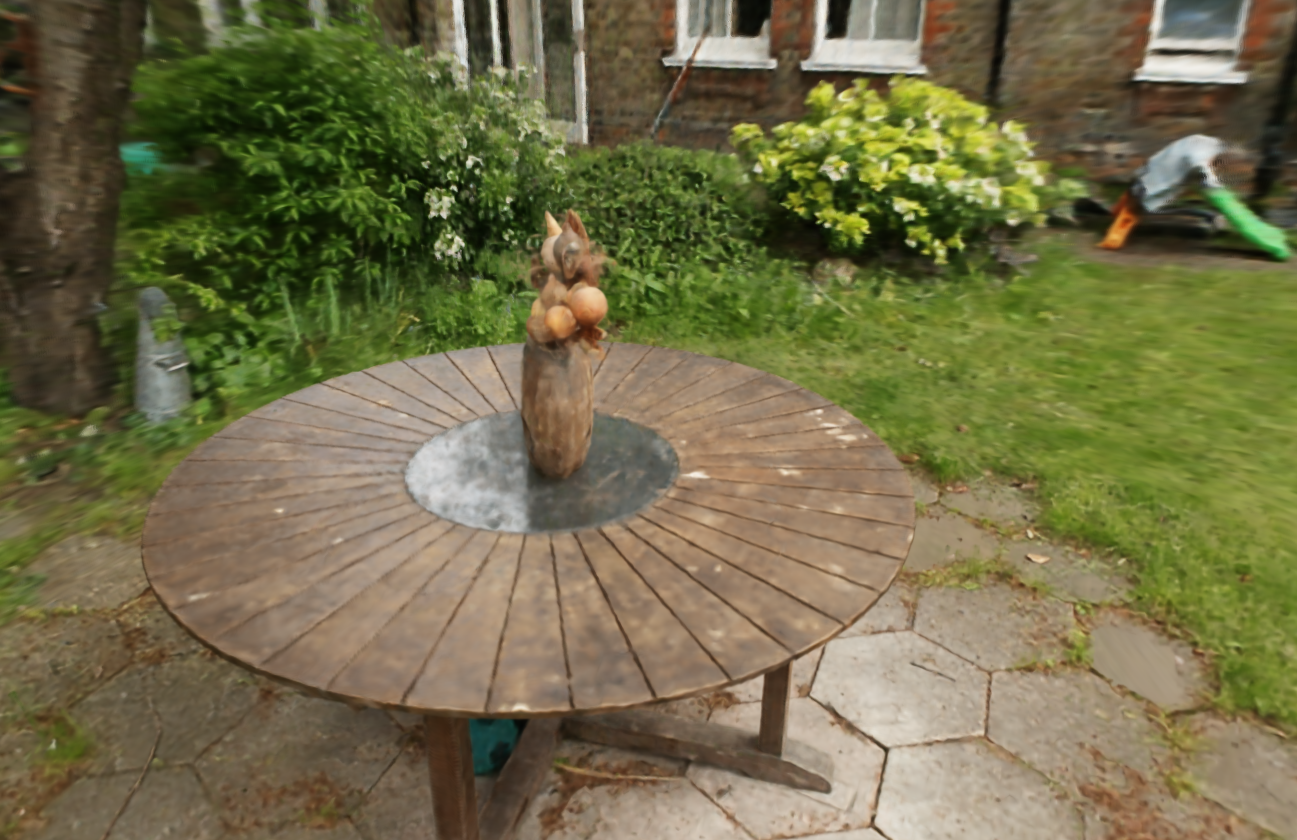}}
    \end{minipage}
    \begin{minipage}{0.24\linewidth}
        \centering
        \subcaptionbox{Removal of Pot}{\includegraphics[width=\textwidth, height=3cm]{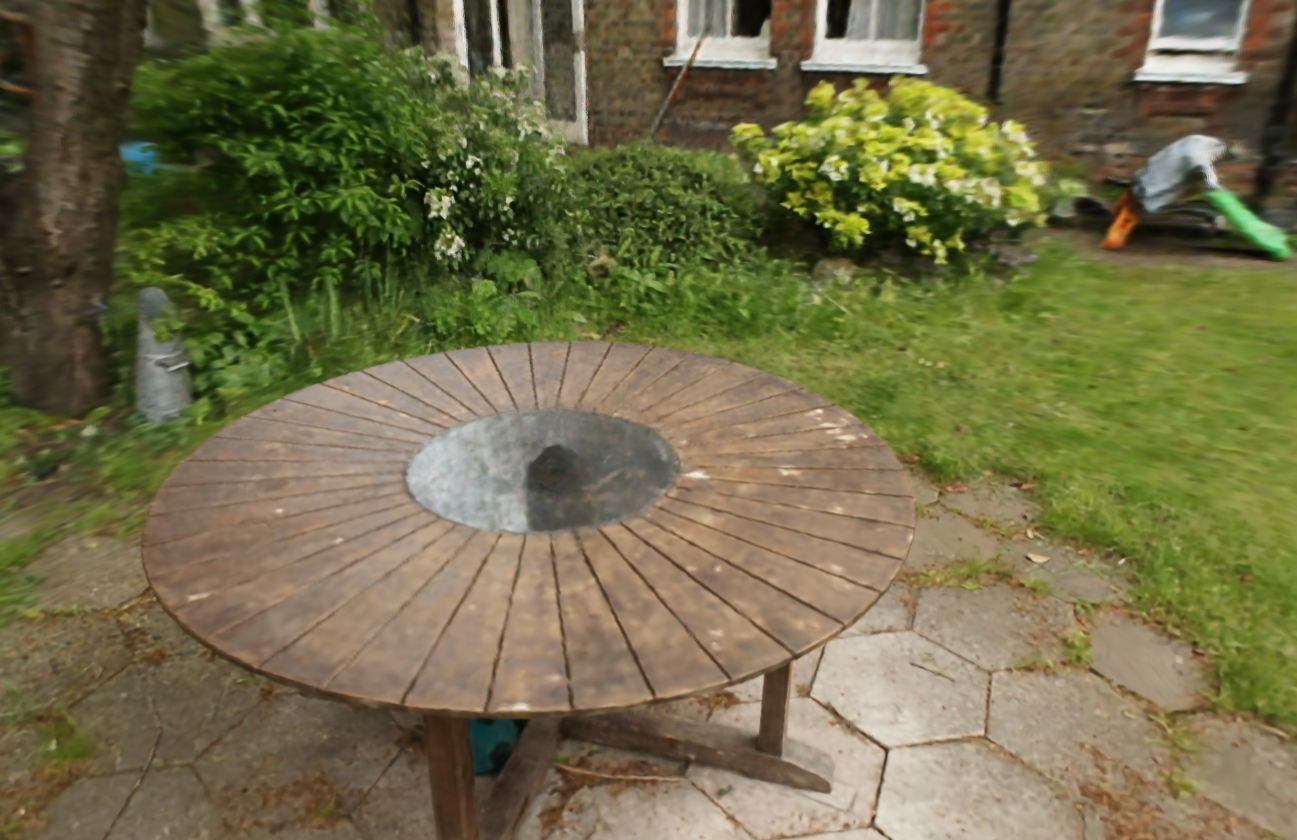}}
    \end{minipage}
    \begin{minipage}{0.24\linewidth}
        \centering
        \subcaptionbox{Composition}{\includegraphics[width=\textwidth, height=3cm]{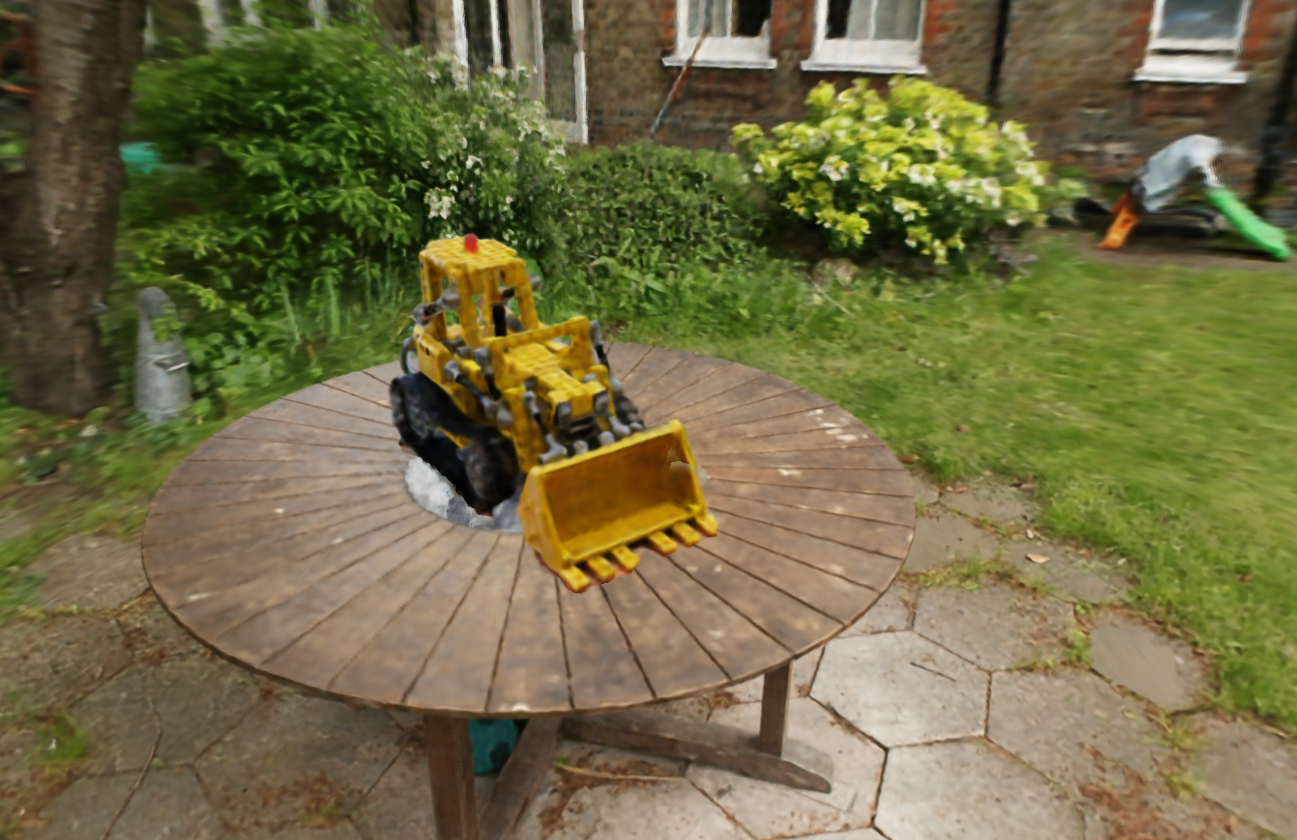}}
    \end{minipage}
    \begin{minipage}{0.24\linewidth}
        \centering
        \subcaptionbox{Style Transfer}{\includegraphics[width=\textwidth, height=3cm]{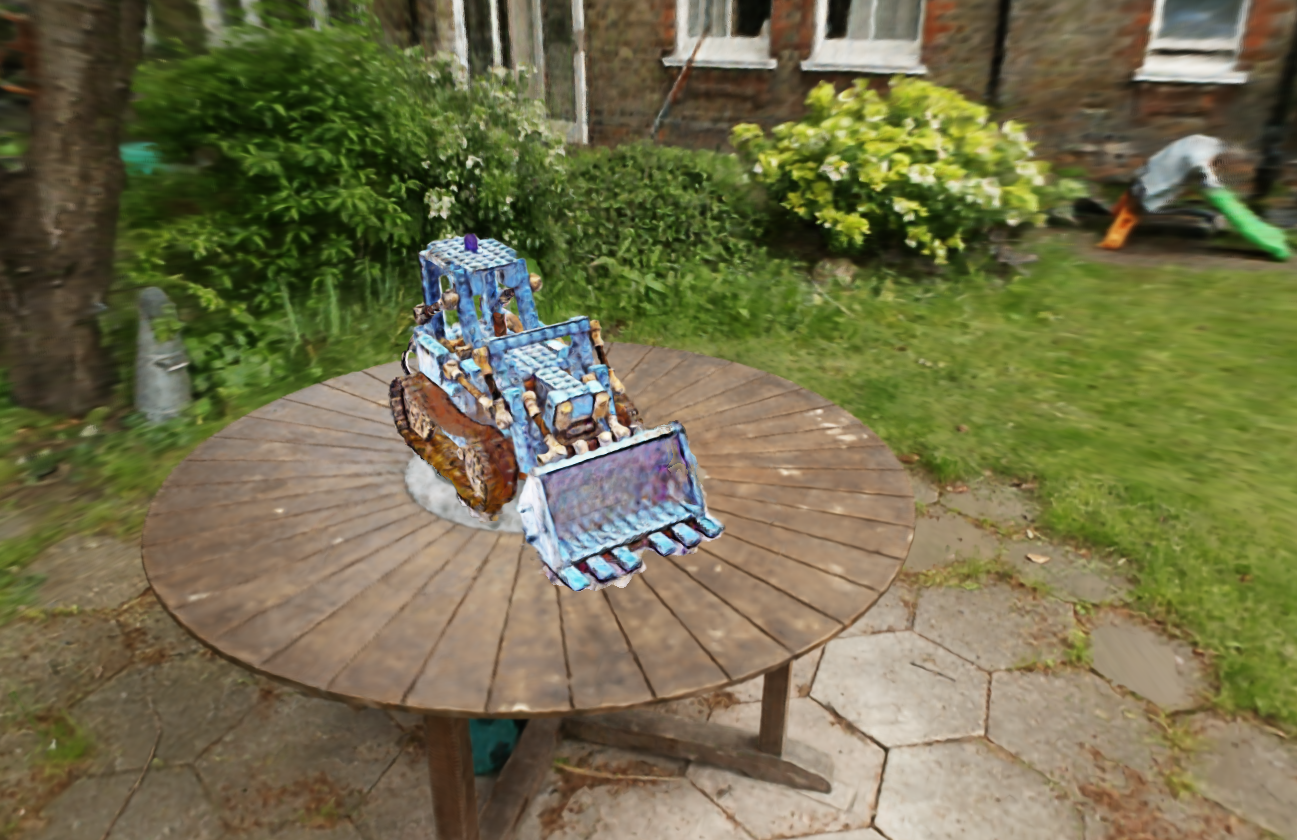}}
    \end{minipage}
    \caption{\emph{Seamless Progressive Scene Editing}:
    {Image (a) is the reference rendered viewpoint. In (b), the pot has been removed. Image (c) shows scene composition. The JCB from \kitchen scene has been placed on the top of the table in the \garden scene. Image (d) shows appearance editing of specific objects. We apply style transfer on just the JCB. For more details please refer to Sec. \ref{sec:sup_scene_editing}.}}
    \label{fig:sup_editing}
\end{figure*}
        \begin{figure}
    \centering
    \begin{minipage}{0.49\linewidth}
        \centering
        \subcaptionbox{With Steel Balls}{{
        \begin{tikzpicture}
            \node[anchor=south west,inner sep=0] (image) at (0,0) {\includegraphics[width=\textwidth]{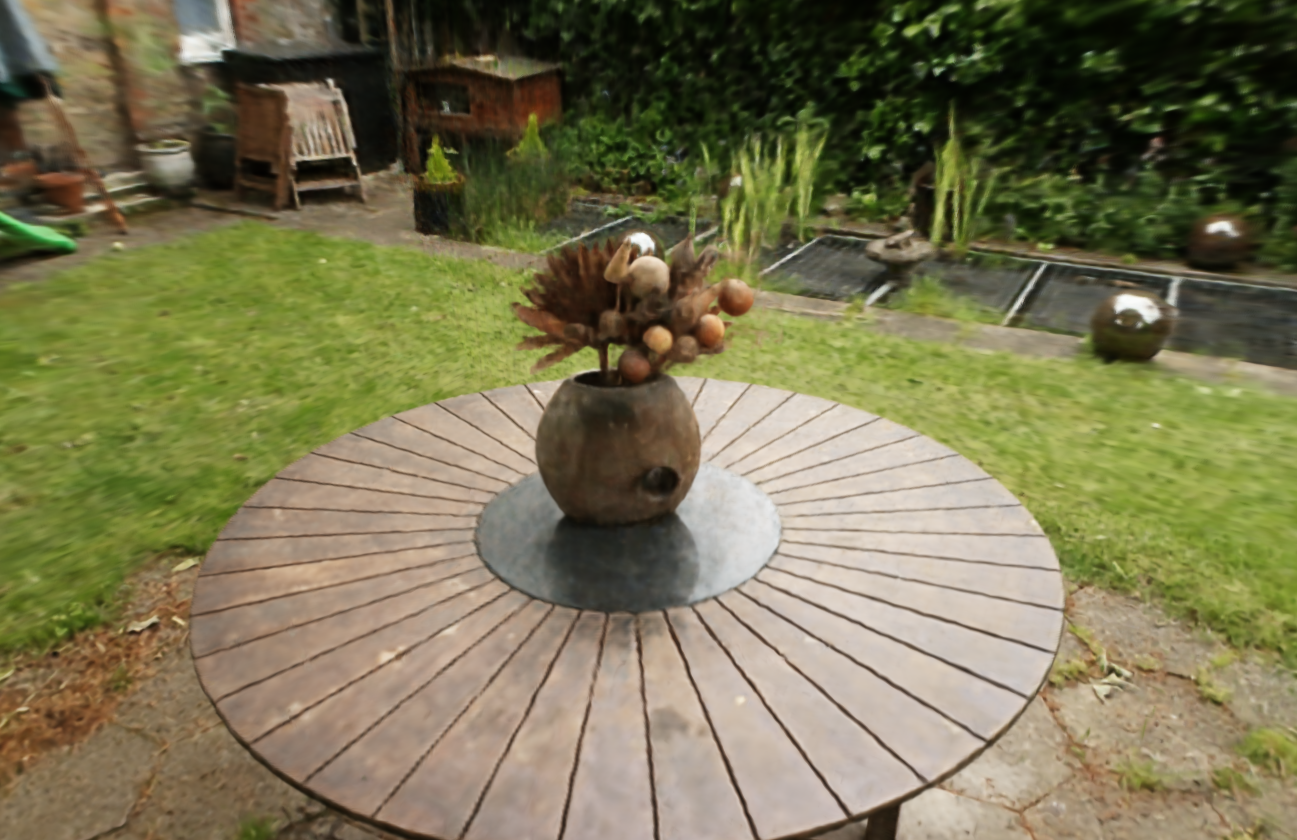}};
            \begin{scope}[x={(image.south east)},y={(image.north west)}]
                \draw[orange,thick] (0.8,0.55) rectangle (1.0, 0.85);
            \end{scope}
        \end{tikzpicture}
        }}
    \end{minipage}
    \begin{minipage}{0.49\linewidth}
        \centering
        \subcaptionbox{Without Steel Balls}{{
        \begin{tikzpicture}
            \node[anchor=south west,inner sep=0] (image) at (0,0) {\includegraphics[width=\textwidth]{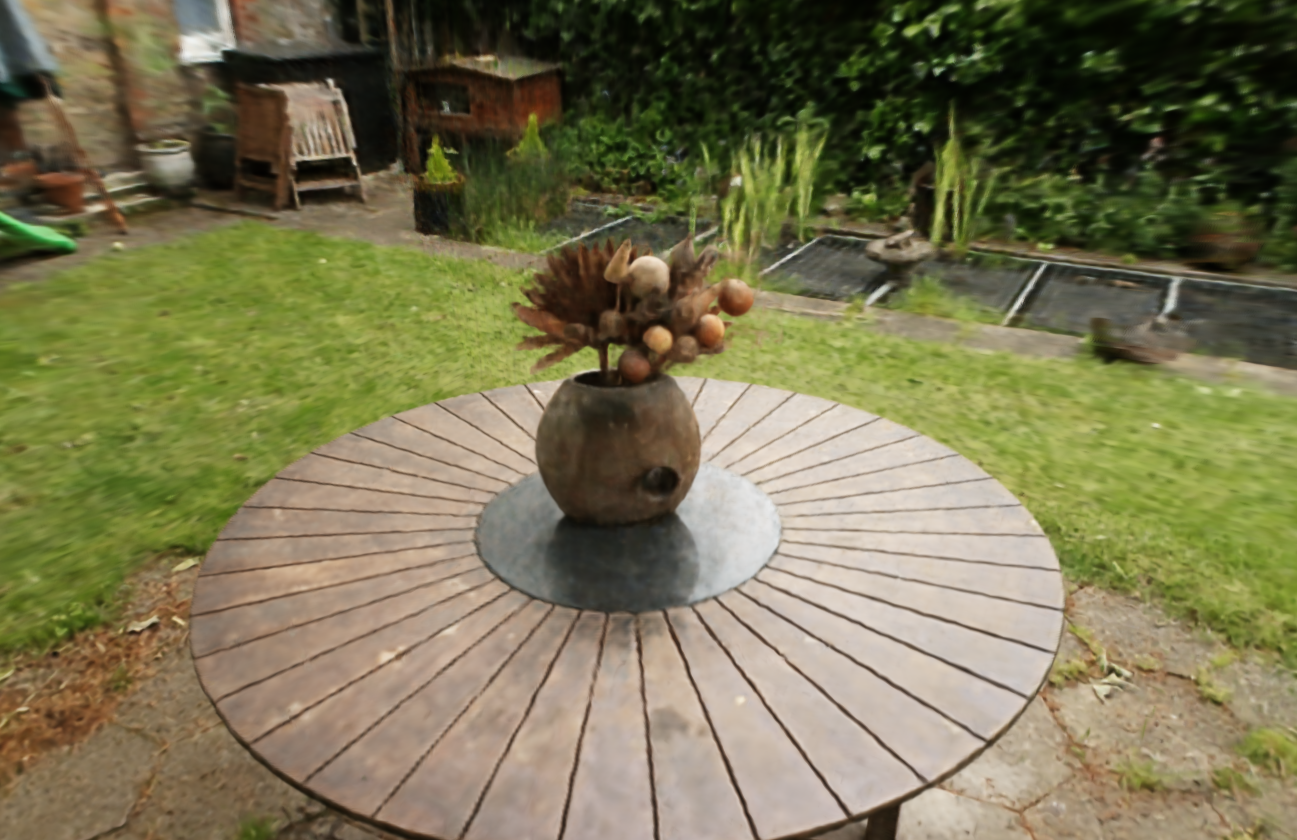}};
        \end{tikzpicture}
        }}
    \end{minipage}
    
    \caption{ \emph{Removal of Steel Balls}:
    {We use the MipNeRF360\cite{mipnerf360} formulation in voxel space for unbounded 360 degree scenes. This gives fewer number of voxels to the background objects compared to the central volume of interest. In this scene, we remove the steel balls appearing in the background region of the scene.}}
    \label{fig:stellballs}
\end{figure}

    }

    \section{Quantitative Analysis}
    {
    
        To quantitatively compare our method on the LLFF Dataset \cite{mildenhall2019llff}, we hand-annotate the segmentation masks for the prominent objects in the \chesstable, \colorfountain, \stove and \shoerack scenes.
        \cref{tab:llff_metric} reports the segmentation metrics for the four scenes. In our method, to predict the segmentation mask, we threshold $\alpha$ to be greater than $0.1$ while rendering.
        This removes the low volumetric density seeping in that contribute negligibly in the rendered visuals.
        
        %        Please note that any such metric calculated on the rendered images is fundamentally incorrect for 3D segmentation especially radiance fields. In the segmentation of radiance fields, artifacts which contribute very little to the radiance may also appear in the segmented image. For instance, in the \chesstable scene, the nearby air appears in the segmented scene which has low volumetric density. To a 2D segmentation metric it may seem prominent, however in 3D applications such as scene composition, it is irrelevant. Therefore, although we have reported the common 2D segmentation metrics, we urge our readers to not rely on it for measuring accuracy of segmentation in radiance fields.

        \begin{table*}[h]
\begin{center}
\begin{tabular}{ |c|c|c|c|c| } 
 \hline
 \textbf{Scene} & \textbf{Metric}  & \textbf{N3F} & \textbf{Ours (Patch)} & \textbf{Ours (Stroke)} \\ 
 \hline
 \multirow{2}{*}{\chesstable} & Mean IoU $\uparrow$ & 0.344 & 0.864 & \textbf{0.912} \\ \cline{2-5}
 & Accuracy $\uparrow$ & 0.820 & 0.985 & \textbf{0.990} \\ \cline{2-5}
 & mAP $\uparrow$ & 0.334 & 0.874 & \textbf{0.916} \\ 
 \hline
 
 \multirow{2}{*}{\colorfountain} & Mean IoU $\uparrow$ & 0.871 & \textbf{0.927} & \textbf{0.927} \\ \cline{2-5}
 & Accuracy $\uparrow$ & 0.979 & \textbf{0.989} & \textbf{0.989} \\ \cline{2-5}
 & mAP $\uparrow$ & 0.871 & \textbf{0.927} & \textbf{0.927} \\ 
 \hline
 
 \multirow{2}{*}{\stove} & Mean IoU $\uparrow$ & 0.416 & \textbf{0.827} & 0.819 \\ \cline{2-5}
 & Accuracy $\uparrow$ & 0.954 & \textbf{0.992} & \textbf{0.992} \\ \cline{2-5}
 & mAP $\uparrow$ & 0.387 & \textbf{0.824} & 0.817 \\ 
 \hline
 
 \multirow{2}{*}{\shoerack} & Mean IoU $\uparrow$ & 0.589 & 0.763 & \textbf{0.861} \\ \cline{2-5}
 & Accuracy $\uparrow$ & 0.913 & 0.965 & \textbf{0.980} \\ \cline{2-5}
 & mAP $\uparrow$ & 0.582 & 0.773 & \textbf{0.869} \\ 
 \hline
\end{tabular}
\end{center}
\caption{\label{tab:llff_metric} This table denotes the Mean IoU (Intersection Over Union), Accuracy and Mean Average Precision measurements for the four LLFF scenes shown in the main paper. The ground truth segmentation masks have been hand-annotated for comparison.}        
\end{table*}
    }

   \section{Region Growing: Bilateral Growth}
    {
        \label{sec:sup_result_region_growth}

        In this section, we discuss the effect of bilateral filtering on the radiance fields and how it improves the final result. Even after employing an efficient feature-matching technique, we often obtain a high-confidence volumetric region with missing constituting parts.
        This is because the content search solely depends on feature distances while ignoring the spatial priors. To resolve this issue we resort to Bilateral search which exploits spatio-semantic domain priors resulting in accurate segmentation constituting all the desired regions of the semantic object.
        This is demonstrated in \cref{fig:bilateral_growth}, where the initial high-confidence region misses the outer leaf of the dry plant. While the bilateral region is growing, we iteratively add more details into the extracted region, finally obtaining desired volumetric content. This content can be further used for various purposes as discussed in \cref{sec:sup_scene_editing}.
        \begin{figure}
    \centering
    \begin{minipage}{0.49\linewidth}
        \centering
        \subcaptionbox{Rendered Image}{\includegraphics[width=\textwidth, height=3cm]{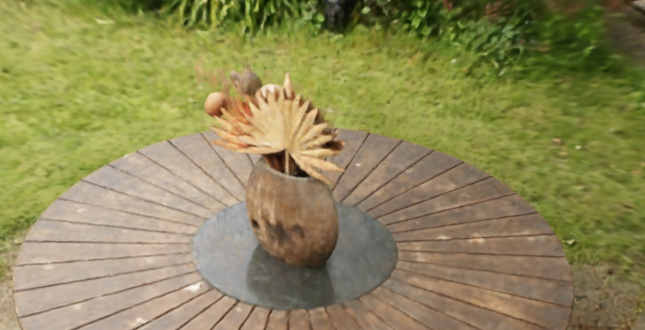}}
    \end{minipage}
    \begin{minipage}{0.49\linewidth}
        \centering
        \subcaptionbox{High Confidence Region}{\includegraphics[width=\textwidth, height=3cm]{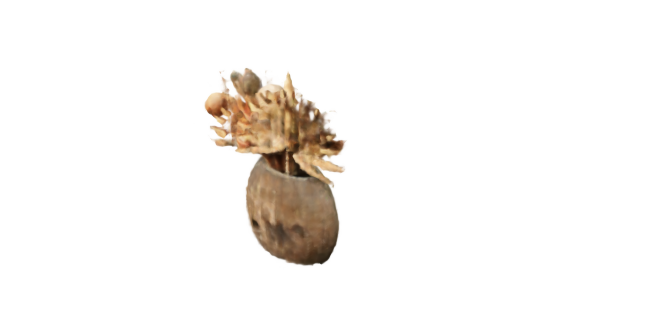}}
    \end{minipage}
    \begin{minipage}{0.49\linewidth}
        \centering
        \subcaptionbox{Iteration 1}{\includegraphics[width=\textwidth, height=3cm]{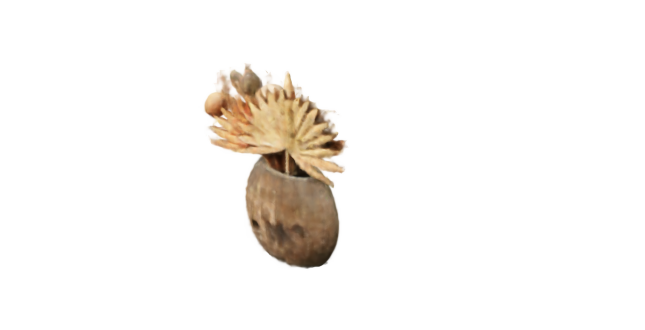}}
    \end{minipage}
    \begin{minipage}{0.49\linewidth}
        \centering
        \subcaptionbox{Iteration 2}{\includegraphics[width=\textwidth, height=3cm]{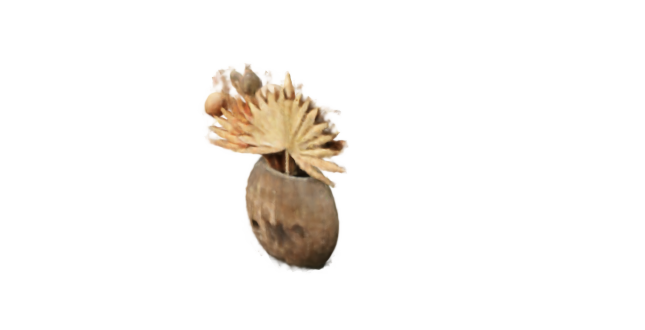}}
    \end{minipage}
    \caption{\emph{Region Growing}:
    {Image (a) is the reference rendered viewpoint. Image (b) is the high confidence region which misses out frontal region of the dry-leaf when extracting the content. Image (c) shows the result obtained after the first iteration of bilateral filtering, which captures most of the desired region of the leaf. Image (d) is the result of the bilateral filtering applied for the second time to include intricate details such as strands around the dry-leaf.}}
    \label{fig:bilateral_growth}
\end{figure}

    }

    \section{Evaluation strategies against SOTA techniques}
    {
        \label{sec:eval_sota}
        \subsection{N3F/DFF}
        {
            \label{sub_sec:eval_n3f_dff}
            As mentioned in the main document, we experiment with various thresholds in the case of N3F/DFF\cite{N3F, DFF}. We report the quantitative metrics (\cref{tab:llff_metric}) of our method against the best results of their methods. N3F/DFF don't produce good results for any threshold as shown in \cref{fig:all_n3f}.
        }
        \subsection{NVOS}
        {
            \label{sub_sec:eval_nvos}
            {
                \begin{table}[t]
    \begin{center}
        \begin{tabular}{|c|c|c|c|c|c|} 
         \hline
         \multicolumn{2}{|c|}{\textbf{NVOS}} & \multicolumn{2}{c}{\textbf{Ours(NVOS Stroke)}}  & \multicolumn{2}{|c|}{\textbf{Our best}}\\
         \hline
         mIOU & mAcc & mIOU & mAcc & mIOU & mAcc\\
         \hline
         70.1 & 92.0 & 83.75 & 96.4 & 90.8 & 98.2\\
         \hline
        \end{tabular}
    \end{center}
    \vspace{-5mm}
   \caption{\label{tab:nvos_comp} Quantitative metrics{(\em mIOU and mAcc)} of NVOS against Ours using NVOS provided strokes and additional strokes using our interactive feedback tool}
\vspace{-5mm}
\end{table}
                To make a fair comparison against NVOS\cite{nvos}, we utilize the masks provided by NVOS and evaluate the quantitative numbers on their dataset. We observe that our method out performs NVOS both qualitatively and quantitatively as shown \cref{fig:nvos_comp} and \cref{tab:nvos_comp} even when using their strokes. Using our own interactive tool with additional strokes achieve much better results.
            }
        }
    }

    \section{Interactive Segmentation}
    {
        Our method provides interactive segmentation capabilities to the user with the incorporation of positive and negative brush strokes similar to GrabCut \cite{grabcut}.
        
        Upon the addition of a new positive stroke, a new segmentation mask $b_p$ is calculated using the procedure described in the main paper. The user has the option to grow this new region using bilateral filtering until not required. The new segmentation mask $b_{new}$ is given by $b \cup b_p$.
        
        When the user adds a negative stroke, a new segmentation mask $b_n$ is calculated. Similar to a positive stroke, the user has the option to grow this region using bilateral filtering until not required. The new segmentation mask $b_{new}$ is given by $b \cap (b \cap b_n)'$ ($X'$ denotes the complement of $X$).
    }
    
    \section{Critical Analysis}
    {   
        \begin{figure}
    \centering
    \begin{minipage}{0.49\linewidth}
        \centering
        \subcaptionbox{\label{fig:trex_r1} Rendered View 1}
         {\includegraphics[width=\linewidth]{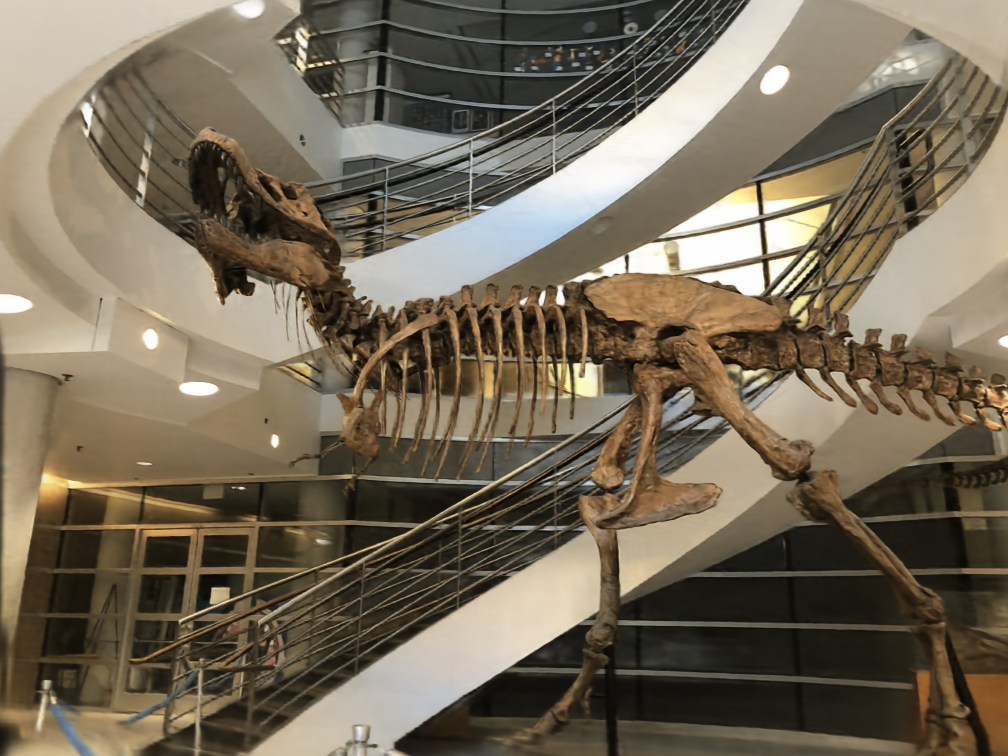}}
    \end{minipage}
    \begin{minipage}{0.49\linewidth}
        \centering
        \subcaptionbox{\label{fig:trex_r2} Rendered View 2}
         {\includegraphics[width=\linewidth]{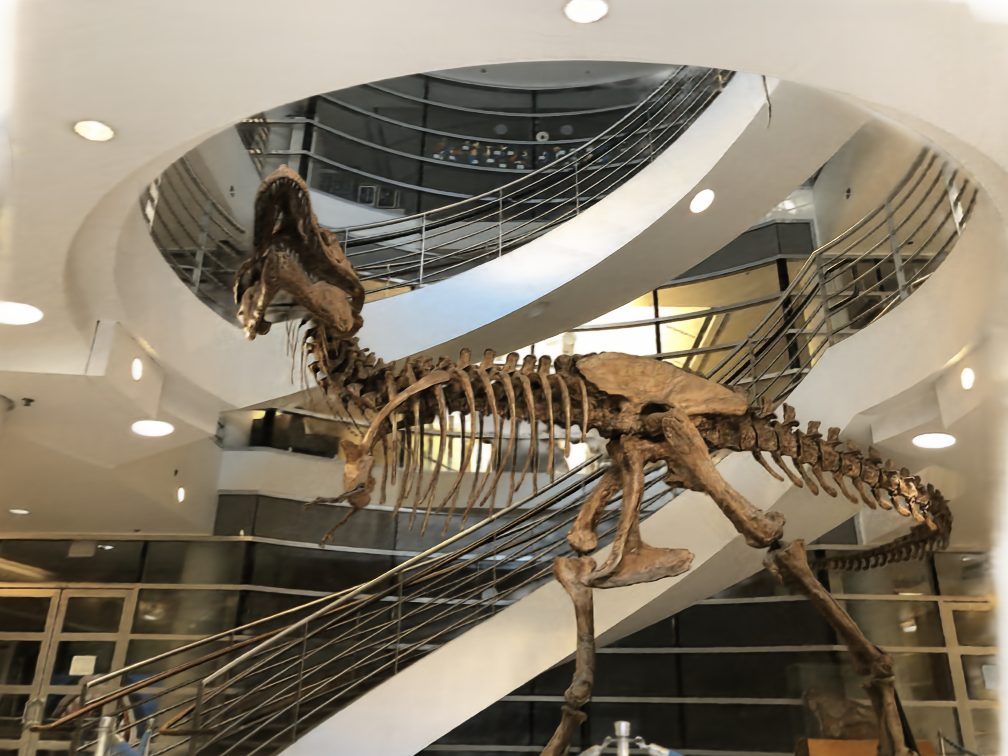}}
    \end{minipage}
    \begin{minipage}{0.49\linewidth}
        \centering
        \subcaptionbox{\label{fig:trex_seg} Segmented Trex}
         {\includegraphics[width=\linewidth]{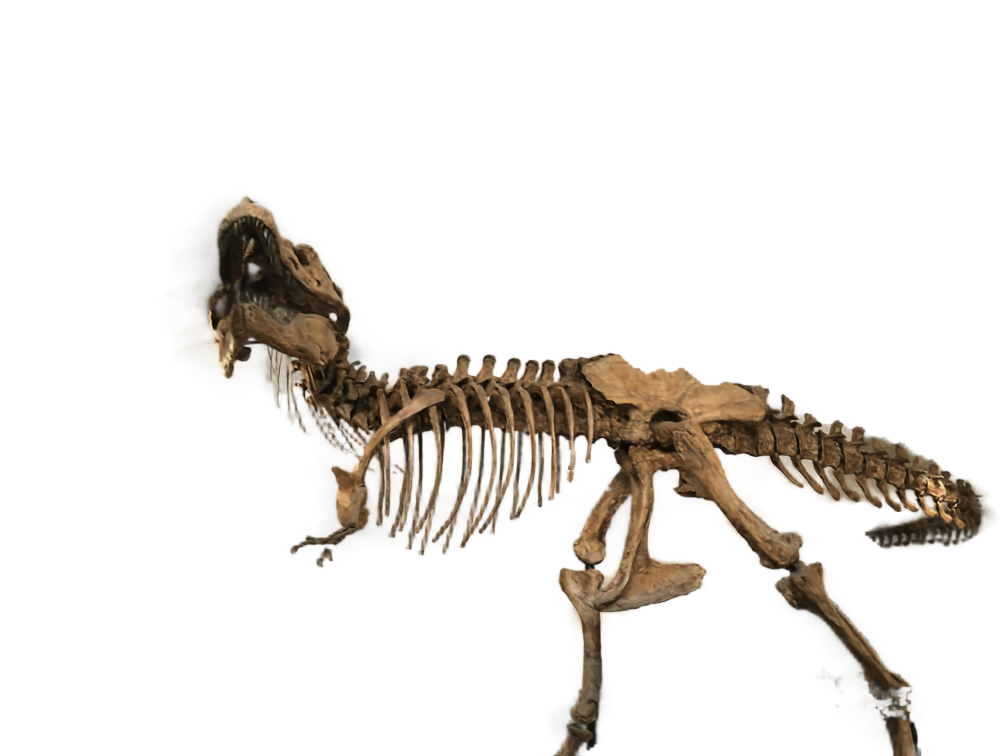}}
    \end{minipage}
    \begin{minipage}{0.49\linewidth}
        \centering
        \subcaptionbox{\label{fig:trex_depth} Depth Map}
        {\includegraphics[width=\linewidth]{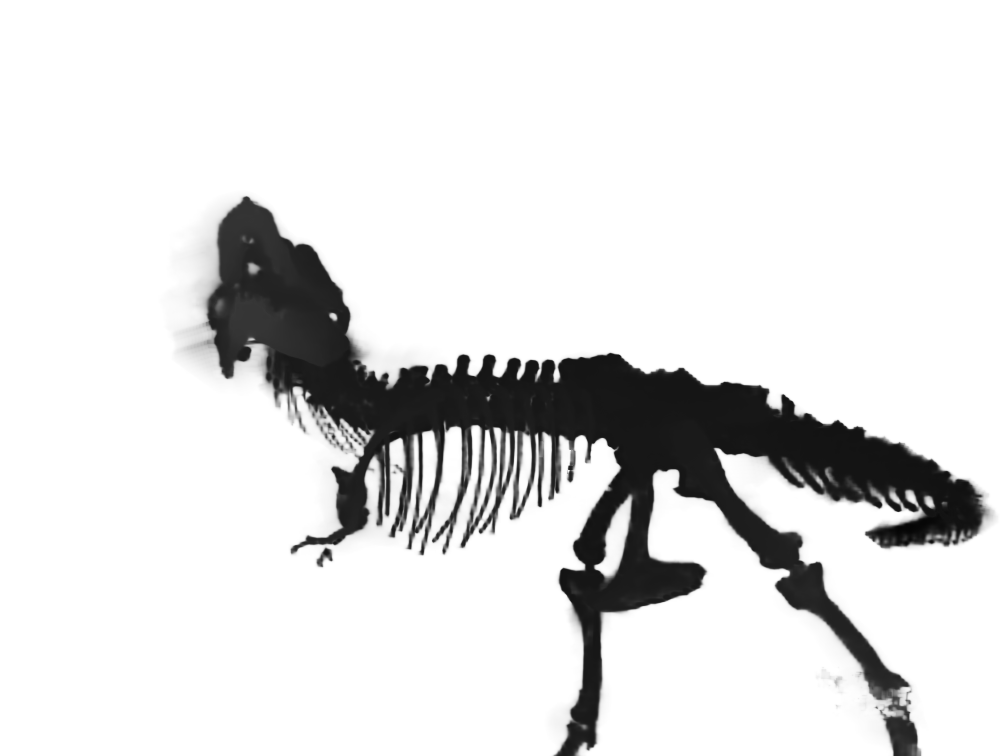}}
    \end{minipage}
     \caption{ \emph{Finer Segmentation}:
    {Images (a) and (b) show rendered views of T-Rex from the LLFF dataset \cite{mildenhall2019llff}. Image (c) shows the segmented output of T-Rex scene. Our method achieves fine-grained segmentation of objects such as the rib-cage bones of T-Rex. However, on close observation, the region near the tail bones background bleeds in. This is due to the wall and the tail-bone lie at the similar depth as shown in the depth map (d). This can be mitigated by having more 3D information (better training views) or higher voxel grid resolution.}}
    \label{fig:trex}
    \vspace{-5mm}
\end{figure}
        \subsection{DINO Features}
        {
            The teacher DINO features calculated on the training set of images are for patches of size 8x8. This method associates a total of 64 pixels to the same feature vector. As shown in \cref{fig:dino_features}, the teacher features appear to be in low resolution due to this. When performing the teacher-student training using the joint loss function, the features learnt by the student are finer in detail due to assistance from volumetric density. Hence, the student surpasses the teacher during distillation. This is evident from \cref{fig:dino_features} as features are allocated with distinct boundaries in the voxel space.
         
        }
        \subsection{Finer Segmentation}
        {
            
            \label{sub_sec:finer_segmentation}
            Our method can segment out fine-grained details such as the ribs of a T-Rex as shown in \cref{fig:trex}. However, it requires accurate 3D information to achieve this. In the T-Rex scene, the tail-bones cannot be distinguished from the wall behind, since the training set images do not cover views which indicate the separation. Therefore, the optimized model containing the wall and the tail bones lie at similar depths as shown in \cref{fig:trex_depth}. Use of additional images covering more viewpoints can circumvent this issue.
        }
        \begin{figure*}
    \begin{minipage}{0.24\linewidth}
    \begin{tikzpicture}
    \draw  node[inner sep=0] (reffern) {{\includegraphics[width=\linewidth]{assets/images/nvos/og_fern.png}}};
    \draw  node[inner sep=0, below = 0.01 of reffern] (reftrex) {{\includegraphics[width=\linewidth]{assets/images/nvos/og_trex.png}}};
    {\color{blue}\draw (+0.25, -2.) node {Reference};}
    \end{tikzpicture}
    \end{minipage}
    \begin{minipage}{0.24\linewidth}
    \centering
    \begin{tikzpicture}
    \draw  node[inner sep=0] (fernnvos) {{\includegraphics[width=\linewidth]{assets/images/nvos/nvos_fern.png}}};
    \draw node[below = 0.01 of fernnvos.south, inner sep=0] (trexnvos){{\includegraphics[width=\linewidth]{assets/images/nvos/nvos_trex.png}}};
    {\color{magenta}\draw (+0.25, -2) node {NVOS};}
    \end{tikzpicture}
    \end{minipage}  
    \begin{minipage}{0.24\linewidth}
    \begin{tikzpicture}
    \draw  node[inner sep=0] (oursnvos_fern){{\includegraphics[width=\linewidth]{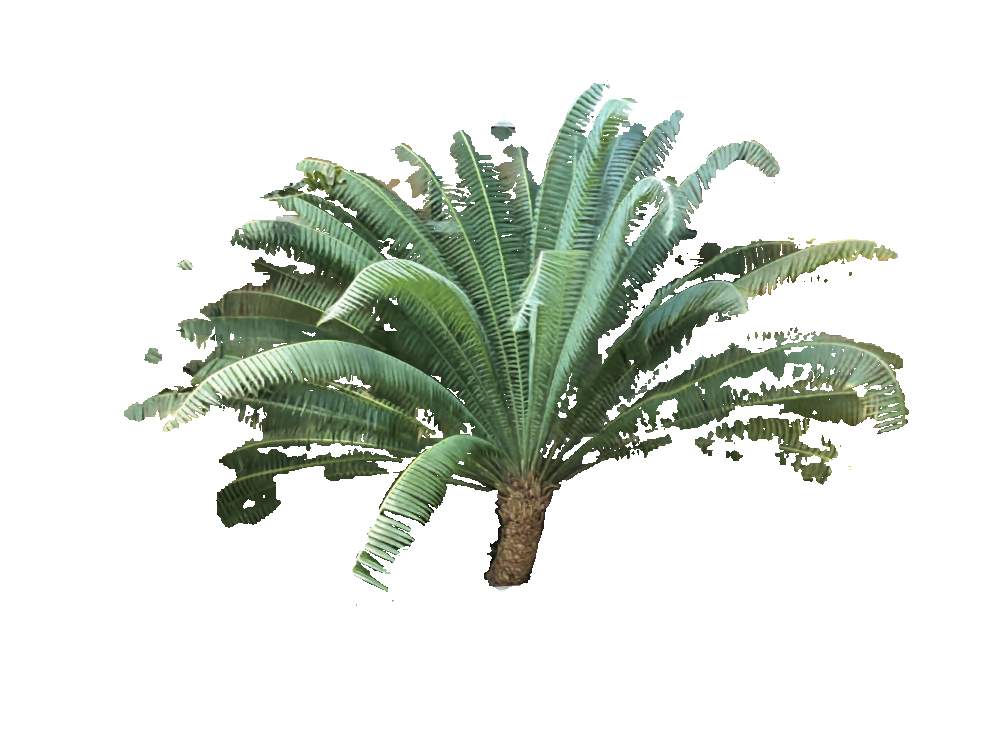}}};
    \draw  node[inner sep=0, below = 0.01 of oursnvos_fern] (oursnvos_trex) {{\includegraphics[width=\linewidth]{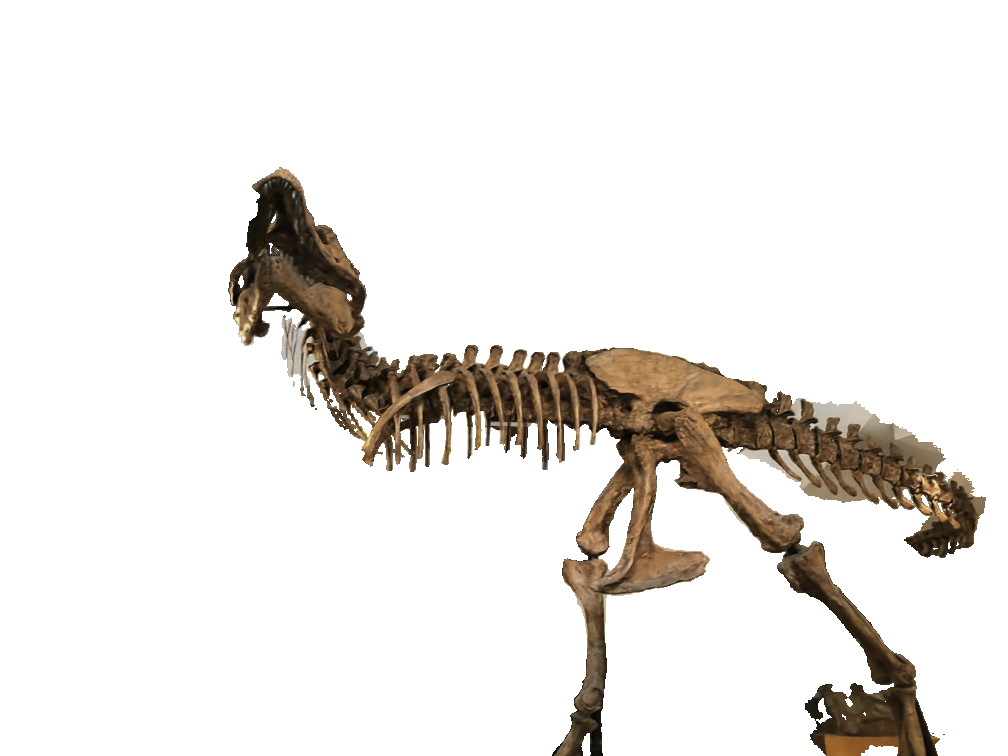}}};
    {\color{orange}\draw (-1, -2) node { Ours };}
    {\color{orange}\draw (-0.0, -2) node { NVOS };}
    {\color{orange}\draw (1, -2) node { Stroke };}
    \end{tikzpicture}
    \end{minipage}  
    \begin{minipage}{0.24\linewidth}
    \begin{tikzpicture}
    \draw  node[inner sep=0] (ours_best_fern){{\includegraphics[width=\linewidth]{assets/images/nvos/best_fern.png}}};
    \draw  node[inner sep=0, below = 0.01 of ours_best_fern] (ours_best_trex){{\includegraphics[width=\linewidth]{assets/images/nvos/best_trex.png}}};
    {\color{teal}\draw (0.25, -2) node { Ours Best};}
    \end{tikzpicture}
    \end{minipage}  
    \caption{{\em left to right:} Reference segmentation using NVOS professionally segmented mask, Result of NVOS\cite{nvos}, Our result using NVOS stroke, Our result using additional strokes. The quantitative comparisons are mentioned in the main document where our method performs better than NVOS even when using NVOS strokes. Please zoom using \emph{Adobe Acrobat/Okular} reader to see the details.}
    \label{fig:nvos_comp}
\end{figure*}
        \begin{figure*}
        \centering
        % garden
        \begin{minipage}{0.24\linewidth}
            \centering
            \frame{\includegraphics[width=\textwidth, height=3cm]{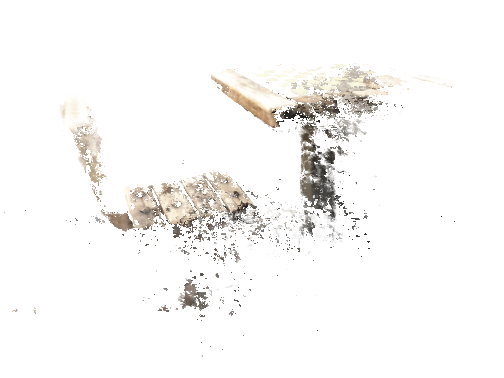}}
        \end{minipage}
        \begin{minipage}{0.24\linewidth}
            \centering
            \frame{\includegraphics[width=\textwidth, height=3cm]{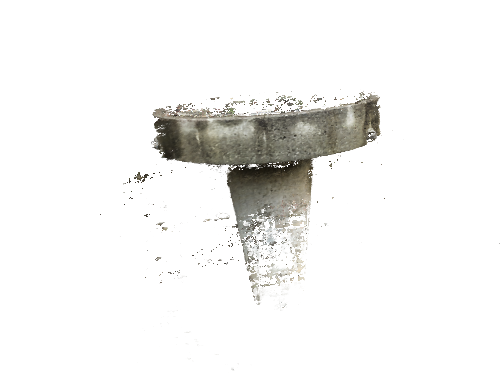}}
        \end{minipage}
        \begin{minipage}{0.24\linewidth}
            \centering
            \frame{\includegraphics[width=\textwidth, height=3cm]{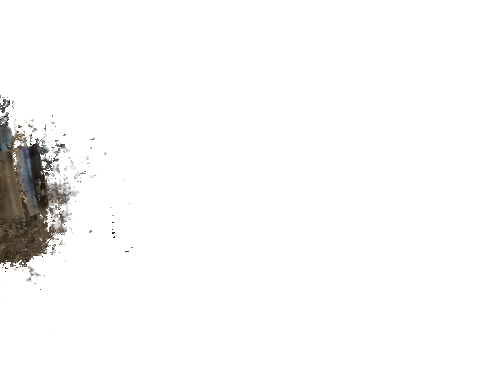}}
        \end{minipage}
        \begin{minipage}{0.24\linewidth}
            \centering
            \frame{\includegraphics[width=\textwidth, height=3cm]{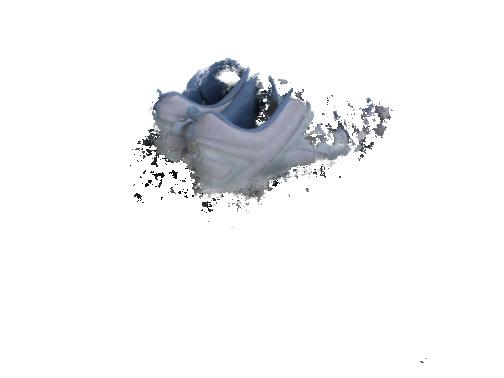}}
        \end{minipage}
        
        \begin{minipage}{0.24\linewidth}
            \centering
            \frame{\includegraphics[width=\textwidth, height=3cm]{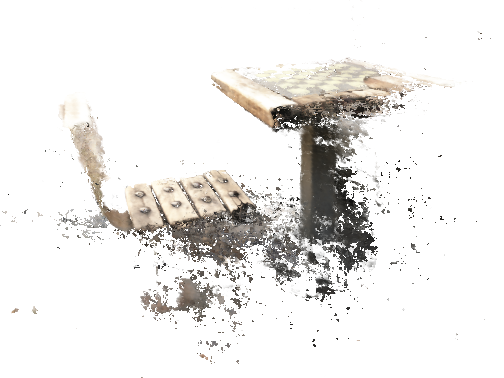}}
        \end{minipage}
        \begin{minipage}{0.24\linewidth}
            \centering
            \frame{\includegraphics[width=\textwidth, height=3cm]{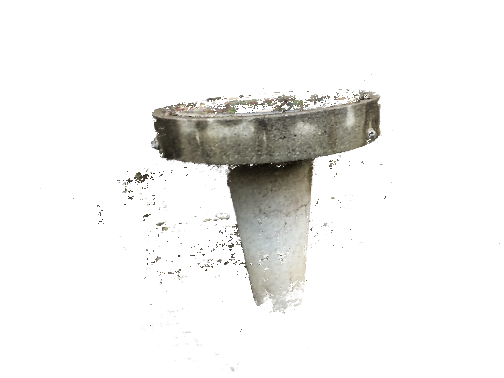}}
        \end{minipage}
        \begin{minipage}{0.24\linewidth}
            \centering
            \frame{\includegraphics[width=\textwidth, height=3cm]{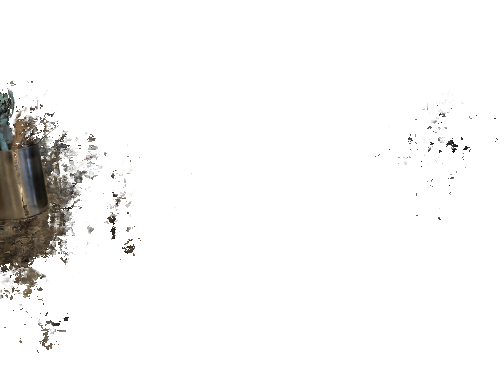}}
        \end{minipage}
        \begin{minipage}{0.24\linewidth}
            \centering
            \frame{\includegraphics[width=\textwidth, height=3cm]{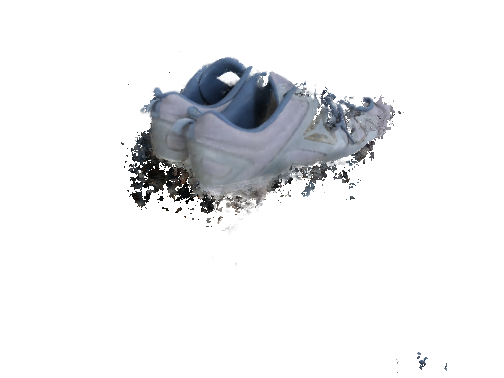}}
        \end{minipage}
        
        \begin{minipage}{0.24\linewidth}
            \centering
            \frame{\includegraphics[width=\textwidth, height=3cm]{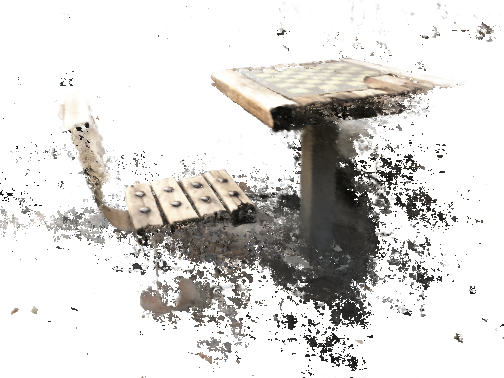}}
        \end{minipage}
        \begin{minipage}{0.24\linewidth}
            \centering
            \frame{\includegraphics[width=\textwidth, height=3cm]{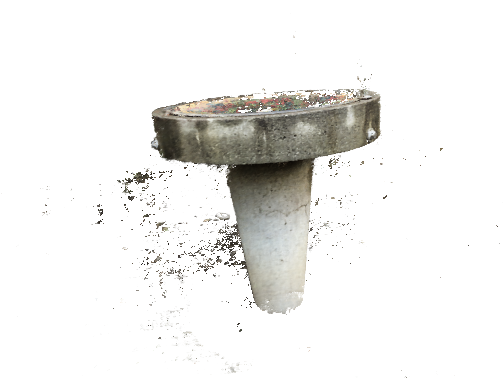}}
        \end{minipage}
        \begin{minipage}{0.24\linewidth}
            \centering
            \frame{\includegraphics[width=\textwidth, height=3cm]{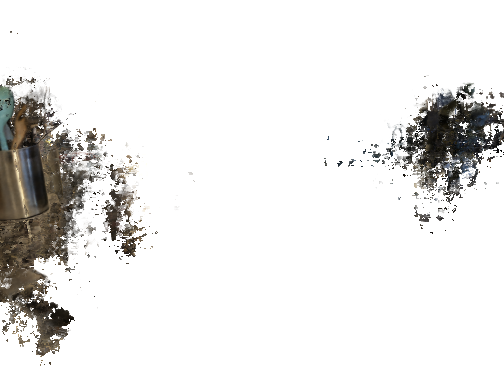}}
        \end{minipage}
        \begin{minipage}{0.24\linewidth}
            \centering
            \frame{\includegraphics[width=\textwidth, height=3cm]{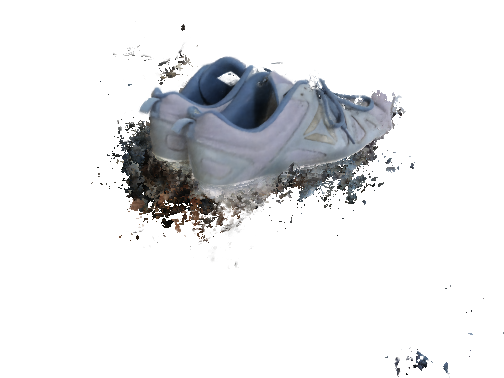}}
        \end{minipage}
        
        \begin{minipage}{0.24\linewidth}
        \centering
            \frame{\includegraphics[width=\textwidth, height=3cm]{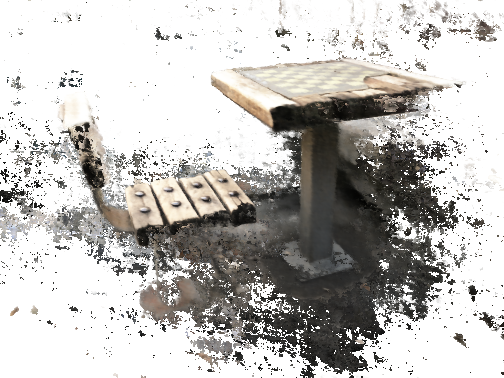}}
            \subcaption{\chesstable}
        \end{minipage}
        \begin{minipage}{0.24\linewidth}
        \centering
            \frame{\includegraphics[width=\textwidth, height=3cm]{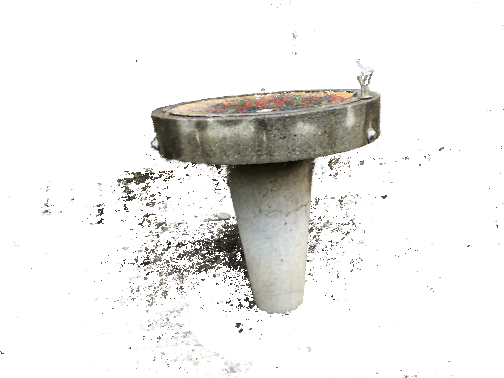}}
            \subcaption{\colorfountain}
        % \subcaptionbox{\colorfountain}
        \end{minipage}
        \begin{minipage}{0.24\linewidth}
        \centering
            \frame{\includegraphics[width=\textwidth, height=3cm]{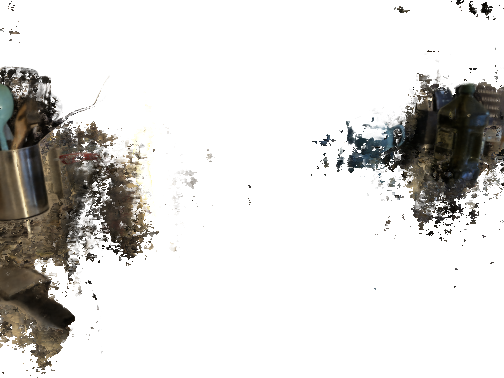}}
            \subcaption{\stove}
        \end{minipage}
        \begin{minipage}{0.24\linewidth}
        \centering
            \frame{\includegraphics[width=\textwidth, height=3cm]{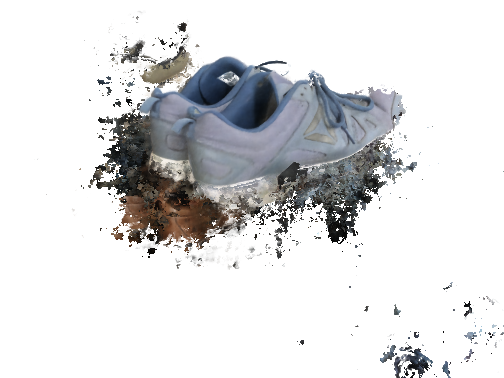}}
            \subcaption{\shoerack}
        % \subcaptionbox{\shoerack}
        \end{minipage}
        \caption{\emph{N3F/DFF Results}:
        {In this figure we show result of DFF/N3F\cite{N3F,DFF} on different thresholds and we reported the best of their method in main document. It can be seen that despite varying the thresholds the result is poorly segmented. The background objects are starting to bleed into the foreground. For the results of our method on the same scenes, please refer to the main paper.}}
        \label{fig:all_n3f}
\end{figure*}
        \begin{figure*}
    \centering
        \includegraphics[width=0.8\linewidth]{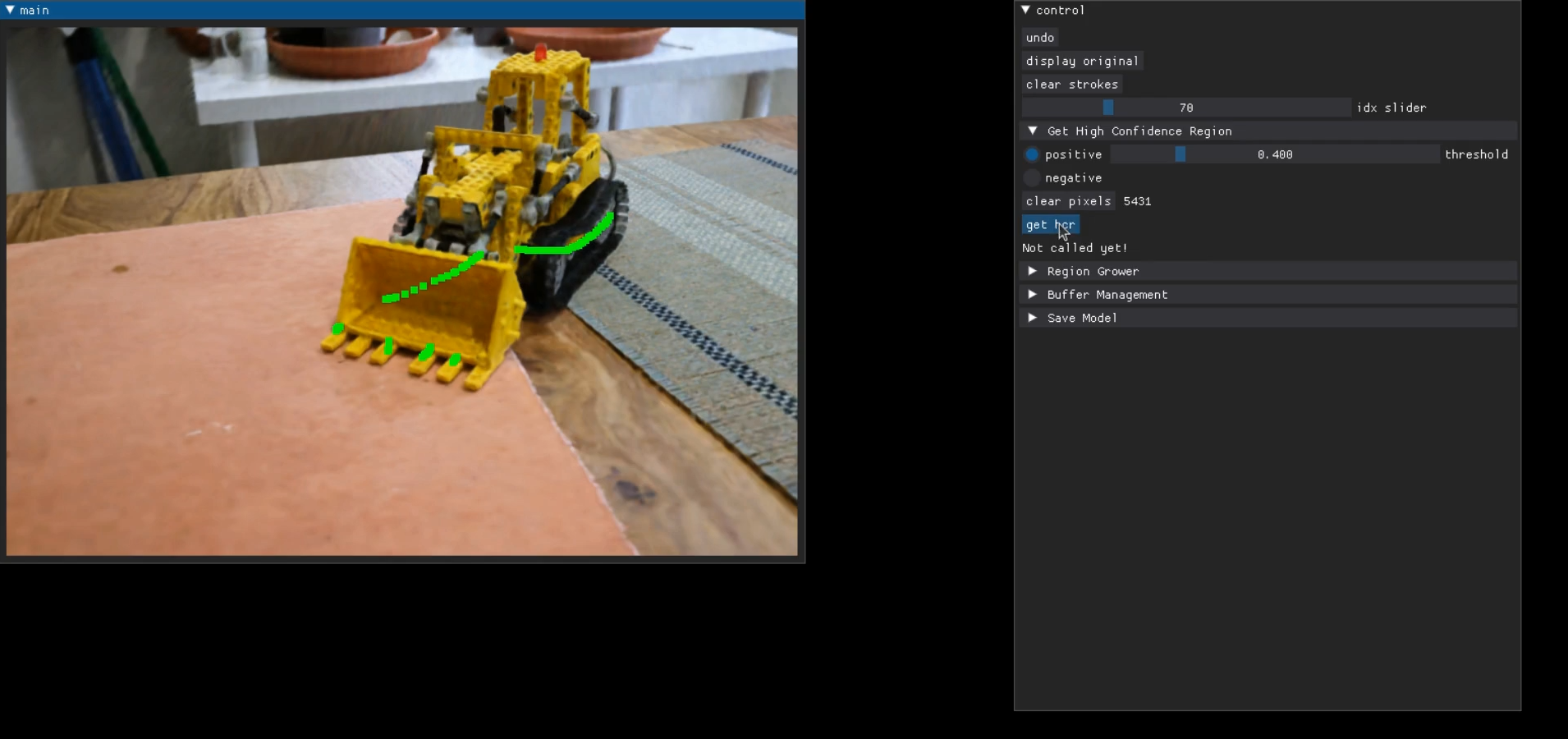}
        \includegraphics[width=0.8\linewidth]{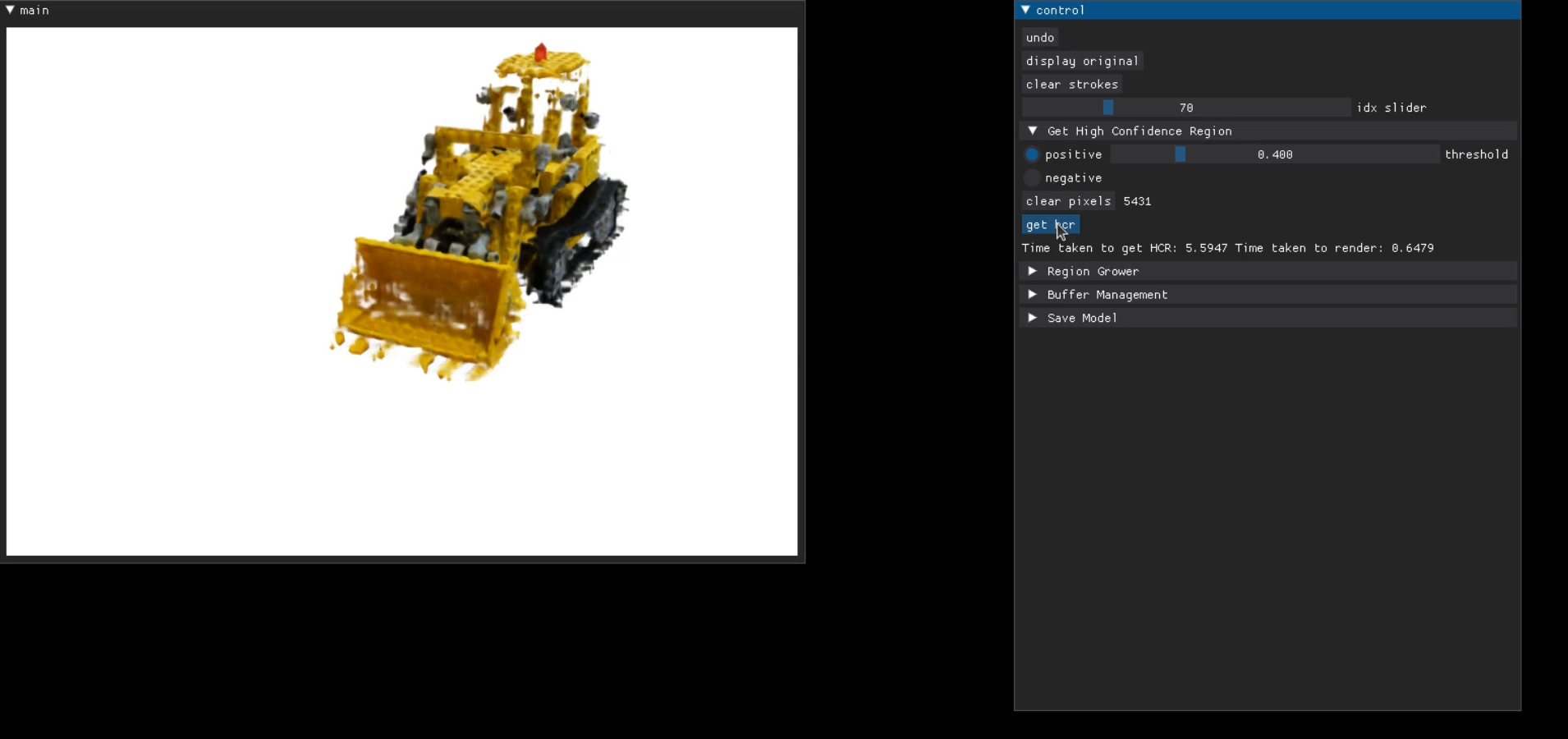}
        \includegraphics[width=0.8\linewidth]{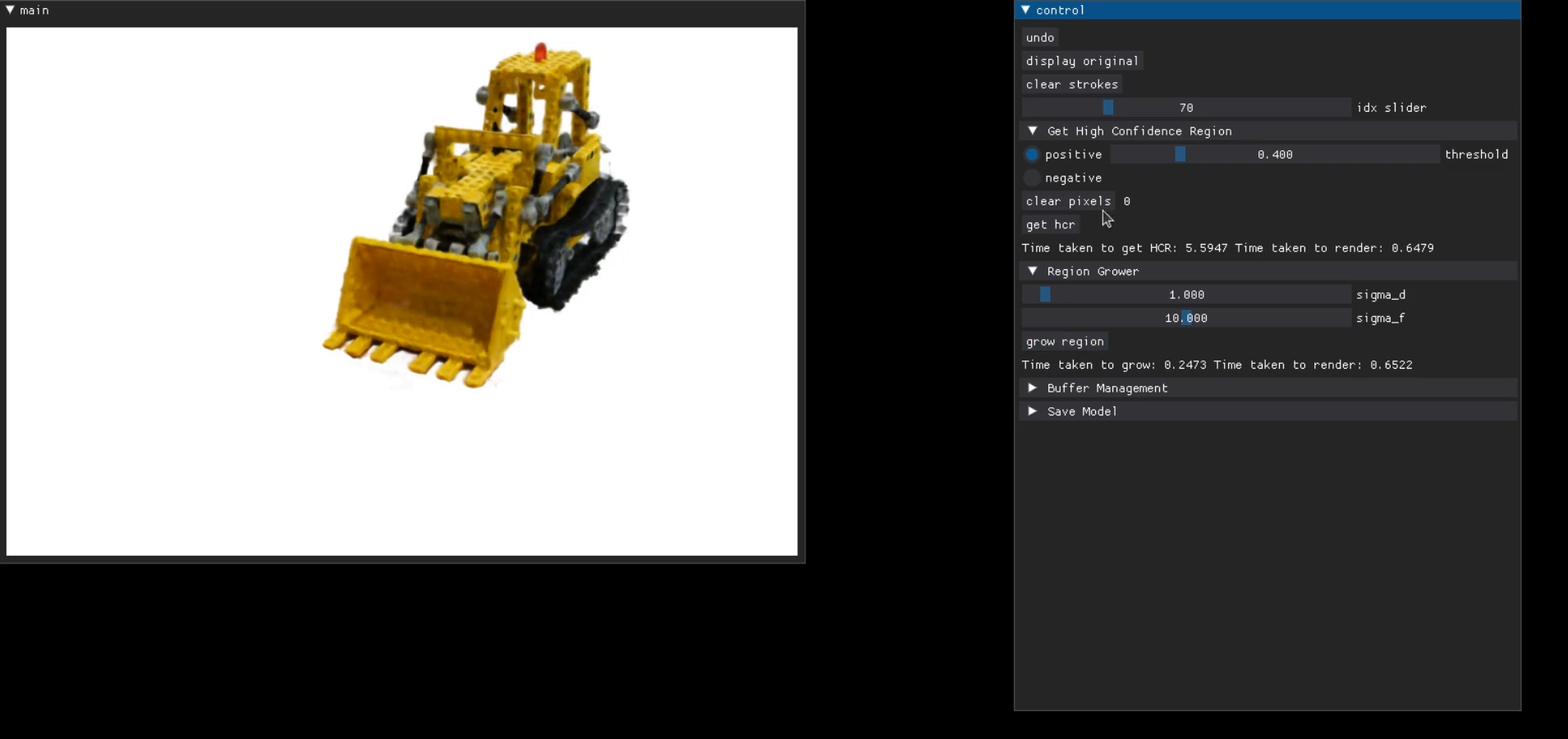}
    \caption{{\em Interactive GUI Tool:} We also release an easy-to-use interactive GUI tool which can be used to draw strokes and segment radiance fields.}
    \label{fig:gui_tool}
\end{figure*}

        \begin{figure*}[!h]
    \centering
    \rotatebox[origin=c]{90}{Teacher DINO\Bstrut}
    \rotatebox[origin=c]{90}{(\garden)\Bstrut}
    \begin{minipage}{0.23\linewidth}
        \centering
        {\includegraphics[width=\textwidth, height=3cm]{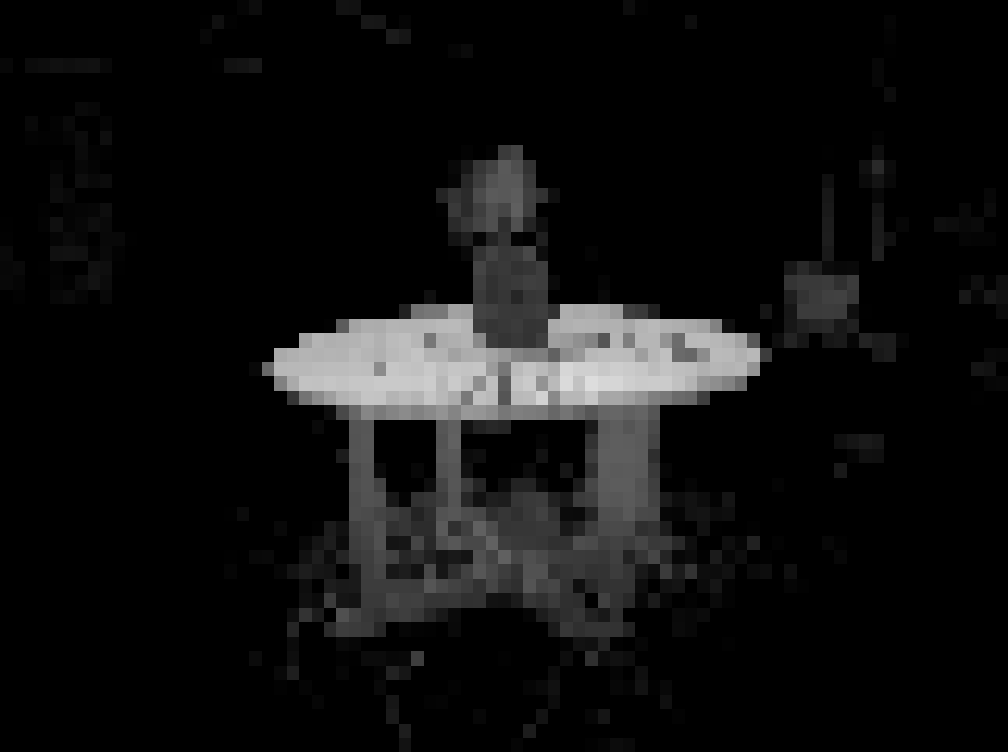}}
    \end{minipage}
    \begin{minipage}{0.23\linewidth}
        \centering
        {\includegraphics[width=\textwidth, height=3cm]{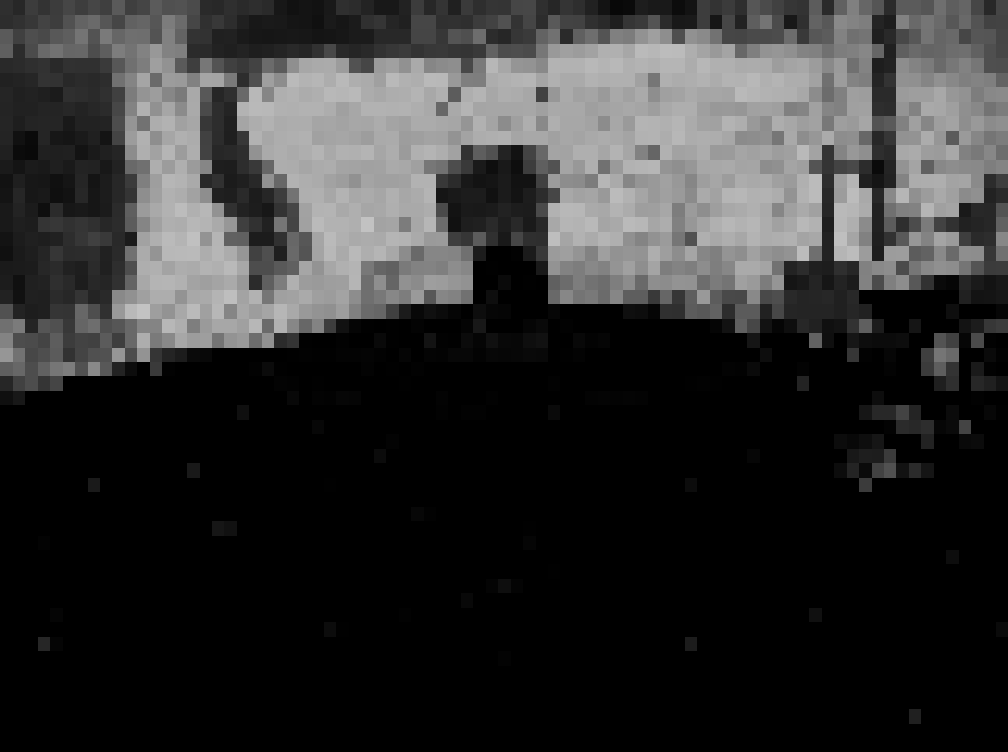}}
    \end{minipage}
    \begin{minipage}{0.23\linewidth}
        \centering
        {\includegraphics[width=\textwidth, height=3cm]{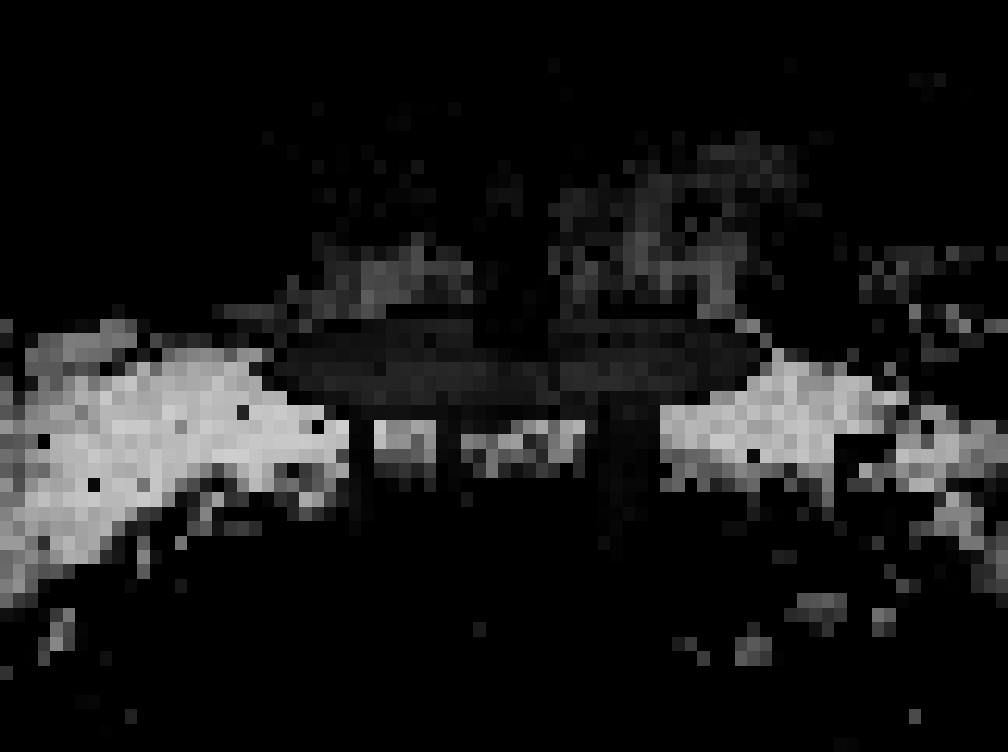}}
    \end{minipage}
    \begin{minipage}{0.23\linewidth}
        \centering
        {\includegraphics[width=\textwidth, height=3cm]{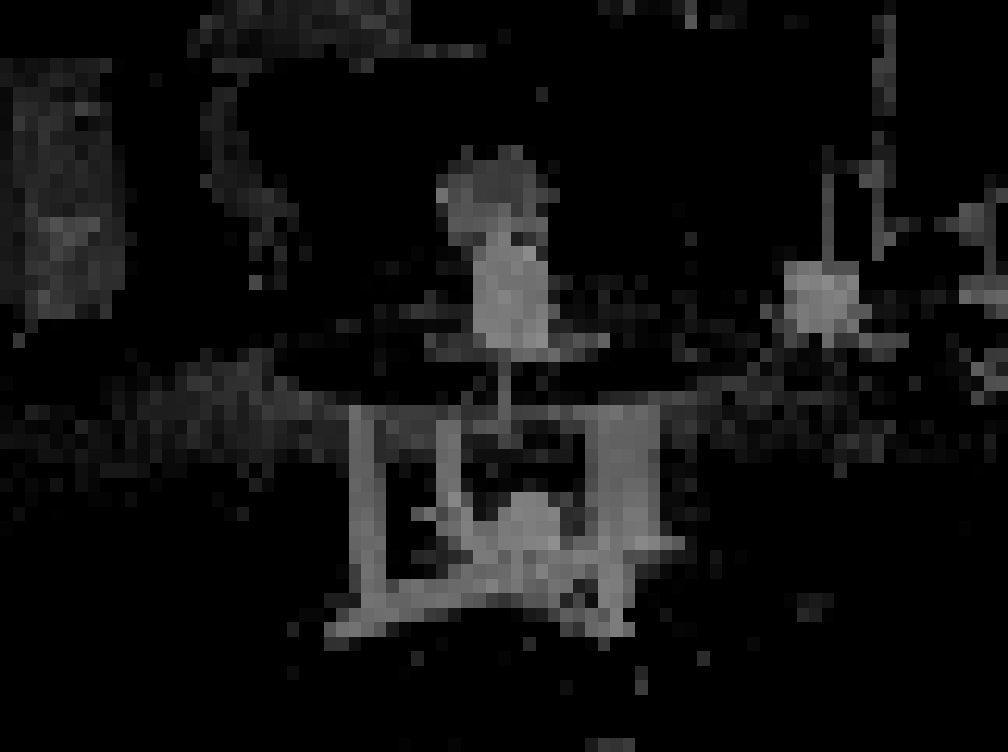}}
    \end{minipage}
    
    \rotatebox[origin=c]{90}{Distilled\Bstrut}
    \rotatebox[origin=c]{90}{(\garden)\Bstrut}
    \begin{minipage}{0.23\linewidth}
        \centering
        {\includegraphics[width=\textwidth, height=3cm]{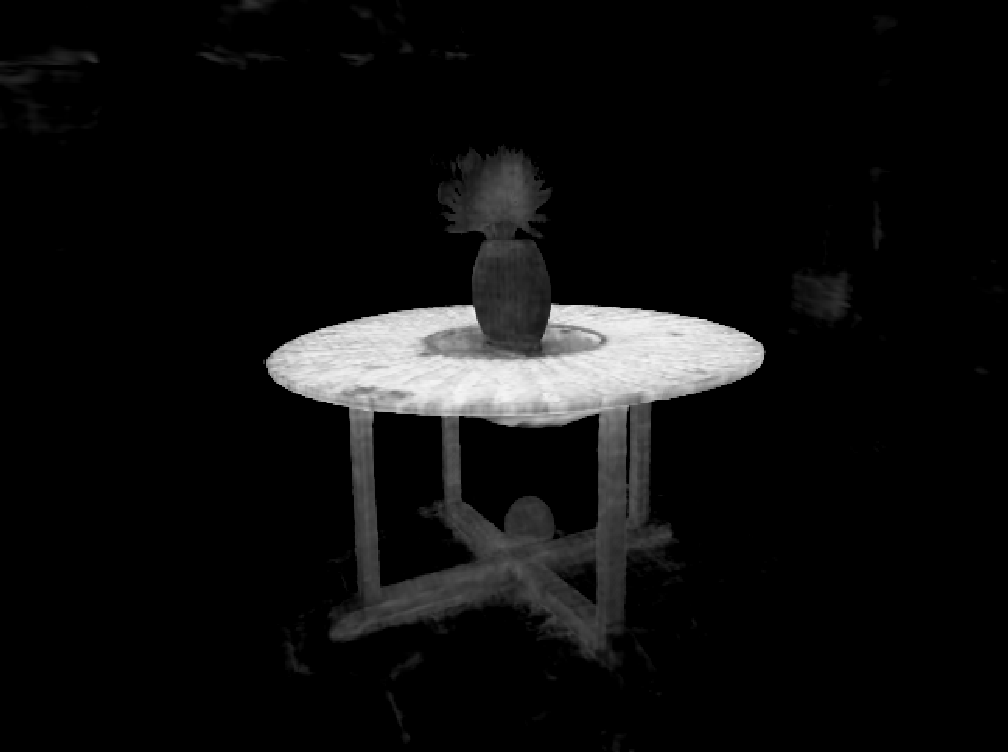}}
    \end{minipage}
    \begin{minipage}{0.23\linewidth}
        \centering
        {\includegraphics[width=\textwidth, height=3cm]{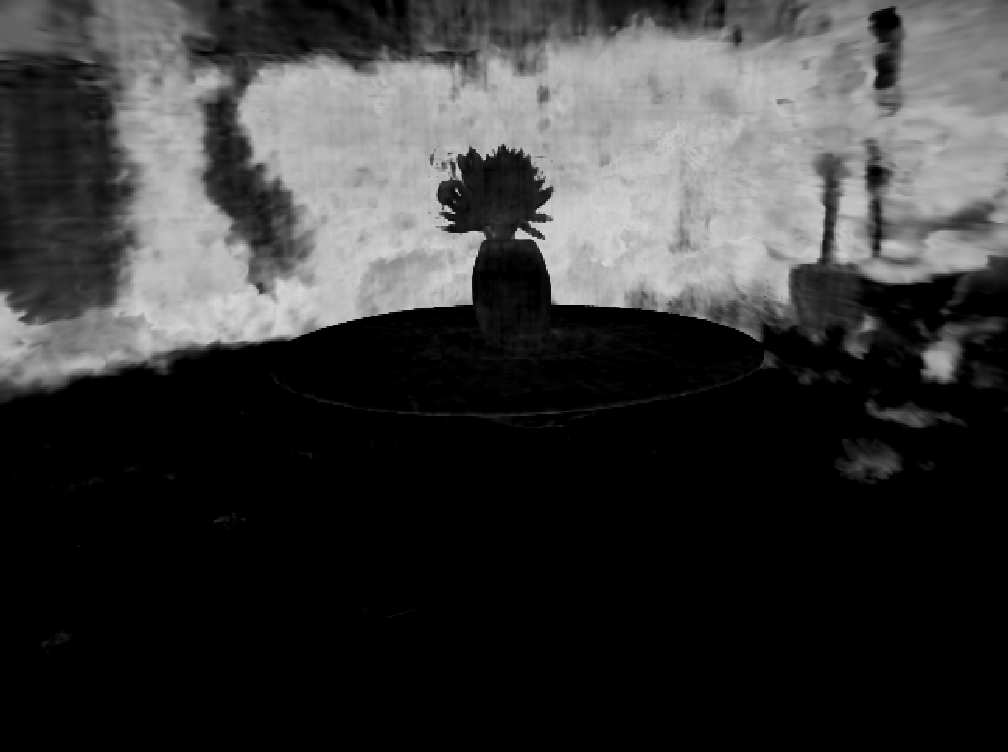}}
    \end{minipage}
    \begin{minipage}{0.23\linewidth}
        \centering
        {\includegraphics[width=\textwidth, height=3cm]{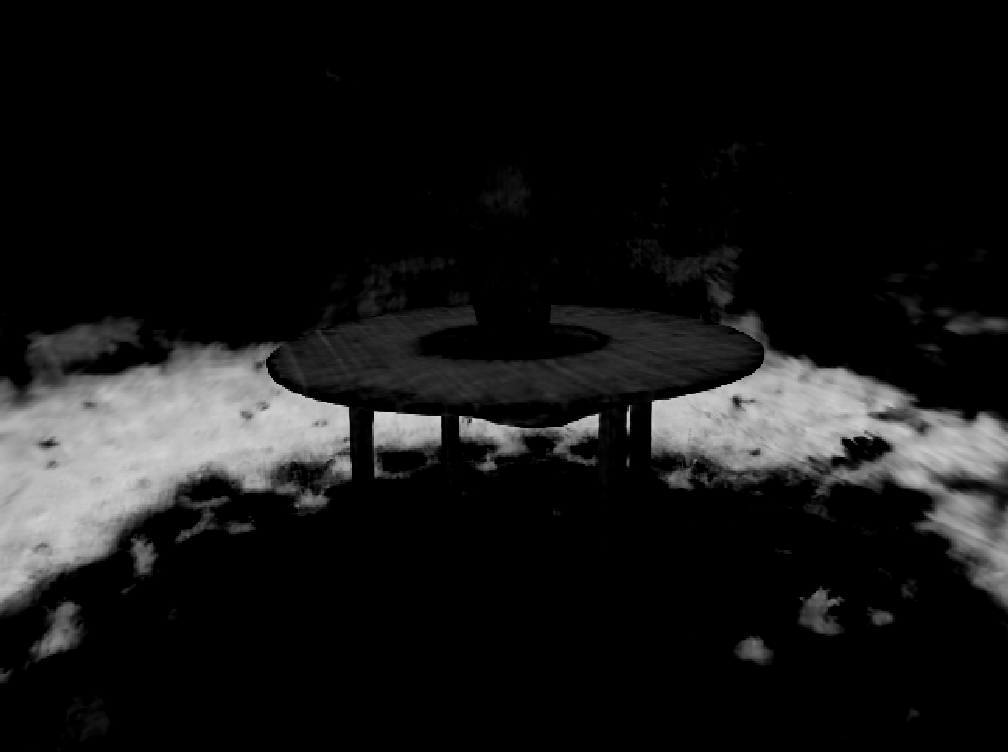}}
    \end{minipage}
    \begin{minipage}{0.23\linewidth}
        \centering
        {\includegraphics[width=\textwidth, height=3cm]{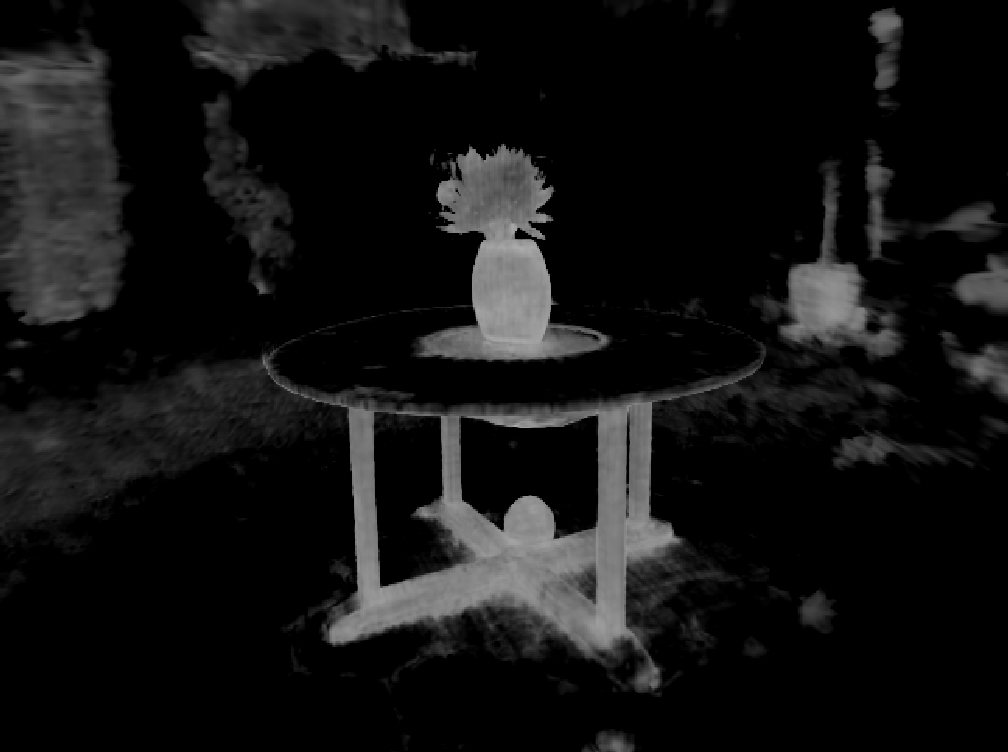}}
    \end{minipage}
    
    % kitchen
    \rotatebox[origin=c]{90}{Teacher DINO\Bstrut}
    \rotatebox[origin=c]{90}{(\kitchen)\Bstrut}
    \begin{minipage}{0.23\linewidth}
        \centering
        {\includegraphics[width=\textwidth, height=3cm]{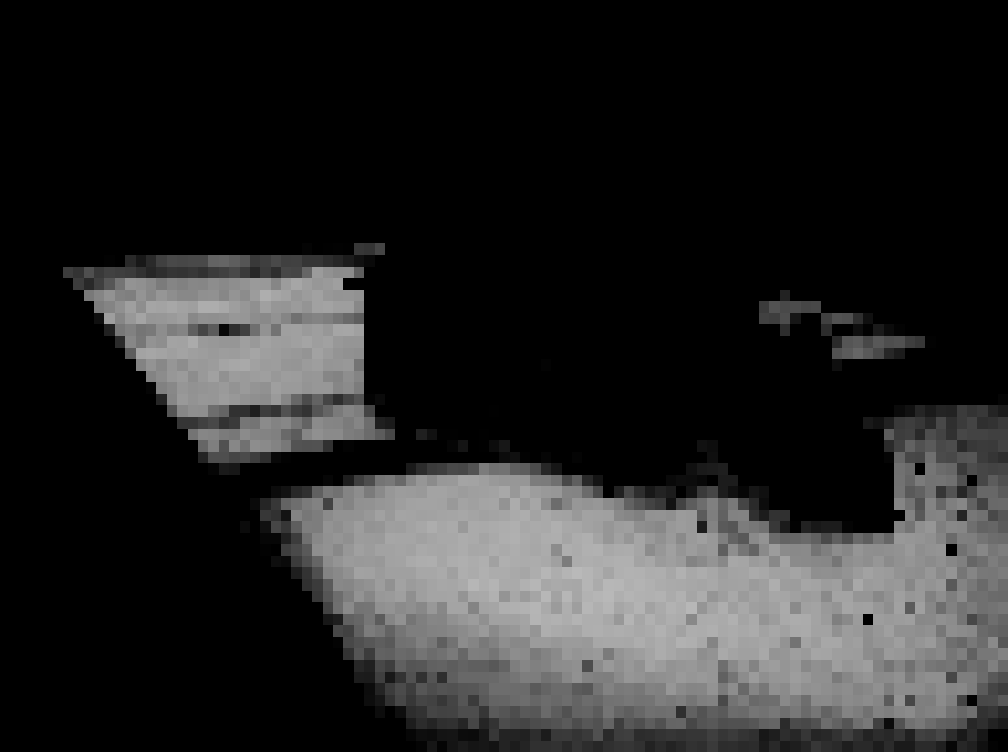}}
    \end{minipage}
    \begin{minipage}{0.23\linewidth}
        \centering
        {\includegraphics[width=\textwidth, height=3cm]{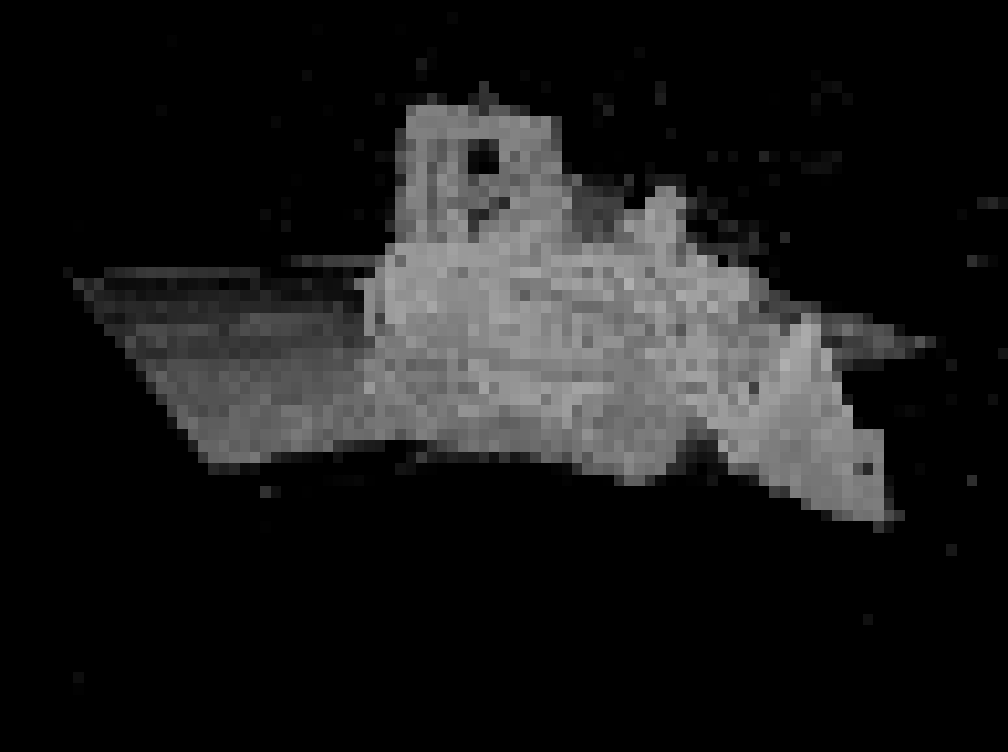}}
    \end{minipage}
    \begin{minipage}{0.23\linewidth}
        \centering
        {\includegraphics[width=\textwidth, height=3cm]{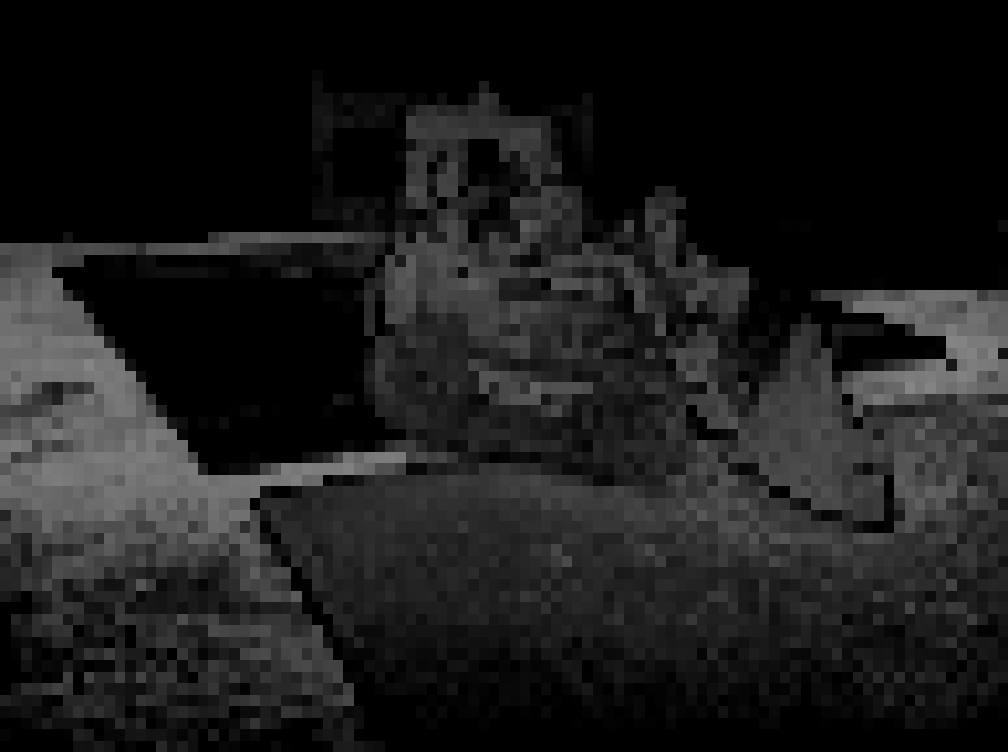}}
    \end{minipage}
    \begin{minipage}{0.23\linewidth}
        \centering
        {\includegraphics[width=\textwidth, height=3cm]{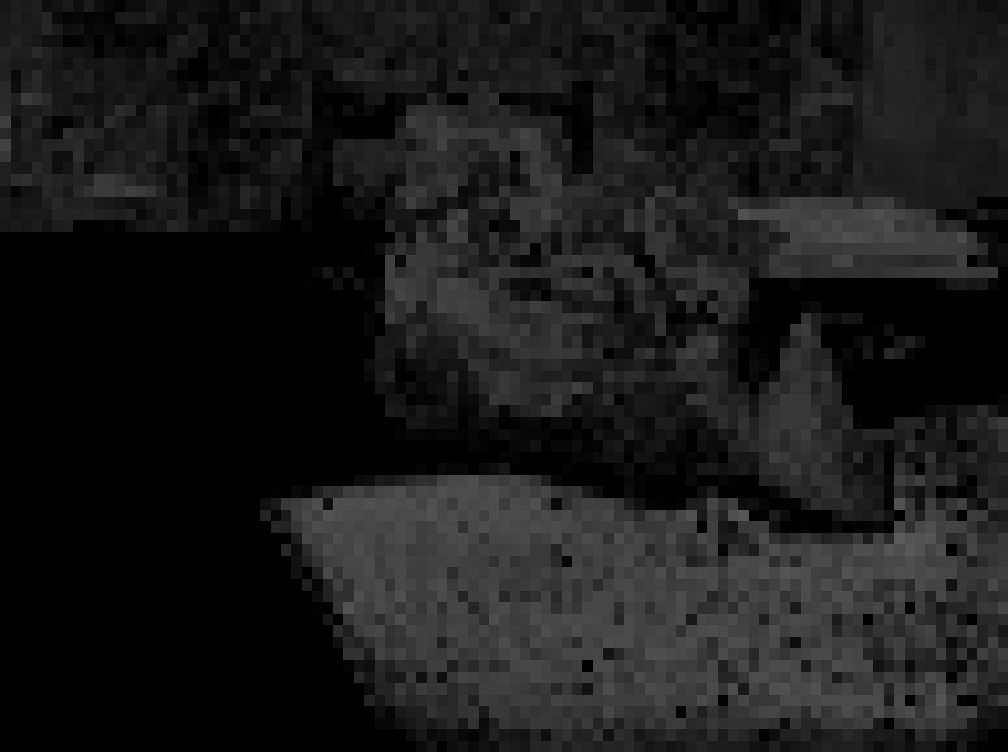}}
    \end{minipage}
    
    \rotatebox[origin=c]{90}{Distilled\Bstrut}
    \rotatebox[origin=c]{90}{(\kitchen)\Bstrut}
    \begin{minipage}{0.23\linewidth}
        \centering
        {\includegraphics[width=\textwidth, height=3cm]{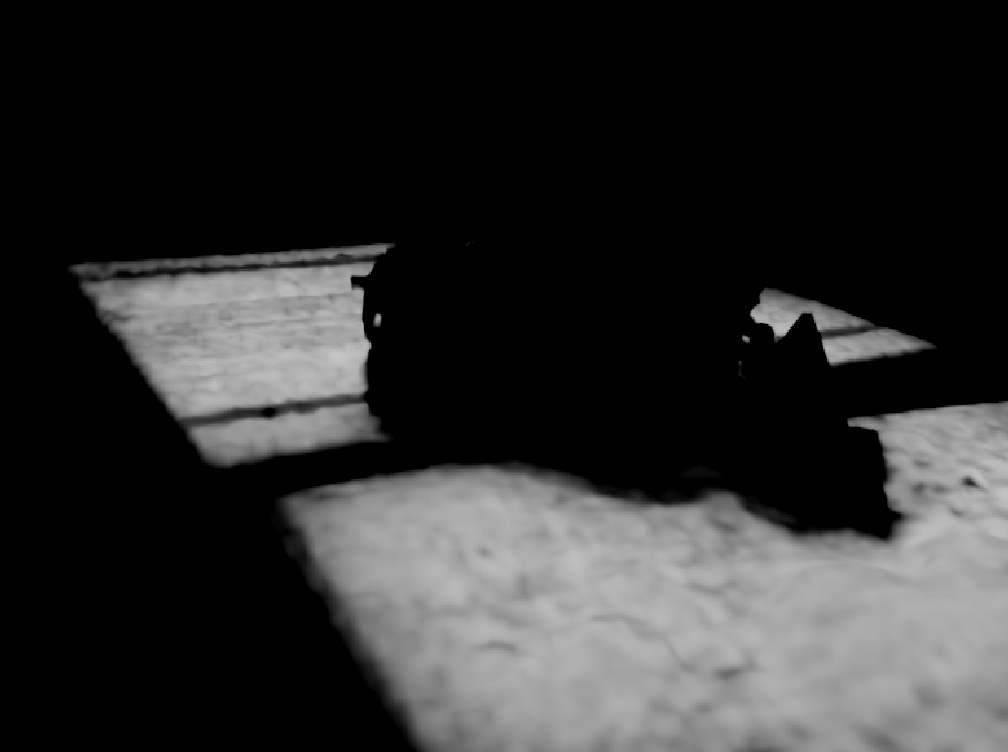}}
    \end{minipage}
    \begin{minipage}{0.23\linewidth}
        \centering
        {\includegraphics[width=\textwidth, height=3cm]{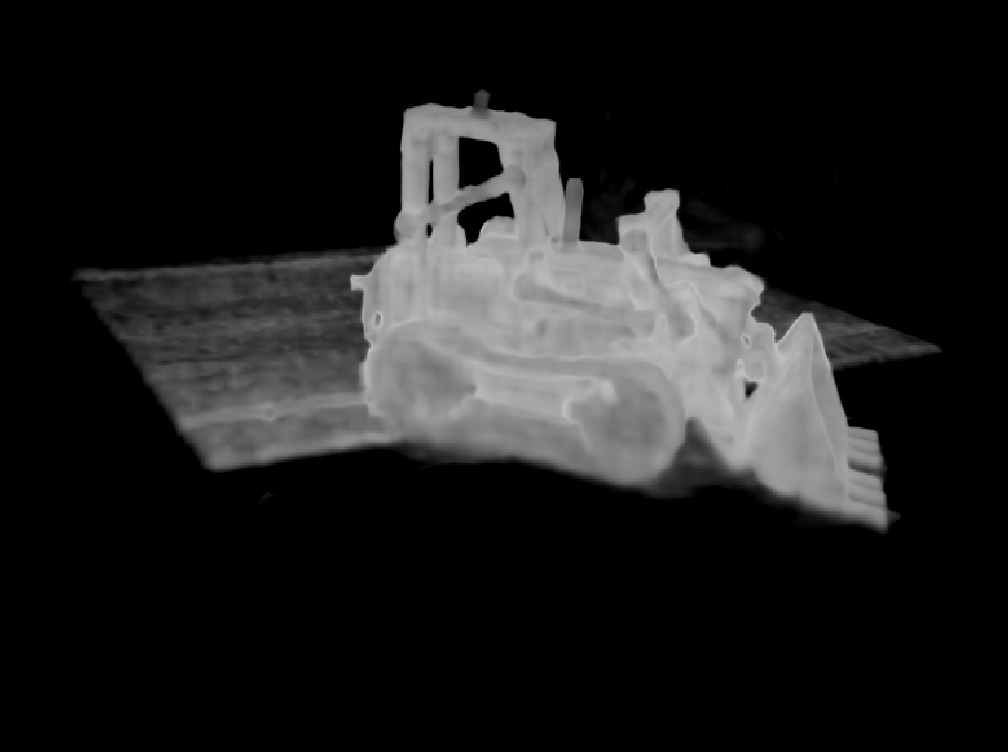}}
    \end{minipage}
    \begin{minipage}{0.23\linewidth}
        \centering
        {\includegraphics[width=\textwidth, height=3cm]{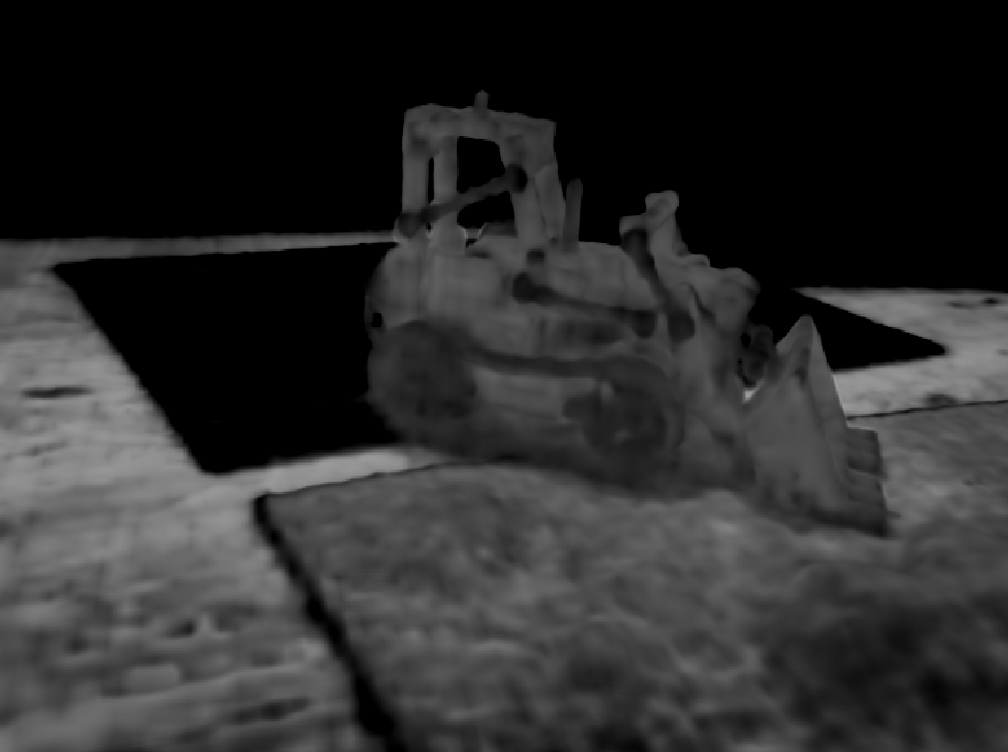}}
    \end{minipage}
    \begin{minipage}{0.23\linewidth}
        \centering
        {\includegraphics[width=\textwidth, height=3cm]{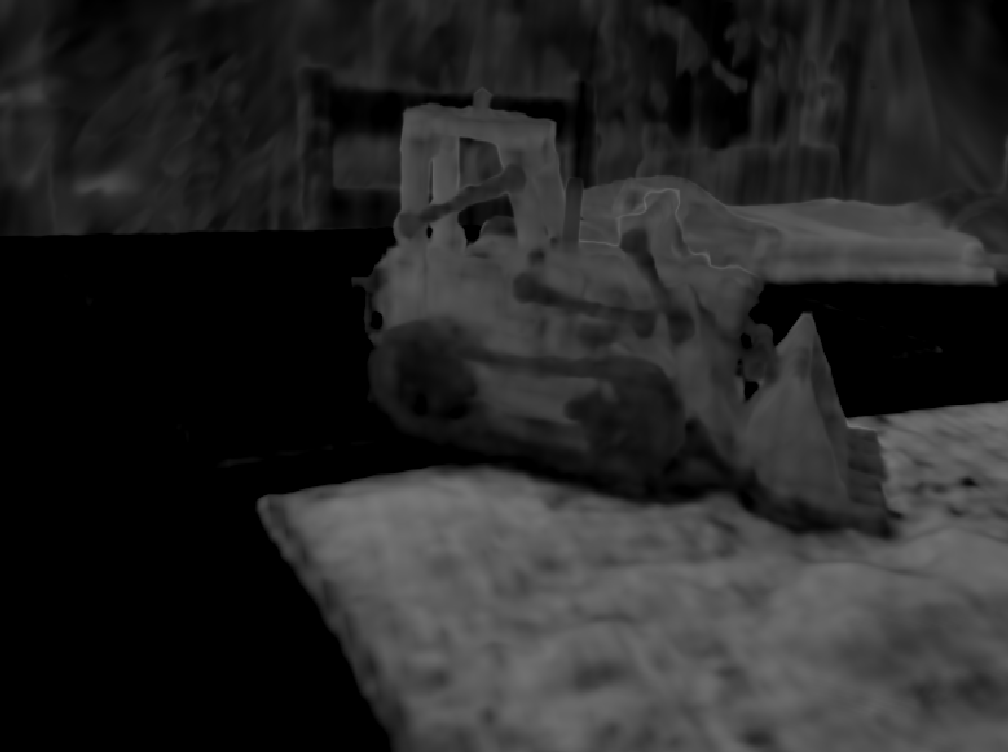}}
    \end{minipage}
    
    \caption{\emph{Student Surpasses Teacher}:
    {The 4 columns of this figure shows the DINO features used as teacher vs the ones learnt by student post optimization. Since, the student learns finer features than the teacher due to assistance from the volumetric density, we can claim that the student surpasses the teacher. This is consistent with the prior art N3F and DFF.}}
    \label{fig:dino_features}
\end{figure*}
    }

    %%%%%%%%% REFERENCES
\end{document}